\PassOptionsToPackage{table}{xcolor}
\documentclass[lettersize,journal]{IEEEtran}
\usepackage{amsmath,amsfonts}
\usepackage{svg}
\usepackage{algorithmic}
\usepackage{algorithm}
\usepackage{array}
\usepackage[caption=false,font=normalsize,labelfont=sf,textfont=sf]{subfig}
\usepackage{textcomp}
\usepackage{stfloats}
\usepackage{url}
\usepackage{verbatim}
\usepackage{graphicx}
\usepackage{cite}
\usepackage{multirow} 
\usepackage{makecell}
\usepackage{array}
\usepackage{booktabs}
\usepackage{skeldoc}
\usepackage{pifont}
\usepackage{footnote}
\usepackage{caption}
\usepackage[utf8]{inputenc}
\definecolor{TUMBlue4}{RGB}{152, 198, 234}
\newcommand{\cmark}{\textcolor{green!60!black}{\ding{51}}}
\newcommand{\xmark}{\textcolor{red!70!black}{\ding{55}}}

\begin{document}

\title{To New Beginnings: A Survey of Unified Perception in Autonomous Vehicle Software}

\author{Loïc Stratil, Felix Fent, Esteban Rivera, Markus Lienkamp
\thanks{The authors are with the Technical University of Munich, Germany; School of Engineering \& Design, Department of Mobility Systems Engineering, Institute of Automotive Technology and Munich Institute of Robotics and Machine Intelligence (MIRMI) (e-mail: loic.stratil@tum.de) Manuscript submitted August 28, 2025.}%
\thanks{This work has been submitted to the IEEE for possible publication. Copyright may be transferred without notice, after which this version may no longer be accessible.}}

\markboth{}%
{Shell \MakeLowercase{\textit{et al.}}: A Sample Article Using IEEEtran.cls for IEEE Journals}

\maketitle

\begin{abstract}
Autonomous vehicle perception typically relies on modular pipelines that decompose the task into detection, tracking, and prediction. While interpretable, these pipelines suffer from error accumulation and limited inter-task synergy. Unified perception has emerged as a promising paradigm that integrates these sub-tasks within a shared architecture, potentially improving robustness, contextual reasoning, and efficiency while retaining interpretable outputs. In this survey, we provide a comprehensive overview of unified perception, introducing a holistic and systemic taxonomy that categorizes methods along task integration, tracking formulation, and representation flow. We define three paradigms -Early, Late, and Full Unified Perception- and systematically review existing methods, their architectures, training strategies, datasets used, and open-source availability, while highlighting future research directions. This work establishes the first comprehensive framework for understanding and advancing unified perception, consolidates fragmented efforts, and guides future research toward more robust, generalizable, and interpretable perception.
\end{abstract}


\section{Introduction}
\label{chap:Introduction}

Autonomous driving software is typically organized into four modules: sensing, perception, planning, and control. Sensing collects information about the environment, mostly through cameras, LiDAR, or RADAR. Perception then interprets this information, providing the basis for planning, which generates safe and efficient trajectories, and control executing these.

In this construct, perception plays a central role, enabling the vehicle to interpret its surroundings, understand the behavior of road users, and anticipate their future motion, thereby making sense of the environment. Perception can further be divided into two key sub-tasks:
\begin{itemize}
\item Localization – Where am I?
\item Scene Understanding – Where is what, going where?
\end{itemize}
While localization ensures precise local and global ego positioning, scene understanding covers detection, tracking, and prediction of surrounding objects, providing a comprehensive view of the ego vehicle's environment. This survey focuses on the latter, summarizing perception as scene understanding.

Perception traditionally follows modular pipelines, where detection, tracking, and prediction are handled in separate sub-modules, as in widely used platforms such as Autoware \cite{kato2018autoware} and Apollo \cite{apollo2017}. Modular pipelines are interpretable but suffer from error accumulation and limited information flow between sub-tasks.

End-to-end autonomous driving stacks, in contrast, map raw sensor data directly to planning and sometimes control outputs \cite{hu2023planning, marcu2023lingoqa, li2024hydra}. This tightly integrates sensing, perception, planning, and control, enabling holistic reasoning across tasks. However, explicit perception outputs are lost, reducing explainability and complicating validation in safety-critical settings. In addition, learning the entire driving task at once is highly complex, making such systems difficult to train. Despite these challenges, end-to-end approaches are increasingly adopted both in research, such as UniAD \cite{hu2023planning}, and in industry, with Wayve \cite{marcu2023lingoqa} and NVIDIA \cite{li2024hydra}.

\begin{figure*}[!t]
\centering
\includegraphics[width=6in]{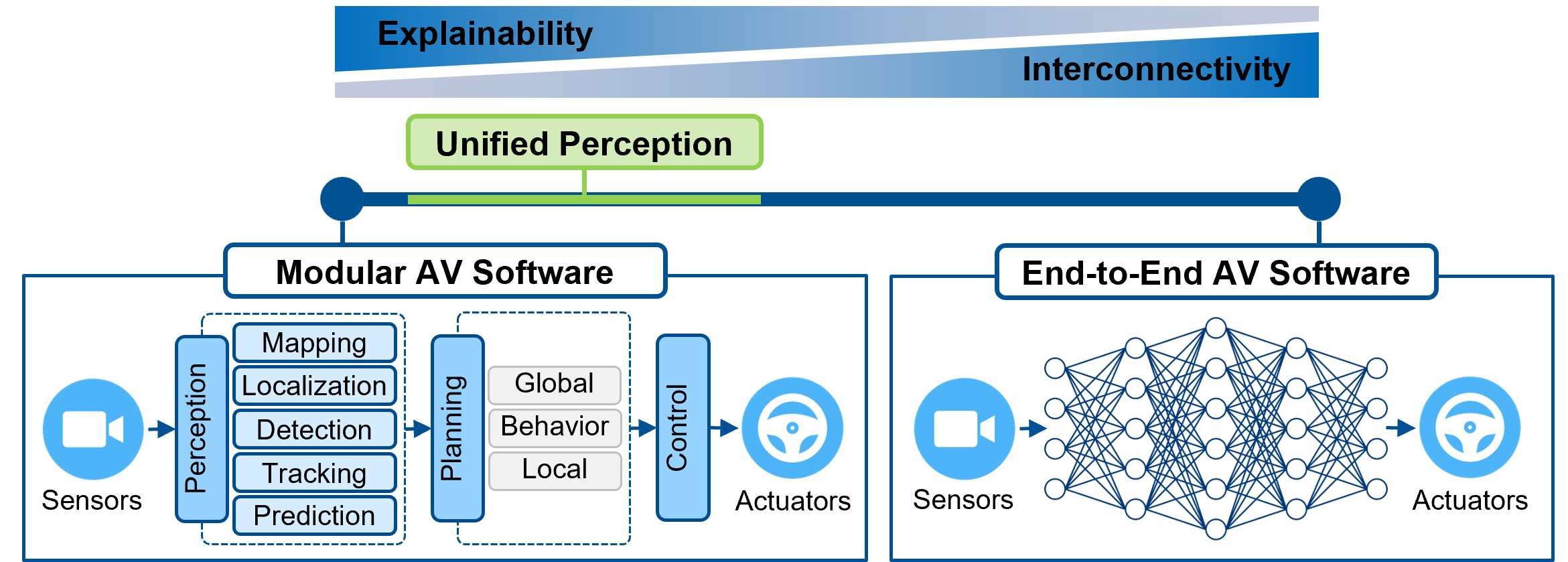}
\caption{Perception is either modular or implicitly learned within end-to-end driving software stacks. Unified perception, an extension to modular approaches, balances explainability and interconnectivity towards robust and generalized perception.}
\label{Modular_vs_E2E}
\end{figure*}

This contrast highlights a key gap: current perception systems either prioritize modular interpretability or global reasoning through interconnectivity, as illustrated in Fig. ~\ref{Modular_vs_E2E}. Unified perception, as an extension of modular pipelines, addresses this gap, integrating scene understanding into a unified module while retaining interpretable perception outputs. Several methods have been proposed in this direction, but vary in scope and design, and no systematic overview exists. 

This survey consolidates existing unified perception approaches into a holistic and systemic taxonomy based on input, intermediate and output representation as well as tracking formulation. From this taxonomy, three levels of task integration are deduced: Early Unified Perception (EUP), unifying detection and tracking; Late Unified Perception (LUP), unifying tracking and prediction; and Full Unified Perception (FUP), unifying detection, tracking, and prediction. We further provide an overview of such architectures, quantitative analysis, training strategies, datasets used, and open-source availability, highlighting the current state of unified perception.

The remainder of this survey is organized as follows: Sec. \ref{chap:StateOfTheArt} introduces the state of the art of modular perception and highlights the unified perception survey gap. Sec. \ref{chap:UnifiedPerceptionTaxonomy} introduces the proposed taxonomy. Secs. \ref{chap:EarlyUnifiedPerception}–\ref{chap:FullUnifiedPerception} survey the three paradigms of EUP, LUP, and FUP. Sec. \ref{chap:Discussion} discusses insights and open challenges, and Sec. \ref{chap:Conclusion} concludes.

\section{State of the Art}
\label{chap:StateOfTheArt}

Perception is a critical task for autonomous driving, today mainly realized through modular approaches that solve it as a feedforward cascade of clearly defined, self-contained sub-modules. While these perform well on benchmarks, their real-world use exhibits limitations in robustness and generalizability \cite{song2024robustness, karle2022scenario}, motivating the development of unified perception. We first introduce the modular perception state of the art through its survey landscape, then introduce unified perception and highlight the existing survey gap.

\textbf{Modular perception} systems form an integral part of modular AV software (Fig. \ref{Modular_vs_E2E}) through a set of clearly defined sub-modules with dedicated objectives, interfaces, and evaluation metrics, maximizing interpretability and enabling parallel development. Raw sensor data is fed into a detection module to identify objects, a tracking module to associate them over time with unique IDs, and a prediction module to estimate future motion, with human-interpretable bounding box representations as intermediate interfaces that facilitate integration and benchmarking. The state of the art of 3D object detection \cite{song2024robustness, mao20233d, ma20233d, liu2023echoes, wang2023multi, yao2023radar, aung2024review}, multi-object tracking \cite{luo2021multiple, bashar2209multiple, agrawal2024systematic, hassan2024multi, rakai2022data, leon2021review}, and motion prediction \cite{leon2021review, huang2022survey, karle2022scenario, ding2023incorporating, bharilya2024machine, huang2023review, uhlemann2024evaluating} has been extensively surveyed, as introduced below.

\textbf{Object Detection} localizes and classifies objects within sensor inputs into 2D/3D bounding boxes. Mao et al. \cite{mao20233d} provide a holistic 3D detection overview, categorizing methods by sensor input (LiDAR, cameras, multi-modal) and working principle (network architecture) into a widely accepted taxonomy, with challenges, advantages, and limitations per method type and a quantitative comparison on Kitti \cite{geiger2013vision}, nuScenes \cite{caesar2020nuscenes}, and WOD \cite{sun2020scalability}. Song et al. \cite{song2024robustness} survey 3D detection with a focus on robustness, accuracy, and latency, adapting Mao et al.'s categorization while emphasizing each method type's resilience to environmental variation, sensor noise, and calibration errors, and the superior robustness of multi-modal systems. Modality-specific surveys include Ma et al. \cite{ma20233d} for cameras, Aung et al. \cite{aung2024review} for LiDAR, and Liu et al. \cite{liu2023echoes} for RADAR. Multi-modal detection is surveyed by Wang et al. \cite{wang2023multi}, and camera-RADAR fusion by Yao et al. \cite{yao2023radar}.

\textbf{Multi-Object Tracking} (MOT) assigns a consistent identifier to each detected object across timesteps. Modular systems typically use the tracking-by-detection paradigm, feeding detections into model- or learning-based modules, where the central challenge is real-time data association in dynamic, multi-object environments.

Model-based tracking is comprehensively surveyed by Luo et al. \cite{luo2021multiple}, who categorize methods by initialization type, processing mode, and output type (stochastic or deterministic), structured into feature extraction, data association, and track management. Such methods rely on motion-derived features, with Kalman filters widely used to extrapolate and refine trajectories via a prediction-update mechanism. Rakai et al. \cite{rakai2022data} dedicate a survey to data association strategies, evaluating accuracy, robustness, computational cost, and complexity, and identifying limitations in occlusion handling, efficiency, and generalization.

Learning-based tracking instead learns similarity metrics between instances from appearance or motion cues rather than explicitly modeling dynamics. Leon et al. \cite{leon2021review} differentiate model- and learning-based camera tracking, subdividing the latter into convolutional and recurrent models. Bashar et al. \cite{bashar2209multiple} propose a broader, non-exhaustive taxonomy of learning-based camera tracking that exploits rich semantic information, while Agrawal et al. \cite{agrawal2024systematic} refine learning-based MOT into nine architectural families. Hassan et al. \cite{hassan2024multi} provide a broader historical survey of both traditional and deep learning-based tracking, with continued emphasis on tracking-by-detection.

\begin{figure*}[!t]
\centering
\includegraphics[width=5in]{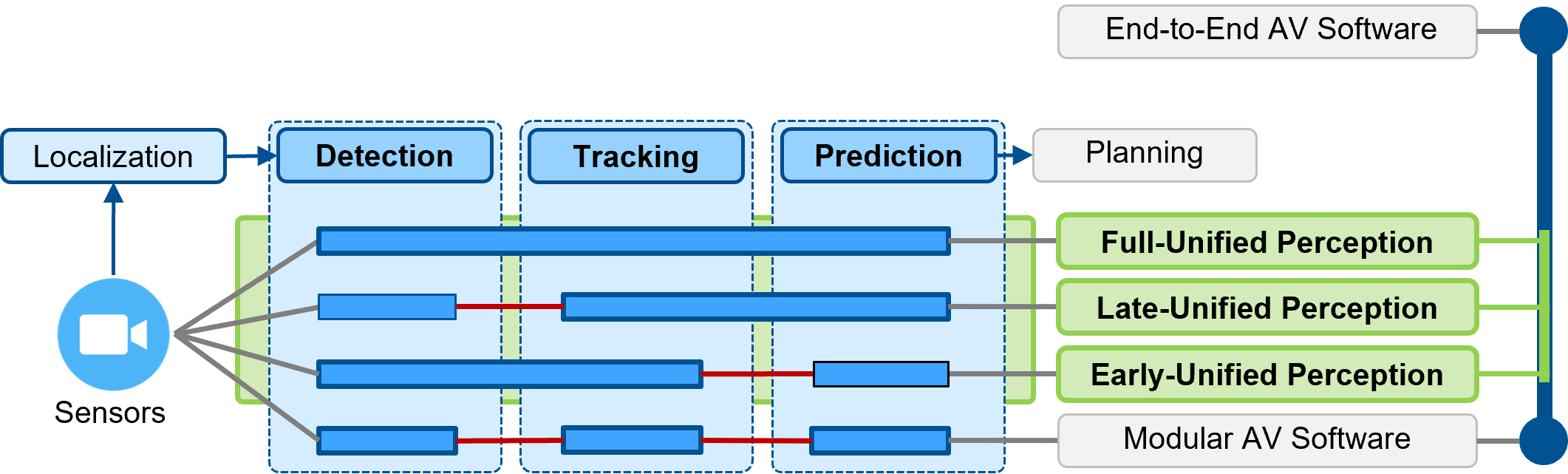}
\caption{Unified perception paradigms: Modular AV stacks perform perception sequentially. Through our taxonomy, we identify EUP, LUP and FUP, which combine detection, tracking, and prediction under different scopes into one joint module.}
\label{Unified_Perception_Overview}
\end{figure*}

\textbf{Motion Prediction} forecasts future object trajectories from tracked states and contextual cues such as agent interactions and road semantics, which strongly influence accuracy \cite{uhlemann2024evaluating}. Methods divide into physics- and learning-based approaches, the latter dominating. Physics-based prediction relies on handcrafted motion models. Karle et al. \cite{karle2022scenario} and Huang et al. \cite{huang2022survey} note these are interpretable and training-free but struggle to capture interactions and complex non-linear behavior. Learning-based methods, which dominate modern research, anticipate motion through supervised training. Huang et al. \cite{huang2022survey} group trajectory prediction by underlying technology (classic machine-, deep-, and reinforcement-learning models), Karle et al. \cite{karle2022scenario} recast the field into pattern- (learned motion patterns) and planning-based (interaction-optimized) methods, Ding et al. \cite{ding2023incorporating} distinguish knowledge-free from knowledge-based approaches by whether traffic priors are explicitly incorporated, and Huang et al. \cite{huang2023review} propose a taxonomy along scene representation, context refinement, and prediction rationality, also covering the emerging direction of occupancy prediction.

While modular perception offers well-defined interfaces and interpretability, its strict task separation limits global scene reasoning and inter-task synergy. Detection errors propagate through tracking and prediction, and useful contextual information is lost at module boundaries. In contrast, end-to-end AV stacks learn the driving task holistically, mapping raw sensor inputs directly to actions. Such approaches are on the rise \cite{chib2023recent} given accessible hardware and data, promising holistic reasoning and generalization across scenarios without intermediate representations or modules. Perception is thus learned implicitly within the system. However, end-to-end surveys \cite{tampuu2020survey, chib2023recent, singh2023end, chen2024end, al2024end} convey that these are highly complex to train, challenging to verify, and lack interpretability, limiting safety-critical deployment. These limitations motivate more balanced solutions to perception.

Unified perception is positioned within a broader spectrum of integration granularities:
\begin{itemize}
    \item \textbf{Modular:} localization, detection, tracking, and prediction are solved independently as separate modules.
    \item \textbf{Unified perception:} scene understanding (detection, tracking, prediction) is integrated into a single module.
    \item \textbf{End-to-end perception:} localization and scene understanding are unified into one stage, also termed first-stage end-to-end~\cite{phillips2021deep}.
    \item \textbf{End-to-end:} localization, scene understanding, and planning are learned jointly, mapping raw sensor data directly to planned trajectories.
\end{itemize}

\textbf{Unified perception} extends modular perception by restructuring sub-modules into larger, integrated components, leveraging modular principles while explicitly enabling global reasoning and inter-task synergy through a shared architecture and deep contextual understanding. Methods that jointly learn localization or planning are regarded as end-to-end perception or end-to-end driving, respectively, and are out of scope. Its key premise is that the scene-understanding sub-tasks (detection, tracking, prediction) are inherently interdependent. Inspired by the holistic learning of end-to-end systems, unified perception leverages shared representations and joint optimization, enhancing information flow while supporting robustness, generalizability, and interpretability.

To formalize this design space, we categorize unified perception into three types, illustrated in Fig.~\ref{Unified_Perception_Overview}, by the degree and stage of task integration:
\begin{itemize}
    \item Early-Unified Perception (EUP): integrates object detection and tracking into a single module. Unlike modular tracking-by-detection, EUP can explicitly retain feature information typically lost between separate detection and tracking stages.
    \item Late-Unified Perception (LUP): integrates tracking and prediction, enabling joint reasoning over an object's current state and future trajectory via shared temporal representations for more coherent, interaction-aware estimation.
    \item Full-Unified Perception (FUP): integrates scene understanding (detection, tracking, prediction) within a single module, the highest level of integration, solving perception as a holistic task and maximizing inter-task synergy and global scene understanding.
\end{itemize}

\begin{table}[h!]
    \centering
    \caption{Survey overview of modular, end-to-end, and unified perception approaches}
    \label{Survey_overview}
    \resizebox{\columnwidth}{!}
    {%
    \begin{tabular}{c c c c c}
        \toprule
        \multirow{2}{*}{\textbf{Type}} & \multicolumn{3}{|c|}{\textbf{Perception}} \\
        \cmidrule(lr){2-4}
        & \multicolumn{1}{|c|}{Detection} & \multicolumn{1}{c|}{Tracking} & \multicolumn{1}{c|}{Prediction} & \\
        \midrule
        \multicolumn{1}{c|}{Modular AV Software} & 
        \multicolumn{1}{c|}{\makecell{\cite{mao20233d}, \cite{song2024robustness}, \\ \cite{ma20233d}, \cite{liu2023echoes}, \\ \cite{wang2023multi}, \cite{yao2023radar}, \\ \cite{aung2024review}}} & 
        \multicolumn{1}{c|}{\makecell{\cite{leon2021review}, \cite{luo2021multiple}, \\ \cite{bashar2209multiple}, \cite{rakai2022data}, \\ \cite{agrawal2024systematic}, \cite{hassan2024multi}}} & 
        \multicolumn{1}{c|}{\makecell{\cite{huang2022survey}, \cite{karle2022scenario}, \\ \cite{ding2023incorporating}, \cite{huang2023review}}} & 
        \\ \cmidrule(lr){1-4}
        
        \multicolumn{1}{c|}{Early-Unified Perception} &  \multicolumn{2}{c|}{\makecell{\textbf{Ours}}} & \multicolumn{1}{c|}{} & \\ \cmidrule(lr){1-4}

        \multicolumn{1}{c|}{Late-Unified Perception} &  & \multicolumn{2}{|c|}{\textbf{Ours}} & \\ \cmidrule(lr){1-4}
        
        \multicolumn{1}{c|}{Full-Unified Perception} & \multicolumn{3}{c|}{\makecell{\cite{dal2024joint}, \textbf{Ours}}} & \\ \midrule
        
        \multicolumn{1}{c|}{End-to-End AV Stacks} & \multicolumn{4}{c}{\makecell{\cite{tampuu2020survey}, \cite{chib2023recent}, \cite{singh2023end}, \cite{chen2024end}, \cite{al2024end}}} \\
        \bottomrule
    \end{tabular}
    }
\end{table}

As summarized in Tab.~\ref{Survey_overview}, the current landscape reflects no holistic coverage of unified perception, a gap this survey closes. Early-Unified Perception (EUP), or joint detection and tracking, is often adopted but has not been explicitly surveyed as a distinct paradigm, despite acknowledgment in recent works such as \cite{agrawal2024systematic, hassan2024multi}. Late-Unified Perception (LUP) remains unaddressed in literature. Full-Unified Perception (FUP) has recently been surveyed by Dal'Col et al.~\cite{dal2024joint} under the term joint detection and prediction, which we refine. Thus, this survey provides a holistic and structured overview of unified perception, with a taxonomy generally applicable across all unified perception approaches that clarifies existing methods and highlights paradigms within this design space.

\section{Unified Perception Taxonomy}

\label{chap:UnifiedPerceptionTaxonomy}

\begin{figure}[h]
\centering
\includegraphics[width=\linewidth]{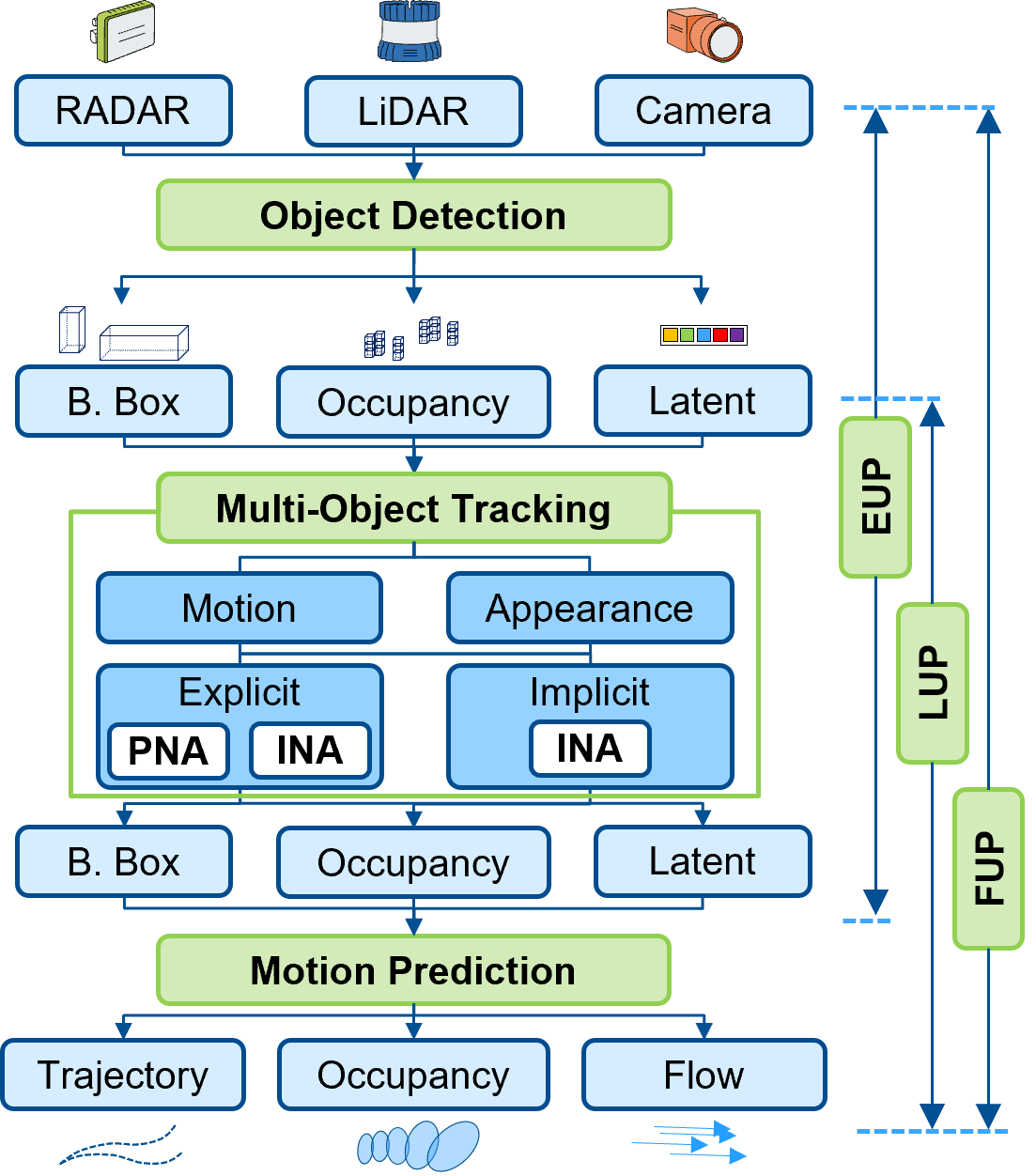}
\caption{Overview of proposed taxonomy with in- and outputs.}
\label{Taxonomy}
\end{figure}

Unified perception, comprising the three paradigms EUP, LUP, and FUP, can be systematically described and analyzed using the holistic taxonomy introduced in Figure~\ref{Taxonomy}. This taxonomy offers a comprehensive framework for understanding how detection, tracking, and prediction tasks are unified from a system perspective. It characterizes unified perception systems based on information flow, system boundaries (input/output), intermediate representations, and integration strategies, without constraining architectural choices.

Any component originally developed under the modular paradigm (Sec.\ref{chap:StateOfTheArt}) can, in principle, be integrated into unified perception, provided it supports joint training. These architectural details are decoupled from the taxonomy itself and discussed in Sec.\ref{chap:Discussion}. Importantly, the central design element across all unified perception paradigms is tracking, which serves as the intermediate stage connecting object detection and motion prediction. Consequently, categorization is centered around tracking in our taxonomy.

We categorize unified perception systems along two dimensions: (1) the nature of cues used and (2) the tracking formulation, distinguishing between explicit and implicit tracking. Thereby, tracking cues (1) fall into two categories. Appearance-based cues describe features often derived from visual characteristics of objects, such as shape, texture, or learned embeddings. Motion cues, in contrast, rely on spatio-temporal properties like velocity, direction, or kinematic constraints. Systems may combine appearance and motion cues to improve robustness.

The second dimension (2) distinguishes between explicit and implicit tracking. Explicit tracking assigns persistent identities to objects across frames, creating temporally consistent tracks via hard, one-to-one associations. Thereby, two dominant mechanisms exist within explicit tracking: post-network association (PNA) and in-network association (INA), as introduced by Cai et al.~\cite{cai2022memot}. 

In PNA, identity assignment is decoupled from the model's inference process. The model produces intermediate outputs such as object features, embeddings, or motion cues, but the actual linking of identities across time is performed in a separate post-processing step. This association typically relies on external heuristics, optimization algorithms, or rule-based matching strategies. Although the model may be trained with temporal supervision, identity persistence is not handled within the network. In contrast, INA, integrates identity tracking directly into the model architecture and training process. The system jointly reasons over temporally extended inputs and assigns consistent identities in a single inference pass. Identity persistence is treated as a learnable function of the model, embedded into its representation and output space, eliminating the need for external association logic. 

Implicit tracking, on the other hand, avoids persistent identity assignment altogether. Instead, it maintains temporal continuity within the model’s latent representations, enabling soft associations across time, making it a form of INA. This allows the system to maintain temporal context and reason about object dynamics, rather than committing to deterministic identity assignments. While this increases flexibility and supports reasoning over multiple hypotheses, it comes at the cost of losing explicit, interpretable tracks.

A key aspect of our taxonomy is the flexible combination of input types, intermediate representations, and output formats within unified perception. Inputs to EUP and FUP consist of raw sensor data such as RADAR, LiDAR, or camera images. Between unified tasks, intermediate representations —bounding boxes, occupancy grids, or latent features— are produced and can be freely combined or propagated. For LUP, the system operates on intermediate representations generated after detection as input. This flexibility allows any combination of inputs, architecture, intermediate formats, and outputs, depending on system design. Outputs are thereby possible in the form of trajectory, occupancy, or flow representations. Critically, the representation type plays a crucial role in information sharing across sub-tasks, directly affecting explainability, interconnectivity (Fig.~\ref{Modular_vs_E2E}), and likely system performance.

In summary, this taxonomy presents a general and systemic framework for describing unified perception systems. It defines a structured design space based on task integration, tracking formulation, and representation flow. It supports both retrospective classification of existing methods and prospective design of new approaches. Following, we survey EUP, LUP, and FUP methods in detail.

\section{Early-Unified Perception}
\label{chap:EarlyUnifiedPerception}

Joint detection and tracking methods, summarized into Early-Unified Perception (EUP), integrate detection and multi-object tracking into a single, jointly trained module, thus the earliest point in the scene understanding pipeline at which modular sub-tasks can be coupled (Sec.~\ref{chap:StateOfTheArt}). 
Within our taxonomy (Sec.~\ref{chap:UnifiedPerceptionTaxonomy}), EUP occupies a distinctive position: tracking, the central element around which the taxonomy is organized, here constitutes the system output rather than an intermediate stage. This shapes EUP's design space directly, where identity assignment must resolve to discrete, persistent object tracks. This is why, unlike LUP and FUP, where tracking is forward connected with prediction, EUP relies exclusively on explicit tracking, realized through both PNA and INA mechanisms. Implicit tracking and occupancy representations, while permitted by the taxonomy, have therefore not yet been proposed in the EUP context. Soft or distributional identity cues offer no advantage when persistent tracks are themselves the desired output.

\begin{figure}[h]
\centering
\includegraphics[width=0.9\linewidth]{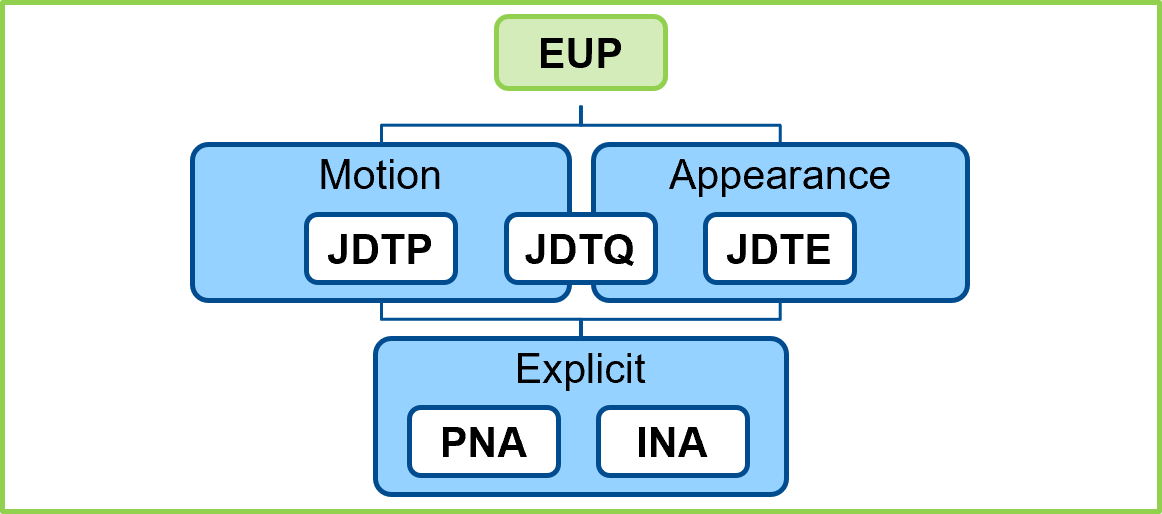}
\caption{The EUP paradigm with its three sub-paradigms: joint detection and tracking by propagation (JDTP), joint detection and tracking by embedding (JDTE), and joint detection and tracking by query (JDTQ). EUP follows an explicit tracking scheme leveraging either PNA or INA, due to the output requirements of tracking.}
\label{Early_Unified_Perception_Taxonomy}
\end{figure}

Following the taxonomy, we identify three EUP sub-paradigms, each defined by the mechanism used to couple detection and tracking and the cue that mechanism primarily exploits: Joint Detection and Tracking by Propagation (JDTP), which predicts object displacement from motion cues to associate instances over time. Joint Detection and Tracking by Embedding (JDTE), which augments detection with learned appearance embeddings and associates objects by embedding similarity, and Joint Detection and Tracking by Query (JDTQ), which treats detection and tracking as a single set-prediction problem, propagating learnable queries that represent, localize, and identify object instances across frames. JDTP and JDTE are each organized around a single dominant cue, though individual methods may incorporate secondary cues, whereas JDTQ bundles position, motion, and appearance into one query representation and therefore spans both. These mechanisms are orthogonal to the representation and association axes of our taxonomy. Methods within a sub-paradigm vary in their intermediate representation and in whether association is resolved via PNA or INA, as detailed below. A visual summary of how these sub-paradigms align with our taxonomy is presented in Fig.~\ref{Early_Unified_Perception_Taxonomy}.

An overview of all EUP methods surveyed is summarized in Tab.~\ref{EUP_table}. On top of the categorization used below, additional key attributes such as number of input timesteps, network architecture, learning strategy, datasets, and code availability are presented. This structure enables readers to understand both the conceptual differences and practical implementations of EUP approaches across the literature.

\setcounter{footnote}{0}
\begin{table*}[!t]
    \centering
    \caption{Early-Unified Perception Approaches: Overview}
    \label{EUP_table}
    \resizebox{\textwidth}{!}
    {%
    \begin{tabular}{l c c c c c c c c r}
        \toprule
        \textbf{Publ.}& \textbf{Year} & \textbf{Modality}\footnotemark & \textbf{Input}\footnotemark & \textbf{Paradigm}\footnotemark & \textbf{Output}\footnotemark & \textbf{Architecture} & \textbf{Learning Strategy}\footnotemark & \textbf{Dataset}\footnotemark & \textbf{Code} \\
        \midrule

        CenterTrack\cite{zhou2020tracking}         & 2020 & I   & 2     & JDTP - PNA  & 2/3D BB  & CenterNet-backbone\cite{zhou2019objects} + Offset Prediction                                                  & Emp. W.L.     & kitti, nuScenes, MOT17                    & \cmark \\ 
        TubeTK\cite{pang2020tubetk}                & 2020 & I   & 8     & JDTP - INA  & 2D BB    & 3D-ResNet\cite{hara2018can} + 3D-FPN + CNN Heads                                                              & Unb. W.L.      & MOT15/16/17                               & \cmark \\
        PermaTrack\cite{tokmakov2021learning}      & 2021 & I   & 1+M   & JDTP - INA  & 2D BB    & CenterTrack\cite{zhou2020tracking} + Spatio-temporal Recurrent Memory Module                                  & Emp. W.L.     & kitti, MOT17                              & \cmark \\
        P3AFormer\cite{zhao2022tracking}           & 2022 & I   & 2     & JDTP - INA  & 2D BB    & ResNet-50\cite{he2016deep} + Swin-Transformer E. + DETR\cite{zhu2020deformable} D. + FlowNet\cite{zhu2017flow}& Emp. W.L.     & kitti, MOT17/20                           & \cmark \\
        QD3DT\cite{hu2022monocular}                & 2021 & I   & 1+M   & JDTP - PNA  & 3D BB    & Faster-RCNN-backbone\cite{ren2015faster} + Quasi-dense Similarity Learning                                    & Emp. W.L.     & kitti, nuScenes, WOD                      & \cmark \\
        STAR-Track\cite{doll2023star}              & 2024 & MVI & 1+M.  & JDTP/Q - INA& 3D BB    & DETR3D\cite{wang2022detr3d} + Latent Motion Model + Learnable Track Embedding                                 & -             & nuScenes                                  & \xmark \\
        CenterPoint\cite{yin2021center}            & 2021 & L   & 1+M   & JDTP - PNA  & 3D BB    & VoxelNet\cite{zhou2018voxelnet}/ PointPillars-backbone\cite{lang2019pointpillars} + CNN + Association         & -             & nuScenes, WOD                             & \cmark \\
        SimTrack\cite{luo2021exploring}            & 2021 & L   & 2     & JDTP - INA  & 3D BB    & VoxelNet\cite{zhou2018voxelnet}/ PointPillars-backbone\cite{lang2019pointpillars} + Hybrid-Time Center/ Movement Branch& Emp. W.L.     & nuScenes, WOD                & \cmark \\
        CenterTube\cite{liu2023centertube}         & 2023 & L   & 5     & JDTP - INA  & 3D BB    & VoxelNet\cite{zhou2018voxelnet}/ PointPillars-backbone\cite{lang2019pointpillars} + Center/ Regression/ Movement Branch& Emp. W.L.     & nuScenes, kitti                  & \xmark \\
        
        VoxelNeXt\cite{chen2023voxelnext}          & 2023 & L   & 1+M   & JDTP - INA  & 3D BB    & 3D Sparse CNN Backbone + Heads + Query Voxel Association                                                      & -             & nuScenes, WOD, Argoverse2                 & \cmark \\
        MMF-JDT\cite{wang2024multi}                & 2024 &I+L  & 1+M   & JDTP - INA  & 3D BB    & 2/3D RPN + 2/3D Trajectory Regression and Refinement + Fusion                                                 & -             & kitti, WOD                                & \cmark \\
        \midrule

        JDE\cite{wang2020towards}                  & 2020 & I   & 1+M   & JDTE - PNA   & 2D BB    & DarkNet-53\cite{redmon2018yolov3} + Prediction Heads                                                          & Unc. W.L.     & MOT16                                     & \cmark \\
        RetinaTrack\cite{lu2020retinatrack}        & 2020 & I   & 1+M   & JDTE - PNA   & 2D BB    & FPN + Task-shared Post-FPN Layers                                                                             & Unb. W.L.     & WOD, MOT17                                & \cmark \\
        Siamese Track-RCNN\cite{shuai2020multi}    & 2020 & I   & 1+M   & JDTE - INA   & 2D BB    & ResNet-101\cite{he2016deep} + FPN + Deformable Convolution                                                    & Unb. W.L.     & MOT16/17, JTA                             & \xmark \\
        CTracker\cite{peng2020chained}             & 2020 & I   & 2     & JDTE - INA   & 2D BB    & ResNet-50\cite{he2016deep} + FPN + Joint Attention                                                            & Emp. W.L.     & MOT16/17                                  & \cmark \\
        DEFT\cite{chaabane2021deft}                & 2021 & I   & 1+M   & JDTE - PNA   & 2/3D BB    & CenterNet-backbone\cite{zhou2019objects} + Embedding Extractor                                                & Unc. W.L.     & kitti, nuScenes, MOT17                    & \cmark \\
        FairMOT\cite{zhang2021fairmot}             & 2021 & I   & 1+M   & JDTE - PNA   & 2D BB    & CenterNet-backbone\cite{zhou2019objects} + Re-ID Branch                                                       & Unc. W.L.     & MOT15/16/17/20                            & \cmark \\
        GSDT\cite{wang2021joint}                   & 2021 & I   & 2     & JDTE - PNA   & 2D BB    & CenterNet-backbone\cite{zhou2019objects} + GNN                                                                & Unc. W.L.     & MOT15/16/17/20                            & \cmark \\
        CorrTracker\cite{wang2021multiple}         & 2021 & I   & 4     & JDTE - PNA   & 2D BB    & CenterNet-backbone\cite{zhou2019objects} + Correlation Learning in Pyramid                                    & Unb. W.L.      & MOT15/16/17/20                            & \xmark \\
        AlignPS\cite{yan2021anchor}                & 2021 & I   & 1+M   & JDTE - PNA   & 2D BB    & FCOS\cite{tian2019fcos} + Feature Aggregation + Re-ID Embedding                                               & Unb. W.L.     & CUHK-SYSU, PRW                            & \cmark \\
        Time3D\cite{li2022time3d}                  & 2022 & I   & 1+M   & JDTE - PNA   & 2/3D BB  & KM3D\cite{li2021monocular} + Attention-based Re-ID Embedding                                                  & Unb. W.L.      & nuScenes                                  & \xmark \\
        TraDeS\cite{wu2021track}                   & 2021 & I   & 3     & JDTE/P - INA & 2/3D BB  & CenterNet-backbone\cite{zhou2019objects}+ Re-ID Embedding + Motion Propagation                                & Unb. W.L.     & nuScenes, MOT16/17, MOTS, Youtube-VIS     & \cmark \\
        Minkowski Tracker\cite{gwak2022minkowski}  & 2022 & L   & 3     & JDTE/P - PNA & 3D BB    & SECOND-based\cite{yan2018second} +spatio-temporal R-CNN                                                        & Emp. W.L.     & nuScenes                                  & \xmark \\
        JMODT\cite{huang2021joint}                 & 2021 &I+L  & 1+M   & JDTE/P - PNA & 3D BB    & RPN + Correlation Backbone + Association                                                                      & Emp. W.L.     & kitti                                     & \cmark \\
        AlphaTrack\cite{zeng2021cross}             & 2021 &MVI+L  & 1+M   & JDTE/P - PNA & 3D BB    & CenterPoint-based \cite{yin2021center} Branch + Embedding + Affinity                                          & Emp. W.L.     & nuScenes                                  & \xmark \\
        3D DetecTrack\cite{koh2022joint}           & 2022 &I+L  &  2    & JDTE - PNA   & 3D BB    & 3D-CVF-backbone\cite{yoo20203d} + Spatio-temporal Gated GNN                                                   & Unb. W.L.     & kitti, nuScenes                           & \xmark \\ 
        \midrule
        
        TransTrack\cite{sun2020transtrack}         & 2020 & I   & 1+M   & JDTQ - PNA  & 2D BB    & DETR-backbone\cite{zhu2020deformable} + Transformer E.-D. + Matching                                          & Emp. W.L.     & MOT17/20                                  & \cmark \\
        TrackFormer\cite{meinhardt2022trackformer} & 2022 & I   & 1+M   & JDTQ - INA  & 2D BB    & DETR-backbone\cite{zhu2020deformable} + Transformer E.-D.                                                     & Emp. W.L.     & MOT17, MOTS20                             & \cmark \\
        TransCenter\cite{xu2022transcenter}        & 2022 & I   & 2     & JDTQ - INA  & 2D BB    & DETR\cite{carion2020end}/ PVT\cite{wang2022pvt} E. + Query Learning Network + Transformer D.                  & Emp. W.L.     & kitti, MOT17/20                           & \cmark \\
        MOTR\cite{zeng2022motr}                    & 2022 & I   & 1+M   & JDTQ - INA  & 2D BB    & DETR-backbone\cite{zhu2020deformable} + Transformer E.-D.                                                     & Emp. W.L.     & MOT17/20                                  & \cmark \\
        MeMOT\cite{cai2022memot}                   & 2022 & I   & 1+M   & JDTQ/E - INA& 2D BB    & DETR-backbone\cite{zhu2020deformable} + Transformer Memory Bank                                               & Emp. W.L.     & MOT16/17/20                               & \xmark \\
        MOTRv2\cite{zhang2023motrv2}               & 2023 & I   & 1+M   & JDTQ - INA  & 2D BB    & YOLOX\cite{ge2021yolox} proposals + MOTR\cite{zeng2022motr}                                                   & Emp. W.L.     & MOT17/20, DanceTrack                      & \cmark \\
        MOTRv3\cite{yu2023motrv3}                  & 2023 & I   & 1+M   & JDTQ - INA  & 2D BB    & YOLOX\cite{ge2021yolox} proposals + MOTR\cite{zeng2022motr}  + Release-Fetch-Supervision                      & Emp. W.L.     & MOT17, DanceTrack                         & \xmark \\
        Co-MOT\cite{yan2023bridging}               & 2023 & I   & 1+M   & JDTQ - INA  & 2D BB    & ResNet-50\cite{he2016deep} + Deformable Transformer E. + Stacked Deformable D.                                & Unb. W.L.      & MOT17, DanceTrack, BDD100K                & \cmark \\
        MeMOTR\cite{gao2023memotr}                 & 2023 & I   & 1+M   & JDTQ - INA  & 2D BB    & ResNet-50\cite{he2016deep} + Transformer E.-D. + Temporal Interaction Module                                  & Emp. W.L.     & MOT17, DanceTrack, BDD100K                & \cmark \\
        LAID\cite{jia2024multi}                    & 2024 & I   & 1+M   & JDTQ - INA  & 2D BB    & YOLOX-backbone\cite{ge2021yolox} + Learnable Associator                                                       & Emp. W.L.     & DanceTrack, SportsMOT                     & \xmark \\
        SambaMOTR\cite{segu2024samba}              & 2024 & I   & 1+M   & JDTQ - INA  & 2D BB    & MOTR\cite{zeng2022motr} E.-D. + Set-of-sequences Model                                                        & Emp. W.L.     & MOT17, DanceTrack, SportsMOT, BFT         & \cmark \\
        MUTR3D\cite{zhang2022mutr3d}               & 2022 & MVI & 1+M   & JDTQ - INA  & 3D BB    & DETR3D\cite{wang2022detr3d} Multi-camera Transformer E.-D.                                             & Emp. W.L.     & nuScenes                                  & \cmark \\
        DQTrack\cite{li2023end}                    & 2023 & MVI & 1+M   & JDTQ - INA  & 3D BB    & DETR3D\cite{wang2022detr3d} E. + Transformer-based Learnable Association                                      & Emp. W.L.     & nuScenes                                  & \cmark \\
        Ada-track\cite{ding2024ada}                & 2024 & MVI & 3     & JDTQ - INA  & 3D BB    & DETR3D-backbone\cite{wang2022detr3d} + Transformer D. + Learnable Association                                 & Emp. W.L.     & nuScenes                                  & \cmark \\
        OneTrack\cite{wang2024onetrack}            & 2024 & MVI & 1+M   & JDTQ - INA  & 3D BB    & StreamPetr-E.\cite{wang2023exploring} + Transformer D. + Masked Attention                                     & Emp. W.L.     & nuScenes                                  & \xmark \\
        MCTR\cite{niculescu2025mctr}               & 2025 & MVI & 1+M   & JDTQ - INA  & 2D BB    & DETR\cite{zhu2020deformable} Embeddings + Transformer D. + Association Module                                 & Unb. W.L.     & MMPTrack                                  & \xmark \\
        JDT3D\cite{cheong2024jdt3d}                & 2024 & L   & 1+M   & JDTQ/P - INA& 3D BB    & VoxelNet\cite{zhou2018voxelnet} + Transformer D. + Past/Future Reasoning                                      & Emp. W.L.     & nuScenes                                  & \cmark \\
        Mask4Former\cite{yilmaz2024mask4former}    & 2024 & L   & 2     & JDTQ - INA  & 3D Seg.   & DETR-inspired\cite{carion2020end} + Mask Module + Query Refinement                                            & Emp. W.L.     & kitti(semanticKITTI)                      & \cmark \\
        MotionTrack\cite{qin2023motiontrack}       & 2023 &MVI+L& 1+M   & JDTQ - INA  & 3D BB    & TransFusion\cite{bai2022transfusion} + Transformer-based Data Association                                     & -             & MOT17/20                                  & \xmark \\
        \bottomrule
        
        \addlinespace[4pt]
       
        \multicolumn{10}{l}{\footnotesize
                            \parbox{29cm}{
                                            \hangindent=4.5em 
                                            \textsuperscript{6} Dataset: (kitti:\cite{geiger2013vision}, nuScenes:\cite{caesar2020nuscenes}, WOD:\cite{sun2020scalability}, Argoverse2:\cite{Argoverse2}, MOT15:\cite{MOTChallenge2015}, MOT16:\cite{MOT16}, MOT17:\cite{MOT16}, MOT20:\cite{MOTChallenge20}, MOTS:\cite{MOTS20}, CUHK-SYSU:\cite{xiao2017joint}, PRW:\cite{zheng2017person}, DanceTrack:\cite{sun2022dancetrack}, JTA:\cite{fabbri2018learning}, BDD100K:\cite{yu2020bdd100k}, SportsMOT:\cite{cui2023sportsmot}, BFT:\cite{zheng2024nettrack}, Youtube-VIS:\cite{yang2019video}, MMPTrack:\cite{han2023mmptrack})%
                                         }%
                           } \\
        \multicolumn{10}{l}{\footnotesize \textsuperscript{5} Learning Strategy: (Emp. W.L.: Empirical Weighted Learning, Unb. W.L.: Unbalanced Weighted Learning, Unc. W.L.: Uncertainty Weighted Learning \cite{kendall2018multi}, -: No Information)} \\
        \multicolumn{10}{l}{\footnotesize \textsuperscript{4} Output: Dimension (2D/ 3D), BB: Bounding Box, Seg.: Segmentation } \\
        \multicolumn{10}{l}{\footnotesize \textsuperscript{3} Paradigm: ([JDTP: Joint Detection and Tracking by Propagation, JDTE: Joint Detection and Tracking by Embedding, JDTQ: Joint Detection and Tracking by Query], [PNA: Post-Network Association, INA: In-Network Association])} \\
        \multicolumn{10}{l}{\footnotesize \textsuperscript{2} Input: ([Number of Frames Input, M: Use of Memory]) } \\
        \multicolumn{10}{l}{\footnotesize \textsuperscript{1} Modality: (I: Image, MVI: Multi-View Image, L: LiDAR)} \\
    
    \end{tabular}
    }
\end{table*}

\subsection{Joint Detection and Tracking by Propagation}

JDTP unifies detection and tracking by leveraging motion cues, thus the spatial displacement of objects across frames, making it the unified counterpart to modular model-based tracking-by-detection. Its reliance on explicit spatial relationships makes JDTP particularly well suited to sensors with high spatial accuracy, such as LiDAR, and it is realized through both PNA- and INA-based designs. \\ \indent
PNA-based JDTP approaches extend established detectors with lightweight motion-estimation heads while resolving identities externally. CenterTrack \cite{zhou2020tracking} augments the keypoint-based detector CenterNet \cite{zhou2019objects} with offset-prediction heads that estimate 2D center displacements between consecutive frames, enabling differentiable, greedy frame-to-frame matching without external optimization. QD3DT \cite{hu2022monocular} extends this principle to 3D, combining a Faster-RCNN detector \cite{ren2015faster} with a motion-aware LSTM and depth-ordering heuristics, though identity assignment still relies on a handcrafted affinity matrix and greedy matching. CenterPoint \cite{yin2021center}, in turn, regresses object velocities directly from temporal point-cloud data through an additional head, associating predicted centers post hoc via spatial proximity.\\ \indent
INA-based JDTP methods instead embed identity association within the network via motion-aware heads, memory modules, or query propagation. TubeTK \cite{pang2020tubetk} reformulates tracking as tube prediction, regressing short bounding-box sequences in a single forward pass, although longer-term association still requires PNA-style tube-level non-maximum suppression. PermaTrack \cite{tokmakov2021learning} extends CenterTrack with ConvGRU-based spatio-temporal memory, enabling recovery from occlusions within an end-to-end trainable framework. Transformer-based designs push this further: P3AFormer \cite{zhao2022tracking} propagates object queries as persistent identity carriers via optical-flow-guided attention, while STAR-Track \cite{doll2023star} encodes dynamic object state into a Latent Motion Model, fully integrating identity management into the attention pipeline without post hoc logic. These INA designs are predominantly camera-based. The same in-network association principle extends naturally to LiDAR.\ \\ \indent
For LiDAR inputs, SimTrack \cite{luo2021exploring} embeds temporal logic directly into detection heads via hybrid-time centerness maps and motion offsets, CenterTube \cite{liu2023centertube} generalizes TubeTK's tube regression to 3D/4D space for spatio-temporal LiDAR linking, and VoxelNeXt \cite{chen2023voxelnext} extends CenterPoint with sparse voxel feature propagation, allowing identity to emerge as a property of persistent spatial features rather than explicit matching. Lastly, the multi-modal MMF-JDT \cite{wang2024multi} fuses image and LiDAR cues, integrating past tracklet features into fusion-based detection heads and resolving associations through shared in-network attention and NMS modules.

\subsection{Joint Detection and Tracking by Embedding}

JDTE unifies detection and tracking by learning appearance embeddings within the network, training the model to distinguish similar from dissimilar objects via contrastive or association-based losses on propagated latent representations, without requiring explicit labels for the embeddings themselves. \\ \indent
PNA-based JDTE methods predict detections and embeddings jointly but resolve identities externally using standard data-association techniques \cite{rakai2022data}. JDE \cite{wang2020towards} and RetinaTrack \cite{lu2020retinatrack} extract detections and embeddings from a shared FPN or RetinaNet backbone, associating them post hoc via cost matrices combining cosine appearance and Mahalanobis motion distances, solved with the Hungarian algorithm \cite{kuhn1955hungarian} or via greedy bipartite matching. FairMOT \cite{zhang2021fairmot} similarly shares a backbone for detection and embedding but applies hierarchical association combining Kalman filtering with Hungarian matching, while DEFT \cite{chaabane2021deft} adds an LSTM-based motion-forecasting module to refine the affinities used for matching. CorrTracker \cite{wang2021multiple} and AlignPS \cite{yan2021anchor} focus on learning more robust correlation- and anchor-free embeddings respectively, but likewise resolve associations externally, as do the graph-based GSDT \cite{wang2021joint} and 3D DetecTrack \cite{koh2022joint}, which refine appearance and motion cues internally yet still match affinities combinatorially post-network. This pattern extends to the multi-modal Time3D \cite{li2022time3d}, JMODT \cite{huang2021joint}, AlphaTrack \cite{zeng2021cross}, and Minkowski Tracker \cite{gwak2022minkowski}, which learn spatio-temporal embeddings or affinity scores internally but still perform identity association as a discrete post-processing step via greedy, Hungarian, or mixed-integer matching strategies. \\ \indent
INA-based JDTE methods instead embed association directly into the inference pipeline, enabling simultaneous prediction of detections and identities without external solvers. Siamese Track-RCNN \cite{shuai2020multi} extends Faster R-CNN \cite{he2016deep} with dedicated tracking and re-identification branches that propagate tracklets and output identity associations within a single forward pass. CTracker \cite{peng2020chained} regresses paired bounding boxes from a dual-branch, shared-weight backbone conditioned on classification and identity-verification attention maps, performing implicit frame-pair association end-to-end. TraDeS \cite{wu2021track} computes re-identification similarity through a cost-volume-based association module, using the resulting tracking offsets to guide a motion-aware feature warper that jointly infers detections and associations without a separate data-association stage.

\subsection{Joint Detection and Tracking by Query}

JDTQ unifies detection and tracking by formulating both as a single set-prediction problem, representing and propagating object instances through learnable query embeddings updated via attention-based mechanisms across frames. This is an inherently INA formulation, as association is performed within the network. Queries are typically split into detection queries for discovering new objects and track queries for propagating known identities, interacting with encoded input features via cross-attention to retrieve, localize, and maintain object representations over time. Because a single query bundles position, motion, and appearance into one high-dimensional representation, JDTQ implicitly includes both motion and appearance cues. Some methods also explicitly model one of these cues on top of the query (denoted JDTQ/P or JDTQ/E in Tab.~\ref{EUP_table}). This is a deliberate, complementary design choice, not an ambiguity in the taxonomy: the query already carries the cue implicitly, and the explicit head reinforces it. 

Early image-based methods established the core mechanisms: TransTrack \cite{sun2020transtrack} fuses detection and track queries via concatenation and late fusion, though residual post-decoding IoU matching makes it partially PNA, while TrackFormer \cite{meinhardt2022trackformer} propagates track queries through decoder memory without external association heuristics. TransCenter \cite{xu2022transcenter} replaces sparse, content-independent queries with dense, image-derived ones for greater robustness to occlusion and small objects, and MOTR \cite{zeng2022motr} extends DETR to video by updating track queries frame-by-frame through decoder interactions, achieving fully integrated detection and association. Subsequent work targets long-term robustness and training stability: MeMOT \cite{cai2022memot} introduces spatio-temporal memory for similarity-based retrieval beyond strict query propagation, MOTRv2 \cite{zhang2023motrv2} and MOTRv3 \cite{yu2023motrv3} refine label-assignment mechanisms and the balance between detection and tracking supervision, Co-MOT \cite{yan2023bridging} promotes cooperative learning between detect and track queries, MeMOTR \cite{gao2023memotr} extends temporal context through long-term memory, LAID \cite{jia2024multi} balances accuracy and training efficiency via pretrained-detector alignment, and SambaMOTR \cite{segu2024samba} improves robustness to occlusion through synchronized memory propagation. \\ \indent
In multi-camera settings, MUTR3D \cite{zhang2022mutr3d} pioneers view-aware deformable attention for cross-view consistency, DQTrack \cite{li2023end} decouples detection and tracking queries with visibility-guided temporal propagation, Ada-track \cite{ding2024ada} adaptively selects queries by visibility and confidence to reduce redundant computation and false positives, OneTrack \cite{wang2024onetrack} fuses queries across views and time for end-to-end spatio-temporal reasoning without post-processing, and MCTR \cite{niculescu2025mctr} adds contrastive query refinement and explicit cross-view temporal correlation for robust identity preservation in crowded, occluded scenes. \\ \indent
For LiDAR, JDT3D \cite{cheong2024jdt3d} adapts detection and track queries to voxelized 3D features via temporal deformable attention, while Mask4Former \cite{yilmaz2024mask4former} introduces mask-based queries that directly yield spatio-temporal segmentation masks for precise localization and association, marking a shift toward dense tracking representations in 3D scenes. Finally, the multi-modal MotionTrack \cite{qin2023motiontrack} fuses vision and point-cloud features through modality-specific cross-attention within joint queries, demonstrating JDTQ's extensibility to heterogeneous sensor combinations.

\subsection{Quantitative EUP Analysis}
Table~\ref{EUP_nuscenes_table} presents the AMOTA and AMOTP metrics as reported by the original researchers on nuScenes~\cite{caesar2020nuscenes}, specifically for the EUP methods where these tracking-specific metrics were made available. Due to variations in modality, backbone detectors, and training configurations, the entries serve to illustrate the reported findings rather than facilitate a controlled comparison. All entries utilizing LiDAR technology exhibit higher AMOTA values compared to those based solely on camera input, with VoxelNeXt~\cite{chen2023voxelnext} achieving the highest recorded value, likely due to the geometric accuracy inherent in point clouds. Given the strong correlation between modality and sub-paradigm, no ranking among sub-paradigms is provided. The highest value stems from a JDTP design, followed closely by JDTE/P entries, while the JDTQ/P method JDT3D still surpasses all camera-only entries. Among the camera-only methods, the JDTQ methods OneTrack and DQTrack showcase the highest reported metrics.

\begin{table}[!t]
    \centering
    \caption{Early-Unified Perception Approaches: nuScenes Test Set Performance, Original Reports}
    \label{EUP_nuscenes_table}
    \begin{tabular}{l c c c c}
        \toprule
        \textbf{Publication} & \textbf{Year} & \textbf{Modality}\textsuperscript{1} & \textbf{AMOTA}$\uparrow$ & \textbf{AMOTP}$\downarrow$ \\
        \midrule

        CenterTrack\cite{zhou2020tracking}        & 2020 & I     & 0.040 & 1.540 \\
        QD3DT\cite{hu2022monocular}               & 2021 & I     & 0.217                       & 1.550 \\
        STAR-Track\cite{doll2023star}             & 2024 & MVI   & 0.430                       & -     \\
        CenterPoint\cite{yin2021center}           & 2021 & L     & 0.630                       & -     \\
        SimTrack\cite{luo2021exploring}           & 2021 & L     & 0.640                       & -     \\
        VoxelNeXt\cite{chen2023voxelnext}         & 2023 & L     & 0.710                       & 0.511 \\
        \midrule

        DEFT\cite{chaabane2021deft}               & 2021 & I     & 0.170\textsuperscript{*}    & -     \\
        Time3D\cite{li2022time3d}                 & 2022 & I     & 0.210                       & 1.360 \\
        Minkowski Tracker\cite{gwak2022minkowski} & 2022 & L     & 0.698                       & -     \\
        AlphaTrack\cite{zeng2021cross}            & 2021 & MVI+L & 0.704                       & -     \\
        \midrule

        MUTR3D\cite{zhang2022mutr3d}              & 2022 & MVI   & 0.270                       & 1.494 \\
        DQTrack\cite{li2023end}                   & 2023 & MVI   & 0.524                       & 1.096 \\
        Ada-track\cite{ding2024ada}               & 2024 & MVI   & 0.479                       & 1.246 \\
        OneTrack\cite{wang2024onetrack}           & 2024 & MVI   & 0.554                       & 1.021 \\
        JDT3D\cite{cheong2024jdt3d}               & 2024 & L     & 0.574                       & 0.837 \\
        \bottomrule

        \addlinespace[4pt]
        \multicolumn{5}{l}{\footnotesize \textsuperscript{1} Modality: (I: Image, MVI: Multi-View Image, L: LiDAR)} \\
        \multicolumn{5}{l}{\footnotesize \textsuperscript{*} Test set not explicitly stated by authors.} \\

    \end{tabular}
\end{table}

\subsection{Comparative EUP Analysis}

Across the EUP landscape, several cross-cutting trends and trade-offs emerge that shape both the state of the art and open research opportunities.

\textbf{From geometric cues to learned representations:} The three sub-paradigms trace a representational trajectory: JDTP grounds association in explicit, geometrically interpretable motion cues (offsets, velocities, tubes). JDTE shifts toward learned but comparatively low-dimensional appearance embeddings, and JDTQ generalizes both into high-dimensional query representations jointly encoding position, appearance, and identity. This mirrors the broader shift discussed in Sec.~\ref{subsec:Intermediate Representations} from bounding-box interfaces toward expressive latent representations, helping explain why JDTQ dominates recent Transformer-based EUP research. This describes the direction of architectural development. In Tab.~\ref{EUP_nuscenes_table}, the greatest value stems from a JDTP design, but the leading entries span several sub-paradigms, and all rely on LiDAR, making modality and sub-paradigm indistinguishable. Within the camera-only categories, the highest values are reported by JDTQ. Consequently, the transition towards query representations is influenced by the joint-optimization depth and differentiability (Sec.~\ref{subsec:Intermediate Representations}), rather than any proven advantage in benchmark performance.

\textbf{The PNA-to-INA shift.} A second trend is the gradual internalization of association. Early methods across all three sub-paradigms, including CenterTrack, JDE, and TransTrack, relied on external, often non-differentiable matching (greedy heuristics, Hungarian assignment, IoU-based NMS), reflecting their modular tracking-by-detection roots. Later methods increasingly embed association into the network, as in STAR-Track's Latent Motion Model, FairMOT's shared embedding space, and MOTR's integrated query propagation. JDTQ stands out as inherently INA by design, which likely explains its rapid adoption, sidestepping the differentiability bottlenecks (Sec.~\ref{subsec:Intermediate Representations}) that still constrain JDTP- and JDTE-based approaches built on bounding-box interfaces.

\textbf{Modality biases and trade-offs.} Sub-paradigm and modality are strongly coupled. JDTP favors LiDAR for its precise spatial cues (CenterPoint, VoxelNeXt, SimTrack), whereas JDTE and JDTQ remain predominantly camera-based, as embeddings and learned queries derive most naturally from visual textures. This trend is reflected in the quantitative results (Tab.~\ref{EUP_nuscenes_table}), where LiDAR-based methods generally report higher AMOTA than camera-only approaches across all three sub-paradigms. The trade-offs are complementary: LiDAR is spatially precise but vulnerable under erratic motion or occlusion. Camera-based JDTE is more occlusion-robust through appearance persistence but ambiguous between similar instances and JDTQ offers the richest representation and long-term reasoning, especially with memory (MeMOT, MeMOTR, SambaMOTR), at an attention-based compute cost that may limit latency-critical deployment.

\textbf{Interpretability and deployment trade-offs.} The sub-paradigms also differ in interpretability and deployment readiness. JDTP and PNA-based JDTE preserve human-interpretable bounding-box and motion-cue interfaces, easing validation and modular integration, but inherit the non-differentiable bottlenecks of their ancestry. JDTQ-based methods achieve the tightest joint optimization, yet their query spaces remain opaque, with identity decisions emerging from attention patterns. This optimization–verifiability tension shapes which paradigms are deployable in near-term safety-critical AV stacks.

\textbf{Open challenges.} Several challenges remain specific to EUP. First, long-term occlusion handling and re-identification are unresolved. Memory-augmented JDTQ methods show promise but have not been systematically benchmarked against JDTP/JDTE under controlled occlusion. Second, sensor-modality coverage is uneven: LiDAR-based JDTE and multi-modal JDTQ remain underexplored relative to camera-based counterparts, leaving open whether embeddings and queries transfer to point-cloud domains. RADAR is absent from EUP altogether, despite its robustness to adverse weather and its native velocity measurements. No surveyed method operates on RADAR input, marking joint detection and tracking on RADAR as entirely unexplored. Third, the PNA-to-INA transition is incomplete even in ostensibly end-to-end designs. TubeTK's tube-level NMS and TransTrack's residual IoU matching show fully differentiable association is still the exception, motivating differentiable matching such as Sinkhorn-based alternatives (\cite{cuturi2013sinkhorn}, Sec.~\ref{subsec:Intermediate Representations}).

\section{Late-Unified Perception}
\label{chap:LateUnifiedPerception}

Joint tracking and prediction methods, summarized into LUP, integrate tracking and prediction into a unified module and are motivated by two key considerations. First, jointly addressing tracking and prediction can reduce identity switches and track fragmentations \cite{weng2022whose}, which directly influence prediction performance \cite{xu2024towards}. Second, shared spatio-temporal information about the ego vehicle's environment can be directly leveraged, enabling coherent reasoning over object dynamics in the past and future.

\begin{figure}[h]
\centering
\includegraphics[width=0.9\linewidth]{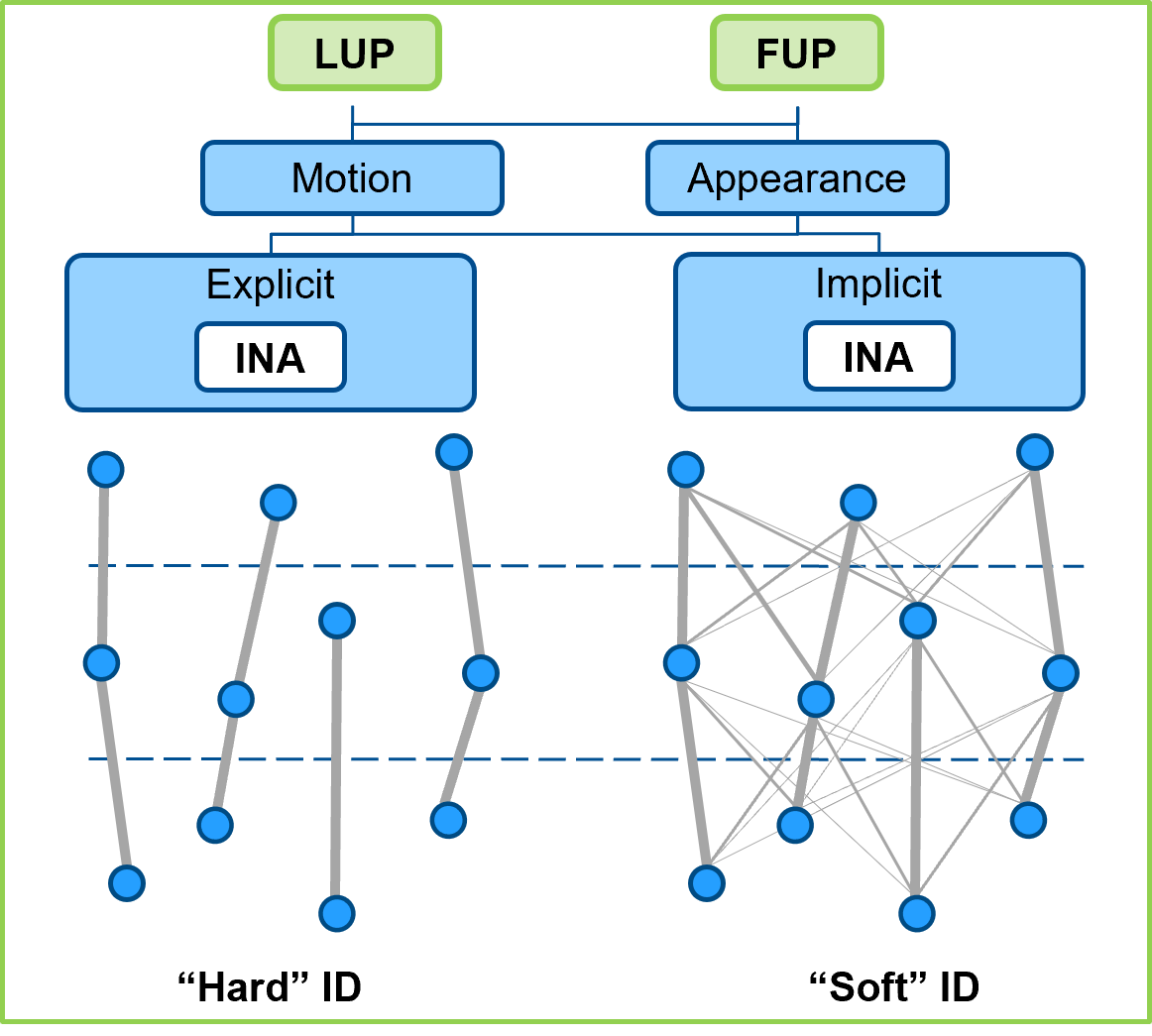}
\caption{Overview of the two tracking paradigms in unified perception. EUP tracks explicitly, outputting unique ID assigned objects. LUP and FUP, through unification, are naturally INA, however tracking can be used explicitly or implicitly in the module.}
\label{LUP_explicit_vs_implicit_tracking}
\end{figure}

Within our taxonomy (Sec.~\ref{chap:UnifiedPerceptionTaxonomy}), LUP occupies the converse position to EUP (Sec.~\ref{chap:EarlyUnifiedPerception}). Tracking here constitutes an intermediate stage that feeds prediction, rather than the  system output. This shift relaxes the constraint that identities must resolve to discrete, persistent tracks, opening the design space to both explicit (hard-association) and implicit (soft-association) tracking formulations, as illustrated in Fig.~\ref{LUP_explicit_vs_implicit_tracking}. Conversely, PNA becomes infeasible in this setting, as resolving associations post hoc would inhibit the joint reasoning process and prevent shared optimization between tracking and prediction, leaving INA as the only viable association mechanism. LUP can take detection results in the form of bounding box, occupancy, or latent representations as input, and output trajectory, occupancy, or flow predictions.

The LUP methods surveyed are summarized in Tab.~\ref{LUP_table}, which provides complementary information about inputs, network architecture, learning strategy, datasets, and code availability in accordance with the survey scope. As most surveyed LUP methods use motion cues, the following discussion is centered around their tracking formulation, working principles, and output representations. Intermediate representation types are pointed out throughout.

\setcounter{footnote}{0}
\begin{table*}[!t]
    \centering
    \caption{Late-Unified Perception Approaches: Overview}
    \label{LUP_table}
    \resizebox{\textwidth}{!}
    {%
    \begin{tabular}{l c c c c c c c c r}
        \toprule
        \textbf{Publ.}& \textbf{Year} & \textbf{Modality}\footnotemark & \textbf{Input}\footnotemark & \textbf{Paradigm}\footnotemark & \textbf{Output}\footnotemark & \textbf{Architecture} & \textbf{Learning Strategy}\footnotemark & \textbf{Dataset}\footnotemark & \textbf{Code} \\
        \midrule

        PTP \cite{weng2021ptp}                      & 2021 & Agn      & BB, (1+M, -)  & M, Imp. T        & T(3, V,P,C)   & GNN + Diversity Sampling + CVAE \cite{lee2017desire}                              & Emp. W.L. & kitti, nuScenes               & \xmark \\ 
        MTP\cite{weng2022mtp}                       & 2022 & Agn      & BB, (4, -)    & M, Imp. T        & T(6, V,P,C)   & Multi-Hypothesis Association + Predictor + Sampling                               &  -        & kitti, nuScenes               & \xmark \\   
        AffiniPred \cite{weng2022whose}             & 2022 & Agn      & BB, (4, Map)  & M, Imp. T        & T(6, V,P,C)   & Affinity-based Transformer + CVAE                                                 & Emp. W.L. & kitti, nuScenes               & \xmark \\ 
        TTFD \cite{zhang2023towards}                & 2023 & Agn      & BB, (4, -)    & M, Imp. T        & T(3, V)       & GRU-based Affinity-Motion E.-D.                                                   & Emp. W.L. & nuScenes, Argoverse           & \xmark \\  
        PF-Track\cite{pang2023standing}             & 2023 & MVI    & LR, (1+M, -)  & A + M, Exp. T    & T(4, V,P,C)   & Attention-based Past\cite{ngiam2021scene} + Future Reasoning                                           & Emp. W.L. & nuScenes                      & \cmark \\
        StreamMOTP \cite{zhuang2024streammotp}      & 2024 & Agn      & BB, (4, -)    & M, Exp. T        & T(6, V,P,C)   & Spatio-temporal E. + Interaction + Dual-stream Predictor                          & Emp. W.L. & nuScenes                      & \xmark \\ 
        Uhlemann et al. \cite{uhlemannexploring}    & 2025 & Agn      & BB, (20, Map) & M, Imp. T        & O(6, P)       & Transformer E.\cite{uhlemann2025snapshot}-D.\cite{zhou2022hivt}                   & Emp. W.L. & Argoverse2                    & \xmark \\ 

        \bottomrule

        \addlinespace[4pt]
        
        \multicolumn{10}{l}{\footnotesize \textsuperscript{6} Dataset: (kitti:\cite{geiger2013vision}, nuScenes:\cite{caesar2020nuscenes}, Argoverse: \cite{chang2019argoverse}, Argoverse2:\cite{Argoverse2}) }\\
        \multicolumn{10}{l}{\footnotesize \textsuperscript{5} Learning Strategy: (Emp. W.L.: Empirical Weighted Learning, Unb. W.L.: Unbalanced Weighted Learning, Unc. W.L.: Uncertainty Weighted Learning \cite{kendall2018multi}, -: No Information)} \\
        \multicolumn{10}{l}{\footnotesize \textsuperscript{4} Output: T: Trajectory, O: Occupancy ([Prediction Horizon in s], [Class Types Predicted (V: Vehicle, P: Pedestrian, C: Cyclist)]) } \\
        \multicolumn{10}{l}{\footnotesize \textsuperscript{3} Paradigm: ([M: Motion, A: Appearance], [Imp. T: Implicit Tracking, Exp. T: Explicit Tracking])} \\
        \multicolumn{10}{l}{\footnotesize \textsuperscript{2} Input: BB: Bounding Boxes, LR: Latent Representations (Number of Frames Input, M: Use of Memory, Map: Use of Map Information) } \\
        \multicolumn{10}{l}{\footnotesize \textsuperscript{1} Modality: (Agn: Agnostic, MVI: Multi-View Image)}  \\
        
    \end{tabular}
    }
\end{table*}

Implicit LUP approaches with trajectory output first appear with Weng et al., introducing PTP \cite{weng2021ptp}, a parallel framework whose GNN encodes bounding-box and fed-back trajectory information into shared interaction features for two heads. These are an explicit 3D MOT head and a CVAE-based forecasting head \cite{lee2017desire}, with prediction relying solely on latent GNN features rather than association results. MTP \cite{weng2022mtp} relaxes this further by generating multiple plausible tracking hypotheses via Murty's H-best assignment \cite{cox1996efficient} and letting the prediction module reason jointly over all of them, though the hypotheses are produced by an external, non-differentiable assignment step rather than within the network. AffiniPred \cite{weng2022whose} removes tracklet generation altogether, instead encoding ``soft'' identity cues as cross-frame affinity matrices that a Transformer consumes directly to model interactions and predict trajectories. This is a strictly implicit, latent-representation formulation. TTFD \cite{zhang2023towards} follows a similar affinity-driven philosophy, estimating short- and long-term detection similarities that drive a recurrent state-update process, decoded into trajectories via a social interaction model, thereby preserving identity uncertainty rather than committing to fixed tracks. \\ \indent
Implicit LUP with occupancy output remains comparatively rare. Uhlemann et al. \cite{uhlemannexploring} eliminate object association entirely by leveraging detections clustered with DBSCAN \cite{ester1996density} during training to define per-cluster Gaussian occupancy distributions, using a snapshot-inspired \cite{uhlemann2025snapshot} encoder followed by a HiVT-based \cite{zhou2022hivt} decoder. It is trained to directly predict occupancy from raw bounding boxes, propagating purely latent representations and reasoning over the scene holistically without persistent identities at inference. \\ \indent
Explicit LUP approaches with trajectory output instead propagate uniquely identified instances with encoded motion histories, directly targeting long-term occlusion resistance, interpretability, and stable joint reasoning. PF-Track \cite{pang2023standing} builds a detector-agnostic, SceneTransformer-based \cite{ngiam2021scene} framework around latent query representations. These queries follow the interface of JDTQ-style EUP detectors (Sec.~\ref{chap:EarlyUnifiedPerception}) but the detector itself is not jointly trained, keeping integration at the LUP level. It combines a past-reasoning module that refines object features from historical query states with a future-reasoning module that stabilizes propagation under occlusion via a track-extension mechanism for re-identification upon reappearance. StreamMOTP \cite{zhuang2024streammotp} advances this query-based line through a streaming, memory-bank design that retains long-term latent object features, associates identities via differentiable log-Sinkhorn optimal transport \cite{sarlin2020superglue}, and predicts trajectories through a dual-stream decoder combining current-detection and historical, intention-aware reasoning. This tightly couples long-term identity preservation with prediction.

\subsection{Quantitative LUP Analysis}

Table~\ref{LUP_nuscenes_table} reports minADE and minFDE on nuScenes for the LUP approaches with such results available. As LUP methods operate on pre-computed detections, they are modality-agnostic, so no such distinction is made. Results are comparable only within the two settings S1 and S2 established by Zhuang et al.~\cite{zhuang2024streammotp}, which differ in horizon, evaluated classes, and evaluation-time input. StreamMOTP~\cite{zhuang2024streammotp} reports the lowest errors in both, ahead of PTP~\cite{weng2021ptp} in S1 and of AffiniPred~\cite{weng2022whose} and MTP~\cite{weng2022mtp} in S2. As these methods also differ in architecture and training, the ordering does not support a general statement about tracking formulation.

\begin{table}[!t]
    \centering
    \caption{Late-Unified Perception Approaches: nuScenes Validation/ Test Set Performance, Original Reports}
    \label{LUP_nuscenes_table}
    \resizebox{\columnwidth}{!}{%
    \begin{tabular}{l c c c c c c c}
        \toprule
            \textbf{Publication} & \textbf{Year} & \textbf{Setting}\textsuperscript{1} & \textbf{Modality}\textsuperscript{2} & \textbf{Horizon (s)} & $k$\textsuperscript{3} & \textbf{minADE}$\downarrow$ & \textbf{minFDE}$\downarrow$ \\
            \midrule
            PTP\cite{weng2021ptp}                 & 2021 & S1 (val)  & Agn & 3 & 20 & 1.017 & 1.527 \\
            StreamMOTP\cite{zhuang2024streammotp} & 2024 & S1 (val)  & Agn & 3 & -  & 0.810 & 1.481 \\
            MTP\cite{weng2022mtp}                 & 2022 & S2 (test) & Agn & 6 & 10 & 1.585 & 2.512 \\
            AffiniPred\cite{weng2022whose}        & 2022 & S2 (test) & Agn & 6 & 10 & 0.977 & 1.628 \\
            StreamMOTP\cite{zhuang2024streammotp} & 2024 & S2 (test)  & Agn & 6 & -  & 0.757 & 1.487 \\
        \bottomrule
        \addlinespace[4pt]
        \multicolumn{7}{l}{\footnotesize \textsuperscript{1} Evaluation setting and split: (S1: GT input, all classes, nuScenes val split; S2: vehicle classes,} \\
        \multicolumn{7}{l}{\footnotesize \phantom{\textsuperscript{1}} real tracked detection histories, nuScenes prediction split termed test set in \cite{weng2022mtp,weng2022whose})} \\      
        \multicolumn{8}{l}{\footnotesize \textsuperscript{2} Modality: (Agn: Agnostic)} \\
        \multicolumn{8}{l}{\footnotesize \textsuperscript{3} Best-of-k predicted modes (-: unspecified)} \\
    \end{tabular}
    }
\end{table}

\subsection{Comparative LUP Analysis}

Across the surveyed LUP methods, several development trends and open challenges emerge.

\textbf{Implicit tracking.} Implicit LUP methods trace a clear progression toward looser identity commitments. PTP retains an explicit MOT head, MTP softens this into multiple externally-generated hypotheses, AffiniPred and TTFD replace discrete tracks with learned cross-frame affinities, and Uhlemann et al. dispense with object-wise reasoning altogether in favor of holistic occupancy distributions. This trades interpretability and persistent identity for robustness to association errors and tighter joint optimization \cite{weng2022whose, xu2024towards}. The trend points toward fully identity-free formulations as LUP's endpoint, leaving open whether they robustly scale to dense, multi-class scenes. 

\textbf{Explicit LUP and the EUP connection.} Explicit LUP methods pursue the converse trade-off, prioritizing interpretable, persistent identities suited to occlusion-heavy scenarios and safety verification. Whether this trade-off pays off empirically cannot be read from Tab.~\ref{LUP_nuscenes_table}. StreamMOTP reports the lowest errors in both settings, but differs from PTP, MTP and AffiniPred in architecture and memory design as well. Notably, both PF-Track and StreamMOTP adopt the query-based interface popularized in JDTQ-style EUP (Sec.~\ref{chap:EarlyUnifiedPerception}), reinforcing that standardized, interoperable intermediate representations would benefit unified perception as a whole.

\textbf{Trajectory-dominant, occupancy-scarce outputs.} Despite the taxonomy's allowance for occupancy and flow outputs, trajectory prediction dominates current LUP research. This leaves open whether the benefits of holistic, identity-free occupancy reasoning generalize beyond a single proof-of-concept, marking occupancy-centric LUP as an underexplored direction that bridges toward occupancy-based FUP (Sec.~\ref{chap:FullUnifiedPerception}).

\textbf{Memory as a unifying mechanism.} Long-term temporal reasoning increasingly relies on explicit memory regardless of tracking formulation. TTFD's recurrent affinity-state updates and StreamMOTP's persistent memory bank both extend context beyond adjacent frames, suggesting that memory-augmented designs, already prominent in JDTQ-based EUP (Sec.~\ref{chap:EarlyUnifiedPerception}), are emerging as a cross-paradigm strategy against occlusion and long-horizon uncertainty.

\textbf{Open challenges.} Since post hoc association would sever joint reasoning, LUP must resolve association within a differentiable, jointly-optimized pipeline. This shift calls for further development of model architecture and the design of differentiable association mechanisms that match the performance of their established non-differentiable counterparts, echoing the PNA-to-INA challenge seen in EUP. LUP methods also consume detection outputs rather than raw sensor streams, which hurts when detection is noisy or incomplete. Finally, the scarcity of occupancy- and flow-output methods and the narrow detector-agnostic validation of approaches like PF-Track leave output representations and cross-detector generalization open for this paradigm.

\section{Full-Unified Perception}
\label{chap:FullUnifiedPerception}

Joint detection, tracking and prediction methods, summarized into FUP, integrate scene understanding as one unified module, enabling holistic reasoning about the perception task. FUP approaches are increasingly being proposed and have, to some extent, been surveyed by Dal'Col et al. \cite{dal2024joint} as "joint detection and prediction" methods. Our scope differs in two respects. First, their review includes methods extending to planning \cite{cui2021lookout, khurana2022differentiable, sadat2020perceive, zeng2019end, casas2021mp3, hu2022st, hu2023planning, ye2023fusionad, jiang2023vad} or localization, which we strictly exclude from FUP (Fig.~\ref{Unified_Perception_Overview}) and regard as end-to-end driving or end-to-end perception, respectively (Sec.~\ref{chap:StateOfTheArt}). FUP is confined to scene understanding. Second, their analysis is primarily input–output focused, whereas we characterize methods systemically within our taxonomy (Sec.~\ref{chap:UnifiedPerceptionTaxonomy}).

Within our taxonomy, FUP represents the highest level of task integration. Detection, tracking, and prediction are addressed jointly within a shared module rather than chained across separate stages. As in LUP (Sec.~\ref{chap:LateUnifiedPerception}), tracking here constitutes an intermediate stage whose outputs feed prediction rather than terminating the pipeline. This is why FUP, like LUP, accommodates both implicit and explicit tracking formulations and likewise excludes PNA, as post hoc association would, in either paradigm, cut the joint reasoning process and prevent shared optimization across tasks. FUP additionally subsumes detection itself into this joint reasoning, distinguishing it from LUP, where detections are fed as input.

\begin{figure}[h]
\centering
\includegraphics[width=\linewidth]{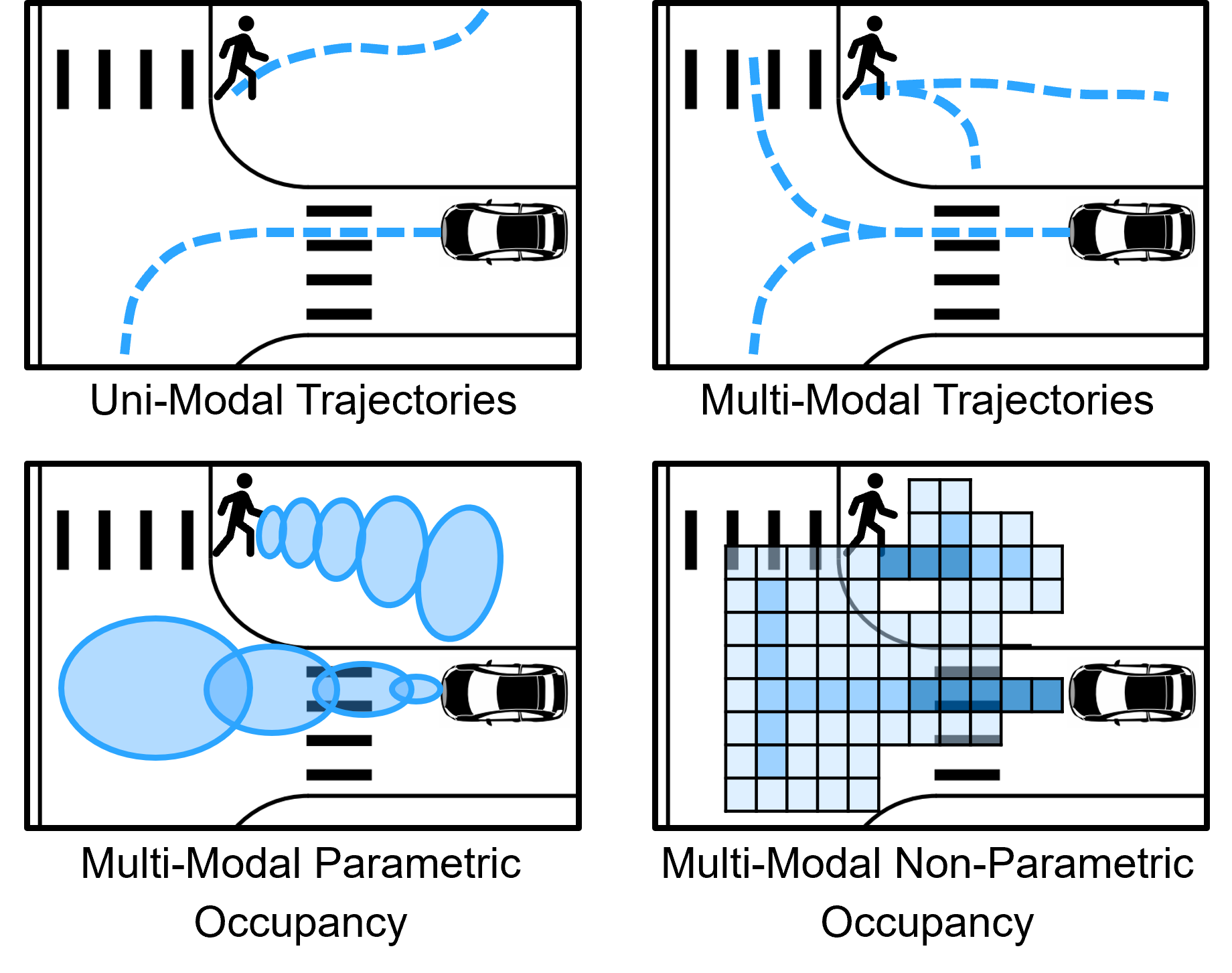}
\caption{Common prediction output representation types. For one, uni- or multi-modal trajectories are used. For the other, occupancy, in parametric and non-parametric representations can be output.}
\label{LUP_FUP_trajectory_representations}
\end{figure}

From that, FUP can be designed with camera, LiDAR, and RADAR combination inputs and output motion in the form of trajectories, occupancy, or flow. Our survey focuses on trajectory and occupancy, illustrated in Fig.~\ref{LUP_FUP_trajectory_representations}, due to their common adoption in real-world driving systems. For FUP with flow output, we refer to Dal'Col et al. \cite{dal2024joint} for deeper insights, noting that our taxonomy holds for these methods. Intermediate representations propagated in FUP can be of bounding box, occupancy, and latent nature, and tracking can be performed implicitly and explicitly, as illustrated in Fig.~\ref{LUP_explicit_vs_implicit_tracking}.

A summary of the surveyed FUP methods is provided in Tab.~\ref{FUP_table}, which provides complementary information to the following method discussion. Because FUP takes raw sensor information, tracking cues are mostly appearance-motion combinations. We discuss methods based on input and output, intermediate representation, as well as their tracking formulation and working principles.
\setcounter{footnote}{0}
\begin{table*}[!t]
    \centering
    \caption{Full-Unified Perception Approaches: Overview}
    \label{FUP_table}
    \resizebox{\textwidth}{!}
    {%
    \begin{tabular}{l c c c c c c c c r}
        \toprule
        \textbf{Publ.}& \textbf{Year} & \textbf{Modality}\footnotemark & \textbf{Input}\footnotemark & \textbf{Paradigm}\footnotemark & \textbf{Output}\footnotemark & \textbf{Architecture} & \textbf{Learning Strategy}\footnotemark & \textbf{Dataset}\footnotemark & \textbf{Code} \\
        \midrule  

        PIP \cite{jiang2022perceive}                & 2022 & MVI    & (4, -)    & M, Imp. T         & T(6, V)       & ResNet-50\cite{he2016deep} + Transformer E.-D. + Motion Interactor                 & Emp. W.L. & nuScenes                      & \xmark \\     %
        JLA \cite{kesa2022multiple}                 & 2022 & I      & (10, -)   & A, Exp. T         & T(2, P)       & FairMOT\cite{zhang2021fairmot} (JDTE, Table~\ref{EUP_table}) + RNN D.              & Unc. W.L. & MOT15/16/17/20                & \xmark \\ 
        S2F2 \cite{chen2022s2f2}                    & 2022 & I      & (1+M, -)  & A, Exp. T         & T(1, P)       & FairMOT\cite{zhang2021fairmot} (JDTE, Table~\ref{EUP_table}) + GRU + Flow D.       & Unc. W.L. & MOT17/20                      & \xmark \\     %
        ViP3D \cite{gu2023vip3d}                    & 2023 & MVI    & (1+M, Map)& A, Exp. T         & T(6, V,P)     & MOTR\cite{zeng2022motr} (JDTQ, Table~\ref{EUP_table}) + Query-based Predictor     & Unb. W.L. & nuScenes                      & \cmark \\     %
        ODTP \cite{cheng2023end}                    & 2023 & MVI    & (1+M, Map)& M, Exp. T         & T(4, V,P,C)   & QD3DT\cite{hu2022monocular} (JDTP, Table~\ref{EUP_table}) + DCENet++\cite{cheng2021exploring}&  - & nuScenes                 & \xmark \\     %
        FaF \cite{luo2018fast}                      & 2018 & L      & (5, -)    & M, Imp. T         & T(1, V)       & One-stage CNN                                                                     & Emp. W.L. & ATG4D                         & \xmark \\     %
        IntentNet \cite{casas2018intentnet}         & 2018 & L      & (10, Map) & M, Imp. T         & T(3, V)       & Two-stream CNN                                                                     & Emp. W.L. & ATG4D                         & \xmark \\     %
        SPF2 \cite{weng2021inverting}               & 2020 & L(RV)  & (2, -)    & M, Imp. T         & T(3, Agn)       & Unsupervised CNN E. + LSTM + CNN D.                                               & Emp. W.L. & kitti, nuScenes               & \xmark \\     %
        LaserFlow \cite{meyer2020laserflow}         & 2020 & L(RV)  & (5, -)    & A + M, Imp. T     & T(3, V,P,C)   & CNN Backbone + Probabilistic Trajectory Head                                      & Emp. W.L. & nuScenes, ATG4D               & \xmark \\     %
        SpAGNN \cite{casas2020spagnn}               & 2020 & L      & (10, Map) & A + M, Imp. T     & T(3, V)       & PIXOR-backbone\cite{yang2018pixor} + Spatial-aware GNN                            & -         & nuScenes, ATG4D               & \xmark \\     %
        PnPNet \cite{liang2020pnpnet}               & 2020 & L      & (5, Map)  & A + M, Exp. T     & T(3, V,P)     & CNN + Discrete-continuous Tracker + Predictor                                     & Unb. W.L. & nuScenes, ATG4D               & \xmark \\     %
        STINet \cite{zhang2020stinet}               & 2020 & L      & (6, Map)  & A + M, Imp. T     & T(3, P)       & CNN Temporal RPN                                                                  & Emp. W.L. & WOD, Lyft                     & \xmark \\     %
        ILVM \cite{casas2020implicit}               & 2020 & L      & (1+M, Map)& A + M, Imp. T     & T(5, V)       & CNN + GNN + Trajectory D.                                                         & Emp. W.L. & nuScenes, ATG4D               & \xmark \\     %
        Casas et al.\cite{casas2020importance}      & 2020 & L      & (1, Map)  & A + M, Imp. T     & T(5, V)       & SpAGNN-inspired\cite{casas2020spagnn} + MTP \cite{cui2019multimodal} Sampler      & Emp. W.L. & nuScenes, ATG4D               & \xmark \\     %
        SDP-Net \cite{zhang2020sdp}                 & 2020 & L      & (5, -)    & M, Exp. T         & T(0.5, V)     & BEV Scene Flow Estimator + CNN Backbone                                           & Emp. W.L. & kitti                         & \xmark \\     %
        RV-FuseNet \cite{laddha2021rv}              & 2021 & L(RV)  & (5, -)    & A + M, Imp. T     & T(3, V)       & CNN Backbone + Per-point Prediction                                               & Unb. W.L. & nuScenes                      & \xmark \\     %
        MultiXNet \cite{djuric2021multixnet}        & 2021 & L      & (10, Map) & M, Imp. T         & T(3, V,P,C)   & IntentNet\cite{casas2018intentnet} + Prediction Refinement Stage                  & Emp. W.L. & nuScenes, ATG4D               & \xmark \\     %
        SDAPNet \cite{ye2021sdapnet}                & 2021 & L      & (5, -)    & A + M, Imp. T     & T(2.5, V)     & EfficientNet-backbone\cite{tan2019efficientnet} + Multi-task Head                 & Emp. W.L. & nuScenes                      & \xmark \\     %
        FutureDet \cite{peri2022forecasting}        & 2022 & L      & (20, -)   & M, Exp. T         & T(3, V,P)     & CenterPoint-backbone\cite{yin2021center} + Backcasting                            & -         & nuScenes                      & \cmark \\     %
        FS-GRU \cite{chen2022fs}                    & 2022 & L      & (1+M, Map)& A + M, Imp. T     & T(3, V)       & FPN-backbone\cite{lin2017feature} + ConvGRU + D.                                  & Emp. W.L. & Argoverse                     & \xmark \\     %
        DeTra \cite{casas2024detra}                 & 2024 & L      & (5, Map)  & A + M, Imp. T     & T(5, V)       & PointNet-backbone\cite{qi2017pointnet} + Refinement Transformer                   & Emp. W.L. & Argoverse2, WOD               & \xmark \\     %
        Interaction Transformer \cite{li2020end}    & 2020 & I+L    & (10, Map) & A + M, Imp. T     & T(3, V)       & CNN E. + Interactive Forecasting Transformer                                      & Emp. W.L. & nuScenes, ATG4D               & \xmark \\     %
        LiRaNet \cite{shah2020liranet}              & 2020 & L+R    & (5, Map)  & M, Imp. T         & T(3, V,P,C)   & Spatio-temporal RADAR Feature Learning + PnPNet\cite{liang2020pnpnet}             & Emp. W.L. & nuScenes, X17k                & \xmark \\     %
        MVFuseNet \cite{laddha2021mvfusenet}        & 2021 & L+L(RV)& (5, Map)  & A + M, Imp. T     & T(3, V,P,C)   & Multi-view Asymmetric U-Net\cite{ronneberger2015u} + CNN Heads                    & Unb. W.L. & nuScenes                      & \xmark \\     %
        Fadadu et al. \cite{fadadu2022multi}        & 2022 & I+L+L(RV)&(10, Map)& A + M, Imp. T     & T(3, V,P,C)   & Multi-view Fusion + MultiXNet\cite{djuric2021multixnet}                           & Emp. W.L. & nuScenes, ATG4D               & \xmark \\     %
        \midrule

        FIERY \cite{hu2021fiery}                    & 2021 & MVI    & (3, -)    & A + M, Exp. T     & O(2, V)       &  LSS\cite{philion2020lift} + CNN-GRU                                              & Unb. W.L. & nuScenes, Lyft                & \cmark \\     %
        BEVerse \cite{zhang2022beverse}             & 2022 & MVI    & (3, Map)  & A + M, Imp. T     & O(2, V,P,C)   &  Transformer E. + Parallel D. Heads                                               & Emp. W.L. & nuScenes                      & \cmark \\     %
        StretchBEV \cite{akan2022stretchbev}        & 2022 & MVI    & (3, -)    & A + M, Imp. T     & O(6, V)       &  Stochastic Residual State-space Temporal Model                                   &  -        & nuScenes                      & \cmark \\     %
        PowerBEV \cite{li2023powerbev}              & 2023 & MVI    & (3, -)    & A + M, Exp. T     & O(2, V)       &  LSS\cite{philion2020lift} + Multi-scale U-Net-inspired\cite{ronneberger2015u} E.-D.  & Unc. W.L. & nuScenes                  & \cmark \\     %
        TBP-Former \cite{fang2023tbp}               & 2023 & MVI    & (3, Map)  & A + M, Imp. T     & O(2, V,P)     &  Spatio-temporal Pyramid Transformer                                              &  -        & nuScenes                      & \cmark \\     %
        SA-GNN \cite{luo2021safety}                 & 2021 & L      & (5, Map)  & A + M, Imp. T     & O(7, P)       &  CNN + Scene-Actor GNN                                                            & Emp. W.L. & nuScenes, ATG4D               & \xmark \\     %
        ImplicitO \cite{agro2023implicit}           & 2023 & L      & (5, Map)  & A + M, Imp. T     & O(5, V)       &  PIXOR\cite{yang2018pixor} E. + Implicit D.                                       & Emp. W.L. & Argoverse2, HighwaySim        & \xmark \\     %
        Khurana et al.\cite{khurana2023point}       & 2023 & L      & (6, -)    & A + M, Imp. T     & O(3, Agn)   &  Voxelization +  Differentiable volumetric Render                                 & -         & nuScenes                      & \cmark \\     %
        Occ4Cast \cite{liu2024lidar}                & 2024 & L      & (10, -)   & A + M, Imp. T     & O(1, Agn)       &  CNN E. + LSTM + D.                                                               & -         & Lyft, Argoverse, ApolloScape  & \cmark \\     %
        FISHING Net \cite{hendy2020fishing}         & 2020 & MVI+L+R& (5, -)    & A + M, Imp. T     & O(2, V,P,C)   &  Per-modality CNN E.-D. + Aggregation                                             & Emp. W.L. & nuScenes, Lyft                & \xmark \\     %
        \bottomrule

        \addlinespace[4pt]

        \multicolumn{10}{l}{\footnotesize \textsuperscript{6} Dataset: (kitti:\cite{geiger2013vision}, nuScenes:\cite{caesar2020nuscenes}, WOD:\cite{sun2020scalability}, Lyft: \cite{houston2021one}, Argoverse: \cite{chang2019argoverse}, Argoverse2:\cite{Argoverse2}, ATG4D: \cite{luo2018fast}, X17k: \cite{shah2020liranet}, HighwaySim: \cite{agro2023implicit}, ApolloScape: \cite{huang2018apolloscape}, MOT15:\cite{MOTChallenge2015}, MOT16:\cite{MOT16}, MOT17:\cite{MOT16}, MOT20:\cite{MOTChallenge20}) }\\
        \multicolumn{10}{l}{\footnotesize \textsuperscript{5} Learning Strategy: (Emp. W.L.: Empirical Weighted Learning, Unb. W.L.: Unbalanced Weighted Learning, Unc. W.L.: Uncertainty Weighted Learning \cite{kendall2018multi}, -: No Information)} \\
        \multicolumn{10}{l}{\footnotesize \textsuperscript{4} Output: T: Trajectory, O: Occupancy ([Prediction Horizon in s], [Class Types Predicted (V: Vehicle, P: Pedestrian, C: Cyclist, Agn: Agnostic to Class)]) } \\
        \multicolumn{10}{l}{\footnotesize \textsuperscript{3} Paradigm: ([M: Motion, A: Appearance], [Imp. T: Implicit Tracking, Exp. T: Explicit Tracking])} \\
        \multicolumn{10}{l}{\footnotesize \textsuperscript{2} Input: (Number of Frames Input, M: Use of Memory, Map: Use of Map Information) } \\
        \multicolumn{10}{l}{\footnotesize \textsuperscript{1} Modality: (I: Image, MVI: Multi-View Image, L: LiDAR, L(RV): LiDAR Range View, R: RADAR)  }  \\
        
    \end{tabular}
    }
\end{table*}

\subsection{Trajectory Output and Implicit Tracking}

Trajectory-based FUP methods leveraging implicit tracking have been proposed for camera, LiDAR, and multi-modal inputs, the order of which we keep in the following. Most propagate latent intermediate representations. \\ \indent
For multi-view camera inputs, PIP \cite{jiang2022perceive} uses a Transformer-based framework in which perception and motion queries interact with agent-wise map features to jointly perform online mapping, detection, and forecasting, with implicit tracking emerging from a unified latent scene representation. \\ \indent
LiDAR-based methods form the largest cluster and trace several recurring design lines. Early architectures include FaF \cite{luo2018fast}, which stacks LiDAR sweeps into a 4D tensor and aggregates overlapping single-stage detections to implicitly handle tracking and short-term forecasting, and IntentNet \cite{casas2018intentnet}, which adds HD maps and a two-stream backbone to jointly output detections, intentions, and trajectories. Graph- and interaction-based approaches refine actor states via message passing: SpAGNN \cite{casas2020spagnn} applies GNNs over a PIXOR backbone \cite{yang2018pixor}, Casas et al. \cite{casas2020importance} extend it with traffic-rule priors and an MTP-based \cite{cui2019multimodal} probabilistic head and ILVM \cite{casas2020implicit} models interactions through a latent-variable graph and samples futures generatively. STINet \cite{zhang2020stinet} instead links bounding-box proposals over time via a Temporal Region Proposal Network and an interactive feature extractor. A further line operates closer to raw sensor space: SPF2 \cite{weng2021inverting} inverts the pipeline by forecasting future point clouds before detecting on them, while LaserFlow \cite{meyer2020laserflow} and RV-FuseNet \cite{laddha2021rv} fuse consecutive range images to predict trajectories through temporal alignment and segmentation. Finally, several methods target refinement and long-term reasoning: MultiXNet \cite{djuric2021multixnet} extends IntentNet with uncertainty-aware, two-stage forecasting. SDAPNet \cite{ye2021sdapnet} fuses multi-frame BEV LiDAR features through a multi-task detection head. FS-GRU \cite{chen2022fs} integrates a ConvGRU with HD-map alignment for long-horizon reasoning, and DeTra \cite{casas2024detra}, the most recent advancement, reformulates FUP as iterative trajectory refinement, updating BEV object queries through Transformer-based pose updates with implicit tracking emerging from temporally consistent query refinement. \\ \indent
Multi-modal methods extend this further. Interaction Transformer \cite{li2020end} fuses LiDAR, camera, and HD-map features into a BEV detection network followed by a recurrent interactive Transformer for spatio-temporal reasoning. LiRaNet \cite{shah2020liranet} additionally incorporates RADAR via modality-specific BEV volumes and a PnPNet-inspired \cite{liang2020pnpnet} prediction backbone. MVFuseNet \cite{laddha2021mvfusenet} jointly processes range-view and BEV representations across consecutive sweeps, and Fadadu et al. \cite{fadadu2022multi} fuse camera and LiDAR (point-cloud and range-view encodings) within a MultiXNet-style \cite{djuric2021multixnet} architecture. Across all of these, temporally consistent trajectory forecasting arises from feature aggregation and cross-modal association rather than explicit identity management.

\subsection{Trajectory Output and Explicit Tracking}

FUP methods with trajectory outputs and explicit tracking formulations have been proposed for camera and LiDAR inputs. Notably, several directly extend EUP backbones (Tab.~\ref{EUP_table}), which naturally employ explicit tracking, into the FUP setting. \\ \indent
JLA \cite{kesa2022multiple} extends FairMOT \cite{zhang2021fairmot} (JDTE) with an RNN-based forecasting head whose predicted future states replace Kalman-filter estimates during a three-step online association process. S2F2 \cite{chen2022s2f2} similarly builds on FairMOT, instead predicting residual future flow fields via a GRU encoder-decoder that adaptively guides association confidence. ViP3D \cite{gu2023vip3d} extends MOTR \cite{zeng2022motr} (JDTQ) with explicit, query-level association supported by a dynamic memory bank and a DETR3D-based \cite{wang2022detr3d} feature extractor, decoding agent queries into future trajectories via a map encoder. ODTP \cite{cheng2023end} couples QD3DT \cite{hu2022monocular} (JDTP) with the trajectory module DCENet++ \cite{cheng2021exploring}: persistent agent queries from QD3DT enable identity association and feed dynamic BEV maps, while DCENet++ applies self-attention and a CVAE to forecast diverse futures, trained directly on noisy tracker outputs via bounding-box intermediate representations. \\ \indent
In the LiDAR domain, PnPNet \cite{liang2020pnpnet} presents a differentiable, end-to-end framework that jointly outputs tracks and trajectories at each timestep through a discrete-continuous tracking module combining learned affinities with Hungarian matching, though gradient flow through this association step remains unaddressed. SDP-Net \cite{zhang2020sdp} instead derives BEV occupancy and flow maps from LiDAR to estimate object and ego-motion, warping past features for temporal aggregation and associating tracks spatially with predicted positions. FutureDet \cite{peri2022forecasting}, building on CenterPoint \cite{yin2021center} (JDTP), directly predicts and backcasts future object locations from LiDAR to infer trajectory-level identity association while supporting multi-modal forecasts.

\subsection{Occupancy Output and Implicit Tracking}

Occupancy-based approaches differ from trajectory-based methods in their output representation while still jointly reasoning about detection, tracking, and prediction within our taxonomy. Methods leveraging implicit tracking have been proposed for camera and LiDAR inputs. \\ \indent
In the camera domain, future occupancy is increasingly predicted via BEV-based approaches that maintain temporal consistency without explicit tracking. BEVerse \cite{zhang2022beverse} lifts multi-view, multi-timestamp image features into an ego-motion-aligned BEV space and decodes them into occupancy through a spatio-temporal encoder. StretchBEV \cite{akan2022stretchbev} instead models temporal dynamics within a stochastic latent space, using a ConvGRU-based recurrent module with stochastic latent variables to diversify long-horizon forecasts. TBP-Former \cite{fang2023tbp} employs a pose-synchronized BEV encoder and spatial-temporal pyramid Transformer to ensure spatial-temporal alignment despite arbitrary camera poses, guided by scene priors from the latest BEV representation. \\ \indent
In the LiDAR domain, SA-GNN \cite{luo2021safety} constructs a Scene-Actor Graph, pedestrians as nodes linked by spatial proximity, the scene as a supernode. This representation is processed via GNN message passing over voxelized LiDAR and map features to jointly predict per-pedestrian and scene-level occupancy distributions. ImplicitO \cite{agro2023implicit} instead employs continuous implicit occupancy representations, with a PIXOR-inspired \cite{yang2018pixor} BEV encoder producing multi-scale features that a query-point decoder converts into dense occupancy and flow forecasts without explicit object association. \\ \indent
A further group abandons object detection altogether, instead propagating latent spatio-temporal features that model scene evolution directly. For LiDAR inputs, Khurana et al. \cite{khurana2023point} estimate future 4D space-time occupancy along sensor rays defined by timestamp, origin, and direction, abstracting away sensor specifics to predict purely when and where occupied space will be encountered. Occ4Cast \cite{liu2024lidar} formulates future occupancy as dense binary estimation over spatio-temporal voxel grids using a convolutional encoder with recurrent LSTM modules. For multi-modal inputs, FISHING Net \cite{hendy2020fishing} fuses independent LiDAR, RADAR, and camera encoder-decoder outputs into a unified, ego-centric, top-down semantic grid for joint perception and prediction.

\subsection{Occupancy Output and Explicit Tracking}

Occupancy-based prediction with explicit tracking has so far only been proposed for image-based inputs. FIERY \cite{hu2021fiery} lifts multi-view monocular features into latent BEV representations via Lift-Splat-Shoot \cite{philion2020lift}, aligns them temporally with ego-motion, and uses a recurrent spatio-temporal model to decode instance centers, offsets, and occupancy. They achieve explicit tracking by warping and matching past instance locations across frames. PowerBEV \cite{li2023powerbev} similarly encodes multi-view images into BEV space but predicts segmentation jointly with backward centripetal flow, explicitly propagating instance IDs for consistent tracking with reduced multi-task complexity.

\subsection{Quantitative FUP Analysis}

Tables~\ref{FUP_traj_table} and~\ref{FUP_occ_table} report trajectory and occupancy prediction results for FUP approaches on the nuScenes validation set.

Table~\ref{FUP_traj_table} must be read with caution, as heterogeneous evaluation protocols make direct comparison largely unreliable. Horizons range from 2.5 to 6 seconds, evaluated class sets differ, ILVM~\cite{casas2020implicit} reports scene-level rather than per-agent metrics, SPF2~\cite{weng2021inverting} reports recall-integrated AADE/AFDE from an end-to-end matching protocol, and upstream detectors and training data differ throughout. Longer horizons inflate displacement errors, so the 6s camera-based PIP~\cite{jiang2022perceive} and ViP3D~\cite{gu2023vip3d} cannot be placed against the shorter-horizon entries. At 3s, all entries are LiDAR-based or LiDAR-inclusive, so no camera-versus-LiDAR comparison at a fixed horizon is possible here, and the multi-modal Interaction Transformer~\cite{li2020end} and LiRaNet~\cite{shah2020liranet} bracket rather than separate from the LiDAR-only entries, which reflects differing training setups as much as fusion design.

Table~\ref{FUP_occ_table} is more internally consistent, as all entries report Future Video Panoptic Quality (VPQ) on the same nuScenes validation set and modality (MVI), making occupancy the FUP subtask with the most homogeneous evaluation. Backbones, resolutions, evaluation ranges, and map usage nevertheless differ (Tab.~\ref{FUP_table}), and StretchBEV~\cite{akan2022stretchbev} reports a 6s horizon. Among the four entries at 2s, TBP-Former~\cite{fang2023tbp} and BEVerse~\cite{zhang2022beverse} report the highest VPQ, ahead of PowerBEV~\cite{li2023powerbev} and FIERY~\cite{hu2021fiery}, indicating progress without supporting a strict ordering.

\begin{table}[!t]
    \centering 
    \caption{Full-Unified Perception Approaches: nuScenes Validation Set Performance, Original Reports, Trajectory Prediction}
    \label{FUP_traj_table}
    \resizebox{\columnwidth}{!}{%
    \begin{tabular}{l c c c c c}
        \toprule
        \textbf{Publication} & \textbf{Year} & \textbf{Modality}\textsuperscript{1} & \textbf{Horizon (s)} & \textbf{ADE}$\downarrow$ & \textbf{FDE}$\downarrow$ \\
        \midrule
        PIP\cite{jiang2022perceive}                & 2022 & MVI   & 6   & 1.230                         & 1.750                         \\
        ViP3D\cite{gu2023vip3d}                    & 2023 & MVI   & 6   & 2.030                         & 2.900                         \\
        ODTP\cite{cheng2023end}                    & 2023 & MVI   & 4   & 0.970                         & 1.860                         \\
        SPF2\cite{weng2021inverting}               & 2020 & L(RV) & 3   & 1.044\textsuperscript{\dag}   & 1.043\textsuperscript{\dag}   \\
        LaserFlow\cite{meyer2020laserflow}         & 2020 & L(RV) & 3   & 0.990                         & -                             \\
        SpAGNN\cite{casas2020spagnn}               & 2020 & L     & 3   & 0.960                         & -                             \\
        PnPNet\cite{liang2020pnpnet}               & 2020 & L     & 3   & 0.580\textsuperscript{\ddag}  & 0.930\textsuperscript{\ddag}  \\
        ILVM\cite{casas2020implicit}               & 2020 & L     & 5   & 0.760\textsuperscript{\S}     & 1.530\textsuperscript{\S}     \\
        RV-FuseNet\cite{laddha2021rv}              & 2021 & L(RV) & 3   & 1.200                         & -                             \\
        SDAPNet\cite{ye2021sdapnet}                & 2021 & L     & 2.5 & 0.648                         & -                             \\
        Interaction Transformer\cite{li2020end}    & 2020 & I+L   & 3   & 1.120                         & -                             \\
        LiRaNet\cite{shah2020liranet}              & 2020 & L+R   & 3   & 0.630\textsuperscript{\ddag}  & 1.060\textsuperscript{\ddag}  \\
        \bottomrule
        \addlinespace[4pt]
        \multicolumn{6}{l}{\footnotesize \textsuperscript{1} Modality: (L: LiDAR, L(RV): LiDAR Range View, MVI: Multi-View Image, R: RADAR)} \\
        \multicolumn{6}{l}{\footnotesize \textsuperscript{\dag} Recall-averaged AADE/AFDE; not comparable to standard ADE/FDE.} \\
        \multicolumn{6}{l}{\footnotesize \textsuperscript{\ddag} Vehicles only.} \\
        \multicolumn{6}{l}{\footnotesize \textsuperscript{\S} Scene-level ADE/FDE (sADE/sFDE).} \\
    \end{tabular}
    }
\end{table}

\begin{table}[!t]
    \centering 
    \caption{Full-Unified Perception Approaches: nuScenes Validation Set Performance, Original Reports, Occupancy Prediction}
    \label{FUP_occ_table}
    \resizebox{\columnwidth}{!}{%
    \begin{tabular}{l c c c c}
        \toprule
        \textbf{Publication} & \textbf{Year} & \textbf{Modality}\textsuperscript{1} & \textbf{Horizon (s)} & \textbf{VPQ}$\uparrow$ \\
        \midrule
        FIERY\cite{hu2021fiery}              & 2021 & MVI & 2 & 29.9 \\
        BEVerse\cite{zhang2022beverse}       & 2022 & MVI & 2 & 36.1 \\
        StretchBEV\cite{akan2022stretchbev}  & 2022 & MVI & 6 & 29.0 \\
        PowerBEV\cite{li2023powerbev}        & 2023 & MVI & 2 & 33.9 \\
        TBP-Former\cite{fang2023tbp}         & 2023 & MVI & 2 & 36.9 \\
        \bottomrule
        \addlinespace[4pt]
        \multicolumn{5}{l}{\footnotesize \textsuperscript{1} Modality: (MVI: Multi-View Image)} \\
    \end{tabular}
    }
\end{table}

\subsection{Comparative FUP Analysis}
Across the surveyed FUP landscape, several cross-cutting trends and open challenges emerge.

\textbf{Extension vs. holistic design.} A clear architectural fork separates how FUP is approached. One line extends EUP backbones with a prediction module: JLA and S2F2 on FairMOT (JDTE), ViP3D on MOTR (JDTQ), and ODTP and FutureDet on QD3DT/CenterPoint (JDTP), retrofitting joint detection-and-tracking systems rather than designing all three tasks holistically. The other designs natively integrated architectures whose shared latent or graph-based reasoning spans detection, tracking, and prediction from the ground up (ILVM, SpAGNN, DeTra). This mirrors the EUP observation (Sec.~\ref{chap:EarlyUnifiedPerception}) that stage-level choices propagate forward, raising whether retrofitted systems can match the joint-optimization depth of natively designed ones, or whether the field will converge toward holistic architectures.

\textbf{Output representation trends.} Trajectory-output FUP remains more prevalent and mature than occupancy-output FUP, which gained traction only recently (BEVerse, ImplicitO, Khurana et al., Occ4Cast). A further split is visible: SA-GNN and ImplicitO still reason about discrete actors before producing occupancy, whereas Khurana et al. and Occ4Cast abandon object-centric detection for direct space-time occupancy, avoiding long-tail categorization but forfeiting object-level interpretability and downstream usability. This positions occupancy-centric, detection-free FUP as a promising yet underexplored direction, particularly relative to its faster adoption in modular and LUP contexts.

\textbf{Modality and representation biases.} LiDAR-centric designs dominate trajectory-output FUP, consistent with JDTP's LiDAR affinity in EUP (Sec.~\ref{chap:EarlyUnifiedPerception}), while camera-centric and multi-modal designs are more common among recent occupancy- and BEV-based methods (BEVerse, TBP-Former, FIERY, PowerBEV). Across nearly all sub-categories, latent representations dominate intermediate propagation, reinforcing the shift toward expressive, jointly-optimizable interfaces over interpretable bounding boxes identified for EUP and LUP (Sec.~\ref{subsec:Intermediate Representations}).

\textbf{Implicit-vs-explicit trade-offs.} As in LUP, implicit-tracking FUP methods outnumber explicit ones, prioritizing tight joint optimization through generative or graph-based reasoning (ILVM, DeTra) free from discrete-identity bottlenecks. Explicit-tracking methods (JLA, S2F2, ViP3D, ODTP, PnPNet, SDP-Net, FutureDet, FIERY, PowerBEV) instead preserve persistent, interpretable identities valuable for planning and safety verification, but inherit association bottlenecks from their modular and EUP ancestry. PnPNet's noted lack of gradient flow through its association step shows that EUP's PNA-to-INA differentiability challenge (Sec.~\ref{chap:EarlyUnifiedPerception}) persists even at the highest integration level.

\textbf{Open challenges.} Beyond the unresolved differentiability of association, generalization across heterogeneous sensor configurations remains underexplored, as the LiDAR-centric bias in trajectory-output FUP leaves camera-only and multi-modal robustness untested at scale. Finally, since many FUP methods extend EUP backbones, stage-level choices, such as representation type, tracking formulation, and sensor modality, propagate forward and constrain FUP. Standardizing interoperable EUP/LUP interfaces (Sec.~\ref{chap:EarlyUnifiedPerception}, Sec.~\ref{chap:LateUnifiedPerception}) would let future work combine extension-based efficiency with holistic joint-optimization depth.

\section{Discussion}
\label{chap:Discussion}

Having surveyed EUP, LUP, and FUP, with paradigm-specific trends and challenges addressed in their comparative analyses, we now discuss aspects that span all three: intermediate representations, training strategies, the limits of cross-paradigm comparison, and the future directions they imply. A recurring thread, already visible per paradigm, is the gradual shift from interpretable bounding-box interfaces and external association toward latent representations and in-network or implicit tracking, trading verifiability for joint-optimization depth.

\subsection{Intermediate Representations}
\label{subsec:Intermediate Representations}

\begin{figure}[ht]
\centering
\includegraphics[width=0.9\linewidth]{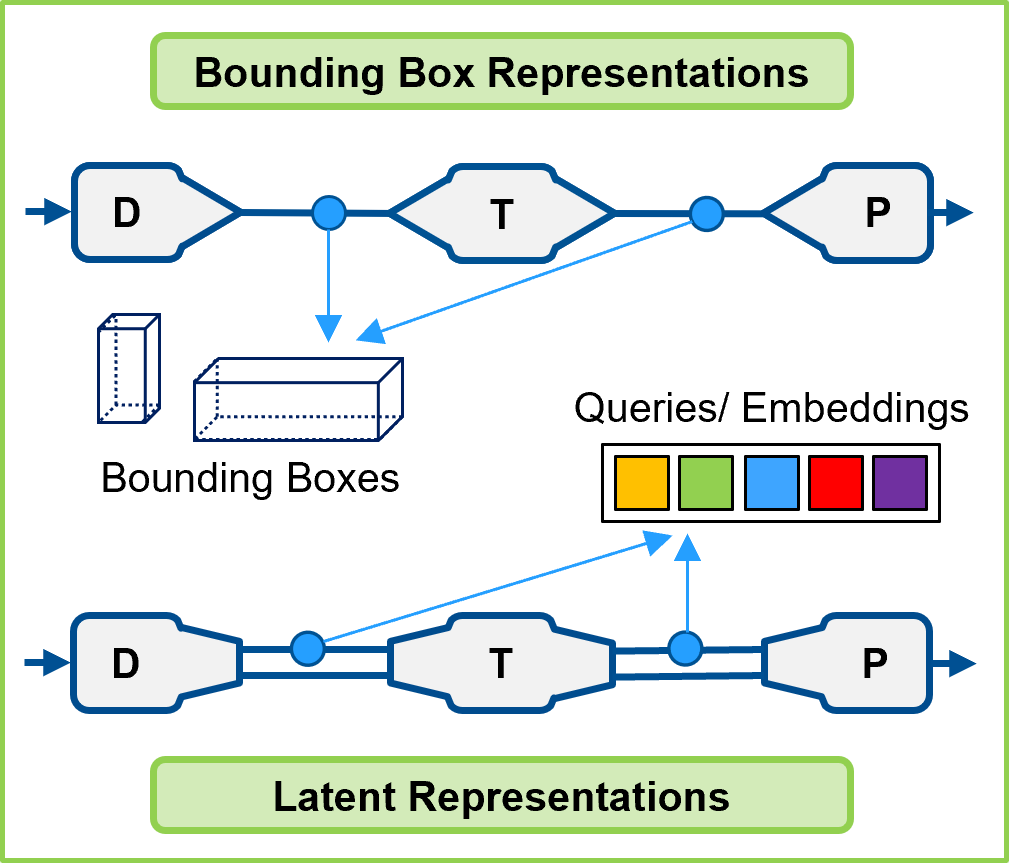}
\caption{Common intermediate representations propagated in unified perception.}
\label{unification_paradigms}
\end{figure}

Representations shared between unified detection-tracking and tracking-prediction structures can be of bounding box, latent, or occupancy nature. As the latter is not yet proposed, we discuss the two former, illustrated in Fig.~\ref{unification_paradigms}.

Bounding box representations (Fig.~\ref{unification_paradigms}, top) directly reuse modular perception components, unifying them through joint training. Gradients are backpropagated across all modules while explainability is preserved through human-interpretable outputs. This faces two key drawbacks, differentiability and information bottlenecks.

Differentiability is essential, as all operations must support gradient flow, yet steps such as argmax in matching or NMS in detection are typically non-differentiable, restricting joint optimization. EUP mitigates this through NMS-free detectors and post-network association that exclude such operations from unified training, as in CenterPoint \cite{yin2021center}. In LUP and FUP, bounding-box interfaces require bounding-box tracking outputs and thus explicit tracking, though explicit tracking also occurs under the latent paradigm. Differentiable matching alternatives such as the Sinkhorn algorithm \cite{cuturi2013sinkhorn} address this, as in StreamMOTP \cite{zhuang2024streammotp}, while FutureDet \cite{peri2022forecasting} instead follows a CenterPoint-based \cite{yin2021center} design that excludes matching from learnable structures.

Information bottlenecks arise because bounding-box interfaces compress rich features into a few geometric parameters, discarding spatial and semantic detail and potentially reducing the model's capacity to learn inter-task dependencies. Latent intermediate representations (Fig.~\ref{unification_paradigms}, bottom) instead relax module boundaries by propagating high-dimensional embeddings such as learned queries, preserving spatial and semantic information, avoiding non-differentiable operations, and enabling flexible designs. Tracking can be implicit, as identity and motion cues are encoded in the latent space, though explicit association remains possible. Unification then arises primarily from structural module combination rather than training, challenging both information bottlenecks and rigid cascading, and enabling tailored approaches such as DeTra \cite{casas2024detra}.

\subsection{Training Strategies}
\label{subsec:Training Strategies}

Training unified perception is inherently challenging, as multiple tasks and their objectives must be balanced at once. Tasks progress at different speeds and may conflict, making loss design and weighting decisive \cite{wang2024onetrack}, with difficulty growing as more tasks are jointly optimized \cite{singh2023end, chen2024end}.

Approaches fall into three complementary families. Loss-level balancing combines task losses through manually tuned weights (empirical weighted learning, standard in modular perception), scheduled weighting such as sigmoid annealing \cite{zhang2023towards}, or uncertainty-based weighting \cite{kendall2018multi}. Optimization-level strategies instead shape how gradients interact, as in DeTra's \cite{casas2024detra} selective gradient stopping between refinement stages, which reduces interference and unexpectedly improves performance. Architectural strategies shape multi-task learning through network structure: shared-yet-modular branches that limit negative transfer \cite{manole2024uolo, wang2024you, he2025md, azak2024unified, wang2025mtfenet}, or auxiliary task-specific heads whose dedicated losses supervise and regularize the shared representation. The most effective combination remains largely empirical. The per-paradigm learning strategies in Tab.~\ref{EUP_table},~\ref{LUP_table}, and~\ref{FUP_table} show no consensus, marking multi-task optimization as a central open problem.

\subsection{Cross-Paradigm Quantitative Comparison}
\label{subsec:Quantitative Comparison}

Each paradigm chapter reports a quantitative analysis within its own scope (Sec.~\ref{chap:EarlyUnifiedPerception},~\ref{chap:LateUnifiedPerception},~\ref{chap:FullUnifiedPerception}). A comparison across paradigms, however, is fundamentally constrained. EUP resolves persistent tracks while LUP and FUP output future trajectories or occupancy, so they are evaluated against different ground truth with different metrics. Even within a paradigm, heterogeneous datasets, horizons, modalities, and upstream detectors complicate fair comparison, limiting what can currently be concluded about the merit of unification depth.

\subsection{Future Research Directions}
\label{subsec:Future Research Directions}

Real-world adoption of unified perception requires advances in evaluation, representation, supervision, and verifiability. Building on the cross-cutting observations above and the per-paradigm challenges, we highlight the most pressing directions.

\noindent\textbf{Standardized and closed-loop evaluation:} A systematic comparison between unified, modular, and end-to-end perception across modalities is still missing \cite{xu2024towards}. Beyond task-specific metrics, closed-loop evaluation, coupling perception with decision-making and control, is needed to reveal how unification affects driving behavior and to make within-paradigm benchmarks (Sec.~\ref{subsec:Quantitative Comparison}) comparable across the field.

\noindent\textbf{Data efficiency and supervision:} The field trends toward larger, increasingly unified datasets, yet unified perception still depends on task-level annotations (boxes, tracks, trajectories), unlike full end-to-end stacks that learn from raw driving logs. Reducing this annotation burden through semi-, self-, or weakly-supervised objectives is a key enabler for scaling. Equally important is the representativeness of the data itself. Single benchmarks such as nuScenes are increasingly insufficient, motivating unified dataset frameworks that harmonize and combine multiple datasets for training and evaluation \cite{feng2024unitraj, dauner2026123d}, alongside emerging large-scale, reasoning-guided datasets \cite{tian2025nuscenes}. Such cross-dataset consolidation would broaden coverage of rare conditions and, by standardizing formats, also ease the cross-paradigm comparison currently hindered by heterogeneous benchmarks (Sec.~\ref{subsec:Quantitative Comparison}).

\noindent\textbf{Interoperable representations and generalization:} Standardized, object- or agent-centric latent interfaces could restore modularity within unified systems, allowing interchangeable submodules without breaking joint training, as hinted by PF-Track \cite{pang2023standing}, JLA \cite{kesa2022multiple}, and ViP3D \cite{gu2023vip3d}. Coupled with deeper (dynamic) map integration \cite{xu2024towards} and class-free or occupancy-based formulations, such interfaces would also improve robustness to rare object categories and heterogeneous sensors.

\noindent\textbf{Toward end-to-end and reasoning-guided perception:} Unified perception forms a natural bridge to end-to-end driving. Emerging BEV- and occupancy-centric representations, foundation models, and world models point toward systems that retain interpretable scene-understanding outputs while approaching the holistic reasoning of end-to-end stacks. Reasoning-based approaches in particular are on the rise within full end-to-end driving \cite{wang2025alpamayo}, increasingly supported by datasets enriched with reasoning traces \cite{tian2025nuscenes}, and open a promising avenue for unified perception. Explicit reasoning over the interdependent detection, tracking, and prediction sub-tasks could strengthen inter-task synergy while preserving interpretable outputs. The central frontier remains how far this extends without sacrificing verifiability.

\noindent\textbf{Safety, reproducibility, and verification:} Integrated designs complicate safety assurance, motivating online runtime monitoring and offline formal verification. Progress is further slowed by limited code release: closed-source contributions account for 31.1\% in EUP, 85.7\% in LUP, and 74.3\% in FUP, underscoring the need for open benchmarks and models.

\section{Conclusion}
\label{chap:Conclusion}

This survey establishes the first holistic taxonomy of unified perception, formalizing three paradigms, Early, Late, and Full Unified Perception, that extend modular scene understanding by jointly addressing detection, tracking, and prediction. Organized around tracking formulation, cue type, and representation flow, the taxonomy both classifies existing methods and frames the design space for new ones.

Positioning the surveyed approaches within this space surfaces a consistent trajectory: A shift from interpretable bounding-box interfaces and external association toward latent representations and in-network or implicit tracking, trading verifiability for joint-optimization depth. Persistent challenges, differentiable association, cross-modal generalization, standardized evaluation, and limited code release, cut across all three paradigms. We hope this survey equips the community to advance unified perception toward more robust, generalizable, and interpretable systems.

\section*{Acknowledgements}
As the first author, Loïc Stratil conceived the study and led its conception, analysis and writing. Felix Fent and Esteban Rivera contributed to conception and revision of the manuscript. Markus Lienkamp made an essential contribution to the conception of the research project and revised the article critically for important intellectual content. He gave final approval of the version to be published and agreed with all aspects of the work. As a guarantor, he accepts responsibility for the article's overall integrity. The authors would like to thank the Munich Institute of Robotics and Machine Intelligence (MIRMI) for their support.

\bibliographystyle{unsrt}
\bibliography{References/references_surveys, References/references_early_unified,  References/references_late_unified,  References/references_full_unified, References/references_other}

@article{chaabane2021deft,
  title={{Deft: Detection embeddings for tracking}},
  author={Chaabane, Mohamed and Zhang, Peter and Beveridge, J Ross and O'Hara, Stephen},
  journal={{arxiv preprint arXiv:2102.02267}},
  year={2021}
}

@inproceedings{meinhardt2022trackformer,
  title={{Trackformer: Multi-object tracking with transformers}},
  author={Meinhardt, Tim and Kirillov, Alexander and Leal-Taixe, Laura and Feichtenhofer, Christoph},
  booktitle={{Proceedings of the IEEE/CVF conference on computer vision and pattern recognition}},
  pages={8844--8854},
  year={2022}
}

@article{zhang2021fairmot,
  title={{Fairmot: On the fairness of detection and re-identification in multiple object tracking}},
  author={Zhang, Yifu and Wang, Chunyu and Wang, Xinggang and Zeng, Wenjun and Liu, Wenyu},
  journal={{International journal of computer vision}},
  volume={129},
  pages={3069--3087},
  year={2021},
  publisher={Springer}
}

@inproceedings{wang2021joint,
  title={{Joint object detection and multi-object tracking with graph neural networks}},
  author={Wang, Yongxin and Kitani, Kris and Weng, Xinshuo},
  booktitle={{2021 IEEE international conference on robotics and automation (ICRA)}},
  pages={13708--13715},
  year={2021},
  organization={IEEE}
}

@inproceedings{tokmakov2021learning,
  title={{Learning to track with object permanence}},
  author={Tokmakov, Pavel and Li, Jie and Burgard, Wolfram and Gaidon, Adrien},
  booktitle={{Proceedings of the IEEE/CVF International Conference on Computer Vision}},
  pages={10860--10869},
  year={2021}
}

@inproceedings{wu2021track,
  title={{Track to detect and segment: An online multi-object tracker}},
  author={Wu, Jialian and Cao, Jiale and Song, Liangchen and Wang, Yu and Yang, Ming and Yuan, Junsong},
  booktitle={{Proceedings of the IEEE/CVF conference on computer vision and pattern recognition}},
  pages={12352--12361},
  year={2021}
}

@inproceedings{yan2021anchor,
  title={{Anchor-free person search}},
  author={Yan, Yichao and Li, Jinpeng and Qin, Jie and Bai, Song and Liao, Shengcai and Liu, Li and Zhu, Fan and Shao, Ling},
  booktitle={{Proceedings of the IEEE/CVF conference on computer vision and pattern recognition}},
  pages={7690--7699},
  year={2021}
}

@article{xu2022transcenter,
  title={{TransCenter: Transformers with dense representations for multiple-object tracking}},
  author={Xu, Yihong and Ban, Yutong and Delorme, Guillaume and Gan, Chuang and Rus, Daniela and Alameda-Pineda, Xavier},
  journal={{IEEE transactions on pattern analysis and machine intelligence}},
  volume={45},
  number={6},
  pages={7820--7835},
  year={2022},
  publisher={IEEE}
}

@inproceedings{lu2020retinatrack,
  title={{Retinatrack: Online single stage joint detection and tracking}},
  author={Lu, Zhichao and Rathod, Vivek and Votel, Ronny and Huang, Jonathan},
  booktitle={{Proceedings of the IEEE/CVF conference on computer vision and pattern recognition}},
  pages={14668--14678},
  year={2020}
}

@article{shuai2020multi,
  title={{Multi-object tracking with siamese track-rcnn}},
  author={Shuai, Bing and Berneshawi, Andrew G and Modolo, Davide and Tighe, Joseph},
  journal={{arXiv preprint arXiv:2004.07786}},
  year={2020}
}

@inproceedings{cai2022memot,
  title={{Memot: Multi-object tracking with memory}},
  author={Cai, Jiarui and Xu, Mingze and Li, Wei and Xiong, Yuanjun and Xia, Wei and Tu, Zhuowen and Soatto, Stefano},
  booktitle={{Proceedings of the IEEE/CVF conference on computer vision and pattern recognition}},
  pages={8090--8100},
  year={2022}
}

@inproceedings{pang2020tubetk,
  title={{Tubetk: Adopting tubes to track multi-object in a one-step training model}},
  author={Pang, Bo and Li, Yizhuo and Zhang, Yifan and Li, Muchen and Lu, Cewu},
  booktitle={{Proceedings of the IEEE/CVF conference on computer vision and pattern recognition}},
  pages={6308--6318},
  year={2020}
}

@inproceedings{peng2020chained,
  title={{Chained-tracker: Chaining paired attentive regression results for end-to-end joint multiple-object detection and tracking}},
  author={Peng, Jinlong and Wang, Changan and Wan, Fangbin and Wu, Yang and Wang, Yabiao and Tai, Ying and Wang, Chengjie and Li, Jilin and Huang, Feiyue and Fu, Yanwei},
  booktitle={{Computer Vision--ECCV 2020: 16th European Conference, Glasgow, UK, August 23--28, 2020, Proceedings, Part IV 16}},
  pages={145--161},
  year={2020},
  organization={Springer}
}

@inproceedings{wang2020towards,
  title={{Towards real-time multi-object tracking}},
  author={Wang, Zhongdao and Zheng, Liang and Liu, Yixuan and Li, Yali and Wang, Shengjin},
  booktitle={{European conference on computer vision}},
  pages={107--122},
  year={2020},
  organization={Springer}
}

@inproceedings{zeng2022motr,
  title={{Motr: End-to-end multiple-object tracking with transformer}},
  author={Zeng, Fangao and Dong, Bin and Zhang, Yuang and Wang, Tiancai and Zhang, Xiangyu and Wei, Yichen},
  booktitle={{European conference on computer vision}},
  pages={659--675},
  year={2022},
  organization={Springer}
}

@inproceedings{zhang2023motrv2,
  title={{Motrv2: Bootstrapping end-to-end multi-object tracking by pretrained object detectors}},
  author={Zhang, Yuang and Wang, Tiancai and Zhang, Xiangyu},
  booktitle={{Proceedings of the IEEE/CVF conference on computer vision and pattern recognition}},
  pages={22056--22065},
  year={2023}
}

@article{yu2023motrv3,
  title={{Motrv3: Release-fetch supervision for end-to-end multi-object tracking}},
  author={Yu, En and Wang, Tiancai and Li, Zhuoling and Zhang, Yuang and Zhang, Xiangyu and Tao, Wenbing},
  journal={{arXiv preprint arXiv:2305.14298}},
  year={2023}
}

@inproceedings{wang2021multiple,
  title={{Multiple object tracking with correlation learning}},
  author={Wang, Qiang and Zheng, Yun and Pan, Pan and Xu, Yinghui},
  booktitle={{Proceedings of the IEEE/CVF conference on computer vision and pattern recognition}},
  pages={3876--3886},
  year={2021}
}

@article{sun2020transtrack,
  title={{Transtrack: Multiple object tracking with transformer}},
  author={Sun, Peize and Cao, Jinkun and Jiang, Yi and Zhang, Rufeng and Xie, Enze and Yuan, Zehuan and Wang, Changhu and Luo, Ping},
  journal={{arXiv preprint arXiv:2012.15460}},
  year={2020}
}

@inproceedings{li2022time3d,
  title={{Time3d: End-to-end joint monocular 3d object detection and tracking for autonomous driving}},
  author={Li, Peixuan and Jin, Jieyu},
  booktitle={{Proceedings of the IEEE/CVF conference on computer vision and pattern recognition}},
  pages={3885--3894},
  year={2022}
}

@inproceedings{zhang2022mutr3d,
  title={{Mutr3d: A multi-camera tracking framework via 3d-to-2d queries}},
  author={Zhang, Tianyuan and Chen, Xuanyao and Wang, Yue and Wang, Yilun and Zhao, Hang},
  booktitle={{Proceedings of the IEEE/CVF Conference on Computer Vision and Pattern Recognition}},
  pages={4537--4546},
  year={2022}
}

@inproceedings{wang2024onetrack,
  title={{OneTrack: Demystifying the Conflict Between Detection and Tracking in End-to-End 3D Trackers}},
  author={Wang, Qitai and He, Jiawei and Chen, Yuntao and Zhang, Zhaoxiang},
  booktitle={{European Conference on Computer Vision}},
  pages={387--404},
  year={2024},
  organization={Springer}
}

@inproceedings{li2023end,
  title={{End-to-end 3d tracking with decoupled queries}},
  author={Li, Yanwei and Yu, Zhiding and Philion, Jonah and Anandkumar, Anima and Fidler, Sanja and Jia, Jiaya and Alvarez, Jose},
  booktitle={{Proceedings of the IEEE/CVF International Conference on Computer Vision}},
  pages={18302--18311},
  year={2023}
}

@article{yan2023bridging,
  title={{Bridging the gap between end-to-end and non-end-to-end multi-object tracking}},
  author={Yan, Feng and Luo, Weixin and Zhong, Yujie and Gan, Yiyang and Ma, Lin},
  journal={{arXiv preprint arXiv:2305.12724}},
  year={2023}
}

@article{jia2024multi,
  title={{Multi-object Tracking by Detection and Query: an efficient end-to-end manner}},
  author={Jia, Shukun and Cao, Yichao and Yang, Feng and Lu, Xin and Lu, Xiaobo},
  journal={{arXiv preprint arXiv:2411.06197}},
  year={2024}
}

@inproceedings{gao2023memotr,
  title={{MeMOTR: Long-term memory-augmented transformer for multi-object tracking}},
  author={Gao, Ruopeng and Wang, Limin},
  booktitle={{Proceedings of the IEEE/CVF International Conference on Computer Vision}},
  pages={9901--9910},
  year={2023}
}

@inproceedings{segu2024samba,
  title = {{Samba: Synchronized Set-of-Sequences Modeling for Multiple Object Tracking}},
  author = {Segu, Mattia and Piccinelli, Luigi and Li, Siyuan and Yang, Yung-Hsu and Van Gool, Luc and Schiele, Bernt},
  booktitle = {International Conference on Learning Representations},
  editor = {Y. Yue and A. Garg and N. Peng and F. Sha and R. Yu},
  pages = {30057--30070},
  volume = {2025},
  year = {2025}
}

@inproceedings{ding2024ada,
  title={{Ada-track: End-to-end multi-camera 3d multi-object tracking with alternating detection and association}},
  author={Ding, Shuxiao and Schneider, Lukas and Cordts, Marius and Gall, Juergen},
  booktitle={{Proceedings of the IEEE/CVF Conference on Computer Vision and Pattern Recognition}},
  pages={15184--15194},
  year={2024}
}

@article{doll2023star,
  title={{Star-track: Latent motion models for end-to-end 3d object tracking with adaptive spatio-temporal appearance representations}},
  author={Doll, Simon and Hanselmann, Niklas and Schneider, Lukas and Schulz, Richard and Enzweiler, Markus and Lensch, Hendrik PA},
  journal={{IEEE Robotics and Automation Letters}},
  volume={9},
  number={2},
  pages={1326--1333},
  year={2023},
  publisher={IEEE}
}

@inproceedings{zhou2020tracking,
  title={{Tracking objects as points}},
  author={Zhou, Xingyi and Koltun, Vladlen and Kr{\"a}henb{\"u}hl, Philipp},
  booktitle={{European conference on computer vision}},
  pages={474--490},
  year={2020},
  organization={Springer}
}

@article{hu2022monocular,
  title={{Monocular quasi-dense 3d object tracking}},
  author={Hu, Hou-Ning and Yang, Yung-Hsu and Fischer, Tobias and Darrell, Trevor and Yu, Fisher and Sun, Min},
  journal={{IEEE Transactions on Pattern Analysis and Machine Intelligence}},
  volume={45},
  number={2},
  pages={1992--2008},
  year={2022},
  publisher={IEEE}
}

@inproceedings{zhao2022tracking,
  title={{Tracking objects as pixel-wise distributions}},
  author={Zhao, Zelin and Wu, Ze and Zhuang, Yueqing and Li, Boxun and Jia, Jiaya},
  booktitle={{European Conference on Computer Vision}},
  pages={76--94},
  year={2022},
  organization={Springer}
}

@inproceedings{niculescu2025mctr,
  title={{MCTR: Multi Camera Tracking Transformer}},
  author={Niculescu-Mizil, Alexandru and Patel, Deep and Melvin, Iain},
  booktitle={{Proceedings of the Winter Conference on Applications of Computer Vision}},
  pages={874--884},
  year={2025}
}

@inproceedings{yin2021center,
  title={{Center-based 3d object detection and tracking}},
  author={Yin, Tianwei and Zhou, Xingyi and Krahenbuhl, Philipp},
  booktitle={{Proceedings of the IEEE/CVF conference on computer vision and pattern recognition}},
  pages={11784--11793},
  year={2021}
}

@inproceedings{chen2023voxelnext,
  title={{Voxelnext: Fully sparse voxelnet for 3d object detection and tracking}},
  author={Chen, Yukang and Liu, Jianhui and Zhang, Xiangyu and Qi, Xiaojuan and Jia, Jiaya},
  booktitle={{Proceedings of the IEEE/CVF Conference on Computer Vision and Pattern Recognition}},
  pages={21674--21683},
  year={2023}
}

@article{liu2023centertube,
  title={{CenterTube: Tracking multiple 3D objects with 4D tubelets in dynamic point clouds}},
  author={Liu, Hao and Ma, Yanni and Hu, Qingyong and Guo, Yulan},
  journal={{IEEE Transactions on Multimedia}},
  volume={25},
  pages={8793--8804},
  year={2023},
  publisher={IEEE}
}

@inproceedings{luo2021exploring,
  title={{Exploring simple 3d multi-object tracking for autonomous driving}},
  author={Luo, Chenxu and Yang, Xiaodong and Yuille, Alan},
  booktitle={{Proceedings of the IEEE/CVF international conference on computer vision}},
  pages={10488--10497},
  year={2021}
}

@article{gwak2022minkowski,
  title={{Minkowski tracker: A sparse spatio-temporal r-cnn for joint object detection and tracking}},
  author={Gwak, JunYoung and Savarese, Silvio and Bohg, Jeannette},
  journal={{arXiv preprint arXiv:2208.10056}},
  year={2022}
}

@inproceedings{cheong2024jdt3d,
  title={{JDT3D: Addressing the Gaps in LiDAR-Based Tracking-by-Attention}},
  author={Cheong, Brian and Zhou, Jiachen and Waslander, Steven},
  booktitle={{European Conference on Computer Vision}},
  pages={161--177},
  year={2024},
  organization={Springer}
}

@inproceedings{yilmaz2024mask4former,
  title={{Mask4former: Mask transformer for 4d panoptic segmentation}},
  author={Yilmaz, Kadir and Schult, Jonas and Nekrasov, Alexey and Leibe, Bastian},
  booktitle={{2024 IEEE International Conference on Robotics and Automation (ICRA)}},
  pages={9418--9425},
  year={2024},
  organization={IEEE}
}

@inproceedings{huang2021joint,
  title={{Joint multi-object detection and tracking with camera-LiDAR fusion for autonomous driving}},
  author={Huang, Kemiao and Hao, Qi},
  booktitle={{2021 IEEE/RSJ International Conference on Intelligent Robots and Systems (IROS)}},
  pages={6983--6989},
  year={2021},
  organization={IEEE}
}

@inproceedings{zeng2021cross,
  title={{Cross-modal 3d object detection and tracking for auto-driving}},
  author={Zeng, Yihan and Ma, Chao and Zhu, Ming and Fan, Zhiming and Yang, Xiaokang},
  booktitle={{2021 IEEE/RSJ International Conference on Intelligent Robots and Systems (IROS)}},
  pages={3850--3857},
  year={2021},
  organization={IEEE}
}

@inproceedings{koh2022joint,
  title={{Joint 3d object detection and tracking using spatio-temporal representation of camera image and lidar point clouds}},
  author={Koh, Junho and Kim, Jaekyum and Yoo, Jin Hyeok and Kim, Yecheol and Kum, Dongsuk and Choi, Jun Won},
  booktitle={{Proceedings of the AAAI Conference on Artificial Intelligence}},
  volume={36},
  pages={1210--1218},
  year={2022}
}

@inproceedings{qin2023motiontrack,
  title={{Motiontrack: Learning robust short-term and long-term motions for multi-object tracking}},
  author={Qin, Zheng and Zhou, Sanping and Wang, Le and Duan, Jinghai and Hua, Gang and Tang, Wei},
  booktitle={{Proceedings of the IEEE/CVF conference on computer vision and pattern recognition}},
  pages={17939--17948},
  year={2023}
}

@article{wang2024multi,
  title={{A Multi-Modal Fusion-Based 3D Multi-Object Tracking Framework with Joint Detection}},
  author={Wang, Xiyang and Fu, Chunyun and He, Jiawei and Huang, Mingguang and Meng, Ting and Zhang, Siyu and Zhou, Hangning and Xu, Ziyao and Zhang, Chi},
  journal={{IEEE Robotics and Automation Letters}},
  year={2024},
  publisher={IEEE}
}

@inproceedings{luo2018fast,
  title={{Fast and furious: Real time end-to-end 3d detection, tracking and motion forecasting with a single convolutional net}},
  author={Luo, Wenjie and Yang, Bin and Urtasun, Raquel},
  booktitle={{Proceedings of the IEEE conference on Computer Vision and Pattern Recognition}},
  pages={3569--3577},
  year={2018}
}

@inproceedings{casas2018intentnet,
  title={{Intentnet: Learning to predict intention from raw sensor data}},
  author={Casas, Sergio and Luo, Wenjie and Urtasun, Raquel},
  booktitle={{Conference on Robot Learning}},
  pages={947--956},
  year={2018},
  organization={PMLR}
}

@inproceedings{casas2020spagnn,
  title={{Spagnn: Spatially-aware graph neural networks for relational behavior forecasting from sensor data}},
  author={Casas, Sergio and Gulino, Cole and Liao, Renjie and Urtasun, Raquel},
  booktitle={{2020 IEEE International Conference on Robotics and Automation (ICRA)}},
  pages={9491--9497},
  year={2020},
  organization={IEEE}
}

@inproceedings{liang2020pnpnet,
  title={{Pnpnet: End-to-end perception and prediction with tracking in the loop}},
  author={Liang, Ming and Yang, Bin and Zeng, Wenyuan and Chen, Yun and Hu, Rui and Casas, Sergio and Urtasun, Raquel},
  booktitle={{Proceedings of the IEEE/CVF Conference on Computer Vision and Pattern Recognition}},
  pages={11553--11562},
  year={2020}
}

@inproceedings{cheng2023end,
  title={{An end-to-end framework of road user detection, tracking, and prediction from monocular images}},
  author={Cheng, Hao and Liu, Mengmeng and Chen, Lin},
  booktitle={{2023 IEEE 26th International Conference on Intelligent Transportation Systems (ITSC)}},
  pages={2178--2185},
  year={2023},
  organization={IEEE}
}

@inproceedings{zhang2020stinet,
  title={{Stinet: Spatio-temporal-interactive network for pedestrian detection and trajectory prediction}},
  author={Zhang, Zhishuai and Gao, Jiyang and Mao, Junhua and Liu, Yukai and Anguelov, Dragomir and Li, Congcong},
  booktitle={{Proceedings of the IEEE/CVF Conference on Computer Vision and Pattern Recognition}},
  pages={11346--11355},
  year={2020}
}

@inproceedings{casas2020implicit,
  title={{Implicit latent variable model for scene-consistent motion forecasting}},
  author={Casas, Sergio and Gulino, Cole and Suo, Simon and Luo, Katie and Liao, Renjie and Urtasun, Raquel},
  booktitle={{Computer Vision--ECCV 2020: 16th European Conference, Glasgow, UK, August 23--28, 2020, Proceedings, Part XXIII 16}},
  pages={624--641},
  year={2020},
  organization={Springer}
}

@inproceedings{casas2020importance,
  title={{The importance of prior knowledge in precise multimodal prediction}},
  author={Casas, Sergio and Gulino, Cole and Suo, Simon and Urtasun, Raquel},
  booktitle={{2020 IEEE/RSJ international conference on intelligent robots and systems (IROS)}},
  pages={2295--2302},
  year={2020},
  organization={IEEE}
}

@inproceedings{li2020end,
  title={{End-to-end contextual perception and prediction with interaction transformer}},
  author={Li, Lingyun Luke and Yang, Bin and Liang, Ming and Zeng, Wenyuan and Ren, Mengye and Segal, Sean and Urtasun, Raquel},
  booktitle={{2020 IEEE/RSJ International Conference on Intelligent Robots and Systems (IROS)}},
  pages={5784--5791},
  year={2020},
  organization={IEEE}
}

@inproceedings{weng2021inverting,
  title={{Inverting the pose forecasting pipeline with SPF2: Sequential pointcloud forecasting for sequential pose forecasting}},
  author={Weng, Xinshuo and Wang, Jianren and Levine, Sergey and Kitani, Kris and Rhinehart, Nicholas},
  booktitle={{Conference on robot learning}},
  pages={11--20},
  year={2021},
  organization={PMLR}
}

@inproceedings{shah2020liranet,
  title={{LiRaNet: End-to-End Trajectory Prediction using Spatio-Temporal Radar Fusion}},
  author={Shah, Meet and Huang, Zhiling and Laddha, Ankit and Langford, Matthew and Barber, Blake and Zhang, Sida and Vallespi-Gonzalez, Carlos and Urtasun, Raquel},
  booktitle={Proceedings of the 2020 Conference on Robot Learning},
  pages={31--48},
  year={2021},
  editor={Kober, Jens and Ramos, Fabio and Tomlin, Claire},
  volume={155},
  series={Proceedings of Machine Learning Research},
  month={16--18 Nov},
  publisher={PMLR}
}

@inproceedings{zhang2020sdp,
  title={{Sdp-net: Scene flow based real-time object detection and prediction from sequential 3d point clouds}},
  author={Zhang, Yi and Ye, Yuwen and Xiang, Zhiyu and Gu, Jiaqi},
  booktitle={{Proceedings of the Asian Conference on Computer Vision}},
  year={2020}
}

@article{meyer2020laserflow,
  title={{Laserflow: Efficient and probabilistic object detection and motion forecasting}},
  author={Meyer, Gregory P and Charland, Jake and Pandey, Shreyash and Laddha, Ankit and Gautam, Shivam and Vallespi-Gonzalez, Carlos and Wellington, Carl K},
  journal={{IEEE Robotics and Automation Letters}},
  volume={6},
  number={2},
  pages={526--533},
  year={2020},
  publisher={IEEE}
}

@inproceedings{laddha2021mvfusenet,
  title={{Mvfusenet: Improving end-to-end object detection and motion forecasting through multi-view fusion of lidar data}},
  author={Laddha, Ankit and Gautam, Shivam and Palombo, Stefan and Pandey, Shreyash and Vallespi-Gonzalez, Carlos},
  booktitle={{Proceedings of the IEEE/CVF Conference on Computer Vision and Pattern Recognition}},
  pages={2865--2874},
  year={2021}
}

@inproceedings{djuric2021multixnet,
  title={{Multixnet: Multiclass multistage multimodal motion prediction}},
  author={Djuric, Nemanja and Cui, Henggang and Su, Zhaoen and Wu, Shangxuan and Wang, Huahua and Chou, Fang-Chieh and San Martin, Luisa and Feng, Song and Hu, Rui and Xu, Yang and others},
  booktitle={{2021 IEEE Intelligent Vehicles Symposium (IV)}},
  pages={435--442},
  year={2021},
  organization={IEEE}
}

@inproceedings{ye2021sdapnet,
  title={{Sdapnet: End-to-end multi-task simultaneous detection and prediction network}},
  author={Ye, Shanding and Yao, Han and Wang, Wenfu and Fu, Yongjian and Pan, Zhijie},
  booktitle={{2021 International Joint Conference on Neural Networks (IJCNN)}},
  pages={1--8},
  year={2021},
  organization={IEEE}
}

@inproceedings{laddha2021rv,
  title={{Rv-fusenet: Range view based fusion of time-series lidar data for joint 3d object detection and motion forecasting}},
  author={Laddha, Ankit and Gautam, Shivam and Meyer, Gregory P and Vallespi-Gonzalez, Carlos and Wellington, Carl K},
  booktitle={{2021 IEEE/RSJ International Conference on Intelligent Robots and Systems (IROS)}},
  pages={7060--7066},
  year={2021},
  organization={IEEE}
}

@inproceedings{hu2021fiery,
  title={{Fiery: Future instance prediction in bird's-eye view from surround monocular cameras}},
  author={Hu, Anthony and Murez, Zak and Mohan, Nikhil and Dudas, Sof{\'\i}a and Hawke, Jeffrey and Badrinarayanan, Vijay and Cipolla, Roberto and Kendall, Alex},
  booktitle={{Proceedings of the IEEE/CVF International Conference on Computer Vision}},
  pages={15273--15282},
  year={2021}
}

@inproceedings{fadadu2022multi,
  title={{Multi-view fusion of sensor data for improved perception and prediction in autonomous driving}},
  author={Fadadu, Sudeep and Pandey, Shreyash and Hegde, Darshan and Shi, Yi and Chou, Fang-Chieh and Djuric, Nemanja and Vallespi-Gonzalez, Carlos},
  booktitle={{Proceedings of the IEEE/CVF Winter Conference on Applications of Computer Vision}},
  pages={2349--2357},
  year={2022}
}

@article{jiang2022perceive,
  title={{Perceive, interact, predict: Learning dynamic and static clues for end-to-end motion prediction}},
  author={Jiang, Bo and Chen, Shaoyu and Wang, Xinggang and Liao, Bencheng and Cheng, Tianheng and Chen, Jiajie and Zhou, Helong and Zhang, Qian and Liu, Wenyu and Huang, Chang},
  journal={{arXiv preprint arXiv:2212.02181}},
  year={2022}
}

@inproceedings{gu2023vip3d,
  title={{Vip3d: End-to-end visual trajectory prediction via 3d agent queries}},
  author={Gu, Junru and Hu, Chenxu and Zhang, Tianyuan and Chen, Xuanyao and Wang, Yilun and Wang, Yue and Zhao, Hang},
  booktitle={{Proceedings of the IEEE/CVF Conference on Computer Vision and Pattern Recognition}},
  pages={5496--5506},
  year={2023}
}

@inproceedings{casas2024detra,
  title={{Detra: A unified model for object detection and trajectory forecasting}},
  author={Casas, Sergio and Agro, Ben and Mao, Jiageng and Gilles, Thomas and Cui, Alexander and Li, Thomas and Urtasun, Raquel},
  booktitle={{European Conference on Computer Vision}},
  pages={326--342},
  year={2024},
  organization={Springer}
}

@inproceedings{chen2022fs,
  title={{Fs-gru: Continuous perception and prediction with inter frame feature sharing}},
  author={Chen, Zhikai and Wang, Yafei and Liu, Xulei and Wang, Xinchang},
  booktitle={{2022 IEEE 25th International Conference on Intelligent Transportation Systems (ITSC)}},
  pages={517--522},
  year={2022},
  organization={IEEE}
}

@inproceedings{chen2022s2f2,
  title={{S2F2: single-stage flow forecasting for future multiple trajectories prediction}},
  author={Chen, Yu-Wen and Yang, Hsuan-Kung and Chiu, Chu-Chi and Lee, Chun-Yi},
  booktitle={{Proceedings of the IEEE/CVF conference on computer vision and pattern recognition}},
  pages={2536--2539},
  year={2022}
}

@article{hendy2020fishing,
  title={{Fishing net: Future inference of semantic heatmaps in grids}},
  author={Hendy, Noureldin and Sloan, Cooper and Tian, Feng and Duan, Pengfei and Charchut, Nick and Xie, Yuesong and Wang, Chuang and Philbin, James},
  journal={{arXiv preprint arXiv:2006.09917}},
  year={2020}
}

@inproceedings{luo2021safety,
  title={{Safety-oriented pedestrian occupancy forecasting}},
  author={Luo, Katie and Casas, Sergio and Liao, Renjie and Yan, Xinchen and Xiong, Yuwen and Zeng, Wenyuan and Urtasun, Raquel},
  booktitle={{2021 IEEE/RSJ International Conference on Intelligent Robots and Systems (IROS)}},
  pages={1015--1022},
  year={2021},
  organization={IEEE}
}

@article{zhang2022beverse,
  title={{Beverse: Unified perception and prediction in birds-eye-view for vision-centric autonomous driving}},
  author={Zhang, Yunpeng and Zhu, Zheng and Zheng, Wenzhao and Huang, Junjie and Huang, Guan and Zhou, Jie and Lu, Jiwen},
  journal={{arXiv preprint arXiv:2205.09743}},
  year={2022}
}

@inproceedings{peri2022forecasting,
  title={{Forecasting from lidar via future object detection}},
  author={Peri, Neehar and Luiten, Jonathon and Li, Mengtian and O{\v{s}}ep, Aljo{\v{s}}a and Leal-Taix{\'e}, Laura and Ramanan, Deva},
  booktitle={{Proceedings of the IEEE/CVF conference on computer vision and pattern recognition}},
  pages={17202--17211},
  year={2022}
}

@inproceedings{akan2022stretchbev,
  title={{Stretchbev: Stretching future instance prediction spatially and temporally}},
  author={Akan, Adil Kaan and G{\"u}ney, Fatma},
  booktitle={{European Conference on Computer Vision}},
  pages={444--460},
  year={2022},
  organization={Springer}
}

@inproceedings{fang2023tbp,
  title={{Tbp-former: Learning temporal bird's-eye-view pyramid for joint perception and prediction in vision-centric autonomous driving}},
  author={Fang, Shaoheng and Wang, Zi and Zhong, Yiqi and Ge, Junhao and Chen, Siheng},
  booktitle={{Proceedings of the IEEE/CVF conference on computer vision and pattern recognition}},
  pages={1368--1378},
  year={2023}
}

@inproceedings{agro2023implicit,
  title={{Implicit occupancy flow fields for perception and prediction in self-driving}},
  author={Agro, Ben and Sykora, Quinlan and Casas, Sergio and Urtasun, Raquel},
  booktitle={{Proceedings of the IEEE/CVF conference on computer vision and pattern recognition}},
  pages={1379--1388},
  year={2023}
}

@inproceedings{khurana2023point,
  title={{Point cloud forecasting as a proxy for 4d occupancy forecasting}},
  author={Khurana, Tarasha and Hu, Peiyun and Held, David and Ramanan, Deva},
  booktitle={{Proceedings of the IEEE/CVF Conference on Computer Vision and Pattern Recognition}},
  pages={1116--1124},
  year={2023}
}

@inproceedings{li2023powerbev,
  title = {{PowerBEV: A Powerful Yet Lightweight Framework for Instance Prediction in Bird’s-Eye View}},
  booktitle = {Proceedings of the Thirty-Second International Joint Conference on Artificial Intelligence},
  publisher = {International Joint Conferences on Artificial Intelligence Organization},
  author = {Li,  Peizheng and Ding,  Shuxiao and Chen,  Xieyuanli and Hanselmann,  Niklas and Cordts,  Marius and Gall,  Juergen},
  year = {2023},
  month = Aug,
  pages = {1080--1088},
  series = {IJCAI-2023},
  collection = {IJCAI-2023}
}

@inproceedings{liu2024lidar,
  title={{Lidar-based 4d occupancy completion and forecasting}},
  author={Liu, Xinhao and Gong, Moonjun and Fang, Qi and Xie, Haoyu and Li, Yiming and Zhao, Hang and Feng, Chen},
  booktitle={{2024 IEEE/RSJ International Conference on Intelligent Robots and Systems (IROS)}},
  pages={11102--11109},
  year={2024},
  organization={IEEE}
}

@inproceedings{khurana2022differentiable,
  title={{Differentiable raycasting for self-supervised occupancy forecasting}},
  author={Khurana, Tarasha and Hu, Peiyun and Dave, Achal and Ziglar, Jason and Held, David and Ramanan, Deva},
  booktitle={{European Conference on Computer Vision}},
  pages={353--369},
  year={2022},
  organization={Springer}
}

@inproceedings{cui2021lookout,
  title={{Lookout: Diverse multi-future prediction and planning for self-driving}},
  author={Cui, Alexander and Casas, Sergio and Sadat, Abbas and Liao, Renjie and Urtasun, Raquel},
  booktitle={{Proceedings of the IEEE/CVF International Conference on Computer Vision}},
  pages={16107--16116   },
  year={2021}
}

@inproceedings{sadat2020perceive,
  title={{Perceive, predict, and plan: Safe motion planning through interpretable semantic representations}},
  author={Sadat, Abbas and Casas, Sergio and Ren, Mengye and Wu, Xinyu and Dhawan, Pranaab and Urtasun, Raquel},
  booktitle={{Computer Vision--ECCV 2020: 16th European Conference, Glasgow, UK, August 23--28, 2020, Proceedings, Part XXIII 16}},
  pages={414--430},
  year={2020},
  organization={Springer}
}

@inproceedings{zeng2019end,
  title={{End-to-end interpretable neural motion planner}},
  author={Zeng, Wenyuan and Luo, Wenjie and Suo, Simon and Sadat, Abbas and Yang, Bin and Casas, Sergio and Urtasun, Raquel},
  booktitle={{Proceedings of the IEEE/CVF conference on computer vision and pattern recognition}},
  pages={8660--8669},
  year={2019}
}

@inproceedings{casas2021mp3,
  title={{Mp3: A unified model to map, perceive, predict and plan}},
  author={Casas, Sergio and Sadat, Abbas and Urtasun, Raquel},
  booktitle={{Proceedings of the IEEE/CVF Conference on Computer Vision and Pattern Recognition}},
  pages={14403--14412},
  year={2021}
}

@inproceedings{hu2022st,
  title={{St-p3: End-to-end vision-based autonomous driving via spatial-temporal feature learning}},
  author={Hu, Shengchao and Chen, Li and Wu, Penghao and Li, Hongyang and Yan, Junchi and Tao, Dacheng},
  booktitle={{European Conference on Computer Vision}},
  pages={533--549},
  year={2022},
  organization={Springer}
}

@article{ye2023fusionad,
  title={{Fusionad: Multi-modality fusion for prediction and planning tasks of autonomous driving}},
  author={Ye, Tengju and Jing, Wei and Hu, Chunyong and Huang, Shikun and Gao, Lingping and Li, Fangzhen and Wang, Jingke and Guo, Ke and Xiao, Wencong and Mao, Weibo and others},
  journal={{arXiv preprint arXiv:2308.01006}},
  year={2023}
}

@inproceedings{jiang2023vad,
  title={{Vad: Vectorized scene representation for efficient autonomous driving}},
  author={Jiang, Bo and Chen, Shaoyu and Xu, Qing and Liao, Bencheng and Chen, Jiajie and Zhou, Helong and Zhang, Qian and Liu, Wenyu and Huang, Chang and Wang, Xinggang},
  booktitle={{Proceedings of the IEEE/CVF International Conference on Computer Vision}},
  pages={8340--8350},
  year={2023}
}

@inproceedings{pang2023standing,
  title={{Standing between past and future: Spatio-temporal modeling for multi-camera 3d multi-object tracking}},
  author={Pang, Ziqi and Li, Jie and Tokmakov, Pavel and Chen, Dian and Zagoruyko, Sergey and Wang, Yu-Xiong},
  booktitle={{Proceedings of the IEEE/CVF conference on computer vision and pattern recognition}},
  pages={17928--17938},
  year={2023}
}

@inproceedings{weng2022whose,
  title={{Whose track is it anyway? Improving robustness to tracking errors with affinity-based trajectory prediction}},
  author={Weng, Xinshuo and Ivanovic, Boris and Kitani, Kris and Pavone, Marco},
  booktitle={{Proceedings of the IEEE/CVF Conference on Computer Vision and Pattern Recognition}},
  pages={6573--6582},
  year={2022}
}

@article{zhang2023towards,
  title={{Towards trajectory forecasting from detection}},
  author={Zhang, Pu and Bai, Lei and Wang, Yuning and Fang, Jianwu and Xue, Jianru and Zheng, Nanning and Ouyang, Wanli},
  journal={{IEEE Transactions on Pattern Analysis and Machine Intelligence}},
  volume={45},
  number={10},
  pages={12550--12561},
  year={2023},
  publisher={IEEE}
}

@inproceedings{weng2022mtp,
  title={{Mtp: Multi-hypothesis tracking and prediction for reduced error propagation}},
  author={Weng, Xinshuo and Ivanovic, Boris and Pavone, Marco},
  booktitle={{2022 IEEE Intelligent Vehicles Symposium (IV)}},
  pages={1218--1225},
  year={2022},
  organization={IEEE}
}

@article{weng2021ptp,
  title={{PTP: Parallelized tracking and prediction with graph neural networks and diversity sampling}},
  author={Weng, Xinshuo and Yuan, Ye and Kitani, Kris},
  journal={{IEEE Robotics and Automation Letters}},
  volume={6},
  number={3},
  pages={4640--4647},
  year={2021},
  publisher={IEEE}
}

@inproceedings{zhuang2024streammotp,
  title={{StreamMOTP: Streaming and unified framework for joint 3D multi-object tracking and trajectory prediction}},
  author={Zhuang, Jiaheng and Wang, Guoan and Zhang, Siyu and Wang, Xiyang and Zhou, Hangning and Xu, Ziyao and Zhang, Chi and Li, Zhiheng},
  booktitle={{Proceedings of the Asian Conference on Computer Vision}},
  pages={3189--3205},
  year={2024}
}

@inproceedings{kesa2022multiple,
  title={{Multiple object tracking and forecasting: Jointly predicting current and future object locations}},
  author={Kesa, Oluwafunmilola and Styles, Olly and Sanchez, Victor},
  booktitle={{Proceedings of the IEEE/CVF Winter Conference on Applications of Computer Vision}},
  pages={560--569},
  year={2022}
}

@inproceedings{uhlemannexploring,
  title = {{Exploring Shared Gaussian Occupancies for Tracking-Free,  Scene-Centric Pedestrian Motion Prediction in Autonomous Driving}},
  booktitle = {Proceedings of the 11th International Conference on Vehicle Technology and Intelligent Transport Systems},
  publisher = {SCITEPRESS - Science and Technology Publications},
  author = {Uhlemann,  Nico and W\"{o}rdehoff,  Melina and Lienkamp,  Markus},
  year = {2025},
  pages = {100--112}
}

@inproceedings{kato2018autoware,
  title={{Autoware on board: Enabling autonomous vehicles with embedded systems}},
  author={Kato, Shinpei and Tokunaga, Shota and Maruyama, Yuya and Maeda, Seiya and Hirabayashi, Manato and Kitsukawa, Yuki and Monrroy, Abraham and Ando, Tomohito and Fujii, Yusuke and Azumi, Takuya},
  booktitle={{2018 ACM/IEEE 9th International Conference on Cyber-Physical Systems (ICCPS)}},
  pages={287--296},
  year={2018},
  organization={IEEE}
}

@misc{apollo2017,
  author = {Baidu Apollo Team},
  title = {{Apollo: Open source autonomous driving platform}},
  year = {2017},
  howpublished = {\url{https://www.apollo.auto/apollo-self-driving}},
  note = {Accessed: April 24, 2025}
}

@inproceedings{marcu2023lingoqa,
  title = {{LingoQA: Visual Question Answering for Autonomous Driving}},
  booktitle = {Computer Vision - ECCV 2024},
  publisher = {Springer Nature Switzerland},
  author = {Marcu,  Ana-Maria and Chen,  Long and H\"{u}nermann,  Jan and Karnsund,  Alice and Hanotte,  Benoit and Chidananda,  Prajwal and Nair,  Saurabh and Badrinarayanan,  Vijay and Kendall,  Alex and Shotton,  Jamie and Arani,  Elahe and Sinavski,  Oleg},
  year = {2024},
  pages = {252--269}
}

@article{li2024hydra,
  title={{Hydra-mdp: End-to-end multimodal planning with multi-target hydra-distillation}},
  author={Li, Zhenxin and Li, Kailin and Wang, Shihao and Lan, Shiyi and Yu, Zhiding and Ji, Yishen and Li, Zhiqi and Zhu, Ziyue and Kautz, Jan and Wu, Zuxuan and others},
  journal={{arXiv preprint arXiv:2406.06978}  },
  year={2024}
}

@inproceedings{hu2023planning,
  title={{Planning-oriented autonomous driving}},
  author={Hu, Yihan and Yang, Jiazhi and Chen, Li and Li, Keyu and Sima, Chonghao and Zhu, Xizhou and Chai, Siqi and Du, Senyao and Lin, Tianwei and Wang, Wenhai and others},
  booktitle={{Proceedings of the IEEE/CVF conference on computer vision and pattern recognition}},
  pages={17853--17862},
  year={2023}
}

@inproceedings{philion2020lift,
  title={{Lift, splat, shoot: Encoding images from arbitrary camera rigs by implicitly unprojecting to 3d}},
  author={Philion, Jonah and Fidler, Sanja},
  booktitle={{Computer Vision--ECCV 2020: 16th European Conference, Glasgow, UK, August 23--28, 2020, Proceedings, Part XIV 16}},
  pages={194--210},
  year={2020},
  organization={Springer}
}

@inproceedings{kendall2018multi,
  title={{Multi-task learning using uncertainty to weigh losses for scene geometry and semantics}},
  author={Kendall, Alex and Gal, Yarin and Cipolla, Roberto},
  booktitle={{Proceedings of the IEEE conference on computer vision and pattern recognition}},
  pages={7482--7491},
  year={2018}
}

@inproceedings{carion2020end,
  title={{End-to-end object detection with transformers}},
  author={Carion, Nicolas and Massa, Francisco and Synnaeve, Gabriel and Usunier, Nicolas and Kirillov, Alexander and Zagoruyko, Sergey},
  booktitle={{European conference on computer vision}},
  pages={213--229},
  year={2020},
  organization={Springer}
}

@article{wang2022pvt,
  title={{Pvt v2: Improved baselines with pyramid vision transformer}},
  author={Wang, Wenhai and Xie, Enze and Li, Xiang and Fan, Deng-Ping and Song, Kaitao and Liang, Ding and Lu, Tong and Luo, Ping and Shao, Ling},
  journal={{Computational visual media}},
  volume={8},
  number={3},
  pages={415--424},
  year={2022},
  publisher={Springer}
}

@article{zhou2019objects,
  title={{Objects as points}},
  author={Zhou, Xingyi and Wang, Dequan and Kr{\"a}henb{\"u}hl, Philipp},
  journal={{arXiv preprint arXiv:1904.07850}  },
  year={2019}
}

@inproceedings{tian2019fcos,
  title={{Fcos: Fully convolutional one-stage object detection}},
  author={Tian, Zhi and Shen, Chunhua and Chen, Hao and He, Tong},
  booktitle={{Proceedings of the IEEE/CVF international conference on computer vision}},
  pages={9627--9636},
  year={2019}
}

@article{redmon2018yolov3,
  title={{Yolov3: An incremental improvement}},
  author={Redmon, Joseph and Farhadi, Ali},
  journal={{arXiv preprint arXiv:1804.02767}  },
  year={2018}
}

@inproceedings{wang2023exploring,
  title={{Exploring object-centric temporal modeling for efficient multi-view 3d object detection}},
  author={Wang, Shihao and Liu, Yingfei and Wang, Tiancai and Li, Ying and Zhang, Xiangyu},
  booktitle={{Proceedings of the IEEE/CVF international conference on computer vision}},
  pages={3621--3631},
  year={2023}
}

@inproceedings{hara2018can,
  title={{Can spatiotemporal 3d cnns retrace the history of 2d cnns and imagenet?}},
  author={Hara, Kensho and Kataoka, Hirokatsu and Satoh, Yutaka},
  booktitle={{Proceedings of the IEEE conference on Computer Vision and Pattern Recognition}},
  pages={6546--6555},
  year={2018}
}

@inproceedings{zhu2017flow,
  title={{Flow-guided feature aggregation for video object detection}},
  author={Zhu, Xizhou and Wang, Yujie and Dai, Jifeng and Yuan, Lu and Wei, Yichen},
  booktitle={{Proceedings of the IEEE international conference on computer vision}},
  pages={408--417},
  year={2017}
}

@article{yan2018second,
  title={{Second: Sparsely embedded convolutional detection}},
  author={Yan, Yan and Mao, Yuxing and Li, Bo},
  journal={{Sensors}},
  volume={18},
  number={10},
  pages={3337},
  year={2018},
  publisher={Multidisciplinary Digital Publishing Institute}
}

@inproceedings{yoo20203d,
  title={{3d-cvf: Generating joint camera and lidar features using cross-view spatial feature fusion for 3d object detection}},
  author={Yoo, Jin Hyeok and Kim, Yecheol and Kim, Jisong and Choi, Jun Won},
  booktitle={{Computer vision--ECCV 2020: 16th European conference, Glasgow, UK, August 23--28, 2020, proceedings, part XXVII 16}},
  pages={720--736},
  year={2020},
  organization={Springer}
}

@inproceedings{bai2022transfusion,
  title={{Transfusion: Robust lidar-camera fusion for 3d object detection with transformers}},
  author={Bai, Xuyang and Hu, Zeyu and Zhu, Xinge and Huang, Qingqiu and Chen, Yilun and Fu, Hongbo and Tai, Chiew-Lan},
  booktitle={{Proceedings of the IEEE/CVF conference on computer vision and pattern recognition}},
  pages={1090--1099},
  year={2022}
}

@article{ren2015faster,
  title={{Faster r-cnn: Towards real-time object detection with region proposal networks}},
  author={Ren, Shaoqing and He, Kaiming and Girshick, Ross and Sun, Jian},
  journal={{Advances in neural information processing systems}},
  volume={28},
  year={2015}
}

@inproceedings{yang2018pixor,
  title={{Pixor: Real-time 3d object detection from point clouds}},
  author={Yang, Bin and Luo, Wenjie and Urtasun, Raquel},
  booktitle={{Proceedings of the IEEE conference on Computer Vision and Pattern Recognition}},
  pages={7652--7660},
  year={2018}
}

@inproceedings{tan2019efficientnet,
  title={{Efficientnet: Rethinking model scaling for convolutional neural networks}},
  author={Tan, Mingxing and Le, Quoc},
  booktitle={{International conference on machine learning}},
  pages={6105--6114},
  year={2019},
  organization={PMLR}
}

@inproceedings{lin2017feature,
  title={{Feature pyramid networks for object detection}},
  author={Lin, Tsung-Yi and Doll{\'a}r, Piotr and Girshick, Ross and He, Kaiming and Hariharan, Bharath and Belongie, Serge},
  booktitle={{Proceedings of the IEEE conference on computer vision and pattern recognition}},
  pages={2117--2125},
  year={2017}
}

@inproceedings{qi2017pointnet,
  title={{Pointnet: Deep learning on point sets for 3d classification and segmentation}},
  author={Qi, Charles R and Su, Hao and Mo, Kaichun and Guibas, Leonidas J},
  booktitle={{Proceedings of the IEEE conference on computer vision and pattern recognition}},
  pages={652--660},
  year={2017}
}

@inproceedings{ronneberger2015u,
  title={{U-net: Convolutional networks for biomedical image segmentation}},
  author={Ronneberger, Olaf and Fischer, Philipp and Brox, Thomas},
  booktitle={{Medical image computing and computer-assisted intervention--MICCAI 2015: 18th international conference, Munich, Germany, October 5-9, 2015, proceedings, part III 18}},
  pages={234--241},
  year={2015},
  organization={Springer}
}

@article{li2021monocular,
  title={{Monocular 3d detection with geometric constraint embedding and semi-supervised training}},
  author={Li, Peixuan and Zhao, Huaici},
  journal={{IEEE Robotics and Automation Letters}},
  volume={6},
  number={3},
  pages={5565--5572},
  year={2021},
  publisher={IEEE}
}

@inproceedings{zhu2020deformable,
  title={{Deformable DETR: Deformable Transformers for End-to-End Object Detection}},
  author={Zhu, Xizhou and Su, Weijie and Lu, Lewei and Li, Bin and Wang, Xiaogang and Dai, Jifeng},
  booktitle={International Conference on Learning Representations},
  year={2021}
}

@inproceedings{wang2022detr3d,
  title={{Detr3d: 3d object detection from multi-view images via 3d-to-2d queries}},
  author={Wang, Yue and Guizilini, Vitor Campagnolo and Zhang, Tianyuan and Wang, Yilun and Zhao, Hang and Solomon, Justin},
  booktitle={{Conference on Robot Learning}},
  pages={180--191},
  year={2022},
  organization={PMLR}
}

@inproceedings{zhou2018voxelnet,
  title={{Voxelnet: End-to-end learning for point cloud based 3d object detection}},
  author={Zhou, Yin and Tuzel, Oncel},
  booktitle={{Proceedings of the IEEE conference on computer vision and pattern recognition}},
  pages={4490--4499},
  year={2018}
}

@inproceedings{lang2019pointpillars,
  title={{Pointpillars: Fast encoders for object detection from point clouds}},
  author={Lang, Alex H and Vora, Sourabh and Caesar, Holger and Zhou, Lubing and Yang, Jiong and Beijbom, Oscar},
  booktitle={{Proceedings of the IEEE/CVF conference on computer vision and pattern recognition}},
  pages={12697--12705},
  year={2019}
}

@article{ge2021yolox,
  title={{Yolox: Exceeding yolo series in 2021}},
  author={Ge, Zheng and Liu, Songtao and Wang, Feng and Li, Zeming and Sun, Jian},
  journal={{arXiv preprint arXiv:2107.08430}},
  year={2021}
}

@article{kuhn1955hungarian,
  title={{The Hungarian method for the assignment problem}},
  author={Kuhn, Harold W},
  journal={{Naval research logistics quarterly}},
  volume={2},
  number={1-2},
  pages={83--97},
  year={1955},
  publisher={Wiley Online Library}
}

@inproceedings{sarlin2020superglue,
  title={{Superglue: Learning feature matching with graph neural networks}},
  author={Sarlin, Paul-Edouard and DeTone, Daniel and Malisiewicz, Tomasz and Rabinovich, Andrew},
  booktitle={{Proceedings of the IEEE/CVF conference on computer vision and pattern recognition}},
  pages={4938--4947},
  year={2020}
}

@article{cox1996efficient,
  title={{An efficient implementation of Reid's multiple hypothesis tracking algorithm and its evaluation for the purpose of virtual tracking}},
  author={Cox, IJ},
  journal={{TPAMI}},
  volume={18},
  year={1996}
}

@inproceedings{ester1996density,
author = {Ester, Martin and Kriegel, Hans-Peter and Sander, J\"{o}rg and Xu, Xiaowei},
title = {{A density-based algorithm for discovering clusters in large spatial databases with noise}},
year = {1996},
publisher = {AAAI Press},
booktitle = {Proceedings of the Second International Conference on Knowledge Discovery and Data Mining},
pages = {226--231}
}

@article{cuturi2013sinkhorn,
  title={{Sinkhorn distances: Lightspeed computation of optimal transport}},
  author={Cuturi, Marco},
  journal={{Advances in neural information processing systems}},
  volume={26},
  year={2013}
}

@inproceedings{cheng2021exploring,
  title={{Exploring dynamic context for multi-path trajectory prediction}},
  author={Cheng, Hao and Liao, Wentong and Tang, Xuejiao and Yang, Michael Ying and Sester, Monika and Rosenhahn, Bodo},
  booktitle={{2021 IEEE International Conference on Robotics and Automation (ICRA)}},
  pages={12795--12801},
  year={2021},
  organization={IEEE}
}

@inproceedings{cui2019multimodal,
  title={{Multimodal trajectory predictions for autonomous driving using deep convolutional networks}},
  author={Cui, Henggang and Radosavljevic, Vladan and Chou, Fang-Chieh and Lin, Tsung-Han and Nguyen, Thi and Huang, Tzu-Kuo and Schneider, Jeff and Djuric, Nemanja},
  booktitle={{2019 IEEE International Conference on Robotics and Automation (ICRA)}},
  pages={2090--2096},
  year={2019},
  organization={IEEE}
}

@inproceedings{lee2017desire,
  title={{Desire: Distant future prediction in dynamic scenes with interacting agents}},
  author={Lee, Namhoon and Choi, Wongun and Vernaza, Paul and Choy, Christopher B and Torr, Philip HS and Chandraker, Manmohan},
  booktitle={{Proceedings of the IEEE conference on computer vision and pattern recognition}},
  pages={336--345},
  year={2017}
}

@inproceedings{uhlemann2025snapshot,
  title={{Snapshot: Towards Application-centered Models for Pedestrian Trajectory Prediction in Urban Traffic Environments}},
  author={Uhlemann, Nico and Zhou, Yipeng and Mohr, Tobias Simeon and Lienkamp, Markus},
  booktitle={{Proceedings of the Winter Conference on Applications of Computer Vision}},
  pages={1152--1162},
  year={2025}
}

@inproceedings{zhou2022hivt,
  title={{Hivt: Hierarchical vector transformer for multi-agent motion prediction}},
  author={Zhou, Zikang and Ye, Luyao and Wang, Jianping and Wu, Kui and Lu, Kejie},
  booktitle={{Proceedings of the IEEE/CVF Conference on Computer Vision and Pattern Recognition}},
  pages={8823--8833},
  year={2022}
}

@inproceedings{ngiam2021scene,
  title={{Scene Transformer: A unified architecture for predicting future trajectories of multiple agents}},
  author={Ngiam, Jiquan and Vasudevan, Vijay and Caine, Benjamin and Zhang, Zhengdong and Lewis Chiang, Hao-Tien and Ling, Jeffrey and Roelofs, Rebecca and Bewley, Alex and Liu, Chenxi and Venugopal, Ashish and Weiss, David J and Sapp, Benjamin and Chen, Zhifeng and Shlens, Jonathon},
  booktitle={International Conference on Learning Representations},
  year={2022}
}

@inproceedings{xiao2017joint,
  title={{Joint detection and identification feature learning for person search}},
  author={Xiao, Tong and Li, Shuang and Wang, Bochao and Lin, Liang and Wang, Xiaogang},
  booktitle={{Proceedings of the IEEE conference on computer vision and pattern recognition}},
  pages={3415--3424},
  year={2017}
}

@inproceedings{zheng2017person,
  title={{Person re-identification in the wild}},
  author={Zheng, Liang and Zhang, Hengheng and Sun, Shaoyan and Chandraker, Manmohan and Yang, Yi and Tian, Qi},
  booktitle={{Proceedings of the IEEE conference on computer vision and pattern recognition}},
  pages={1367--1376},
  year={2017}
}

@inproceedings{he2016deep,
  title={{Deep residual learning for image recognition}},
  author={He, Kaiming and Zhang, Xiangyu and Ren, Shaoqing and Sun, Jian},
  booktitle={{Proceedings of the IEEE conference on computer vision and pattern recognition}},
  pages={770--778},
  year={2016}
}

@article{geiger2013vision,
  title={{Vision meets robotics: The kitti dataset}},
  author={Geiger, Andreas and Lenz, Philip and Stiller, Christoph and Urtasun, Raquel},
  journal={{The international journal of robotics research}},
  volume={32},
  number={11},
  pages={1231--1237},
  year={2013},
  publisher={Sage Publications Sage UK: London, England}
}

@inproceedings{caesar2020nuscenes,
  title={{nuscenes: A multimodal dataset for autonomous driving}},
  author={Caesar, Holger and Bankiti, Varun and Lang, Alex H and Vora, Sourabh and Liong, Venice Erin and Xu, Qiang and Krishnan, Anush and Pan, Yu and Baldan, Giancarlo and Beijbom, Oscar},
  booktitle={{Proceedings of the IEEE/CVF conference on computer vision and pattern recognition}},
  pages={11621--11631},
  year={2020}
}

@inproceedings{sun2020scalability,
  title={{Scalability in perception for autonomous driving: Waymo open dataset}},
  author={Sun, Pei and Kretzschmar, Henrik and Dotiwalla, Xerxes and Chouard, Aurelien and Patnaik, Vijaysai and Tsui, Paul and Guo, James and Zhou, Yin and Chai, Yuning and Caine, Benjamin and others},
  booktitle={{Proceedings of the IEEE/CVF conference on computer vision and pattern recognition}},
  pages={2446--2454},
  year={2020}
}

@inproceedings{han2023mmptrack,
  title={{Mmptrack: Large-scale densely annotated multi-camera multiple people tracking benchmark}},
  author={Han, Xiaotian and You, Quanzeng and Wang, Chunyu and Zhang, Zhizheng and Chu, Peng and Hu, Houdong and Wang, Jiang and Liu, Zicheng},
  booktitle={{Proceedings of the IEEE/CVF Winter Conference on Applications of Computer Vision}},
  pages={4860--4869},
  year={2023}
}

@inproceedings{chang2019argoverse,
  title={{Argoverse: 3d tracking and forecasting with rich maps}},
  author={Chang, Ming-Fang and Lambert, John and Sangkloy, Patsorn and Singh, Jagjeet and Bak, Slawomir and Hartnett, Andrew and Wang, De and Carr, Peter and Lucey, Simon and Ramanan, Deva and others},
  booktitle={{Proceedings of the IEEE/CVF conference on computer vision and pattern recognition}},
  pages={8748--8757},
  year={2019}
}

@inproceedings{Argoverse2,
  author = {Wilson, Benjamin and Qi, William and Agarwal, Tanmay and Lambert, John and Singh, Jagjeet and Khandelwal, Siddhesh and Pan, Bowen and Kumar, Ratnesh and Hartnett, Andrew and Kaesemodel Pontes, Jhony and Ramanan, Deva and Carr, Peter and Hays, James},
  title = {{Argoverse 2: Next Generation Datasets for Self-driving Perception and Forecasting}},
  booktitle = {{Proceedings of the Neural Information Processing Systems Track on Datasets and Benchmarks (NeurIPS Datasets and Benchmarks 2021)}},
  year = {2021}
}

@article{MOTChallenge2015,
	title = {{MOTChallenge 2015: Towards a Benchmark for Multi-Target Tracking}},
	shorttitle = {MOTChallenge 2015},
	journal = {{arxiv preprint arXiv:1504.01942}},
	author = {Leal-Taix\'{e}, L. and Milan, A. and Reid, I. and Roth, S. and Schindler, K.},
	month = apr,
	year = {2015}
}

@article{MOT16,
	title = {{MOT16: A Benchmark for Multi-Object Tracking}},
	journal = {{arxiv preprint arXiv:1603.00831}},
	author = {Milan, A. and Leal-Taix\'{e}, L. and Reid, I. and Roth, S. and Schindler, K.},
	month = mar,
	year = {2016}
}

@article{MOTChallenge20,
    title={{MOT20: A benchmark for multi object tracking in crowded scenes}},
	journal = {{arxiv preprint arXiv:2003.09003}},
	author = {Dendorfer, P. and Rezatofighi, H. and Milan, A. and Shi, J. and Cremers, D. and Reid, I. and Roth, S. and Schindler, K. and Leal-Taix\'{e}, L. },
	month = mar,
	year = {2020}
}

@inproceedings{MOTS20,
  title = {{MOTS: Multi-Object Tracking and Segmentation}},
  booktitle = {2019 IEEE/CVF Conference on Computer Vision and Pattern Recognition (CVPR)},
  publisher = {IEEE},
  author = {Voigtlaender,  Paul and Krause,  Michael and Osep,  Aljosa and Luiten,  Jonathon and Sekar,  Berin Balachandar Gnana and Geiger,  Andreas and Leibe,  Bastian},
  year = {2019},
  month = jun,
  pages = {7934--7943}
}

@inproceedings{sun2022dancetrack,
  title={{Dancetrack: Multi-object tracking in uniform appearance and diverse motion}},
  author={Sun, Peize and Cao, Jinkun and Jiang, Yi and Yuan, Zehuan and Bai, Song and Kitani, Kris and Luo, Ping},
  booktitle={{Proceedings of the IEEE/CVF conference on computer vision and pattern recognition}},
  pages={20993--21002},
  year={2022}
}

@inproceedings{fabbri2018learning,
  title={{Learning to detect and track visible and occluded body joints in a virtual world}},
  author={Fabbri, Matteo and Lanzi, Fabio and Calderara, Simone and Palazzi, Andrea and Vezzani, Roberto and Cucchiara, Rita},
  booktitle={{Proceedings of the European conference on computer vision (ECCV)}},
  pages={430--446},
  year={2018}
}

@inproceedings{yu2020bdd100k,
  title={{Bdd100k: A diverse driving dataset for heterogeneous multitask learning}},
  author={Yu, Fisher and Chen, Haofeng and Wang, Xin and Xian, Wenqi and Chen, Yingying and Liu, Fangchen and Madhavan, Vashisht and Darrell, Trevor},
  booktitle={{Proceedings of the IEEE/CVF conference on computer vision and pattern recognition}},
  pages={2636--2645},
  year={2020}
}

@inproceedings{cui2023sportsmot,
  title={{Sportsmot: A large multi-object tracking dataset in multiple sports scenes}},
  author={Cui, Yutao and Zeng, Chenkai and Zhao, Xiaoyu and Yang, Yichun and Wu, Gangshan and Wang, Limin},
  booktitle={{Proceedings of the IEEE/CVF international conference on computer vision}},
  pages={9921--9931},
  year={2023}
}

@inproceedings{zheng2024nettrack,
  title={{Nettrack: Tracking highly dynamic objects with a net}},
  author={Zheng, Guangze and Lin, Shijie and Zuo, Haobo and Fu, Changhong and Pan, Jia},
  booktitle={{Proceedings of the IEEE/CVF Conference on Computer Vision and Pattern Recognition}},
  pages={19145--19155},
  year={2024}
}

@inproceedings{yang2019video,
  title={{Video instance segmentation}},
  author={Yang, Linjie and Fan, Yuchen and Xu, Ning},
  booktitle={{Proceedings of the IEEE/CVF international conference on computer vision}},
  pages={5188--5197},
  year={2019}
}

@inproceedings{houston2021one,
  title={{One thousand and one hours: Self-driving motion prediction dataset}},
  author={Houston, John and Zuidhof, Guido and Bergamini, Luca and Ye, Yawei and Chen, Long and Jain, Ashesh and Omari, Sammy and Iglovikov, Vladimir and Ondruska, Peter},
  booktitle={{Conference on Robot Learning}},
  pages={409--418},
  year={2021},
  organization={PMLR}
}

@inproceedings{huang2018apolloscape,
  title={{The apolloscape dataset for autonomous driving}},
  author={Huang, Xinyu and Cheng, Xinjing and Geng, Qichuan and Cao, Binbin and Zhou, Dingfu and Wang, Peng and Lin, Yuanqing and Yang, Ruigang},
  booktitle={{Proceedings of the IEEE conference on computer vision and pattern recognition workshops}},
  pages={954--960},
  year={2018}
}

@inproceedings{phillips2021deep,
  title={{Deep multi-task learning for joint localization, perception, and prediction}},
  author={Phillips, John and Martinez, Julieta and B{\^a}rsan, Ioan Andrei and Casas, Sergio and Sadat, Abbas and Urtasun, Raquel},
  booktitle={{Proceedings of the IEEE/CVF Conference on Computer Vision and Pattern Recognition}},
  pages={4679--4689},
  year={2021}
}

@article{manole2024uolo,
  title={{UOLO: A Multitask U-Net YOLO hybrid model for railway scene understanding}},
  author={Manole, Alexandru and Dioșan, Laura-Silvia},
  journal={IEEE Transactions on Intelligent Vehicles},
  year={2024},
  publisher={IEEE}
}

@article{wang2024you,
  title={{You only look at once for real-time and generic multi-task}},
  author={Wang, Jiayuan and Wu, QM Jonathan and Zhang, Ning},
  journal={IEEE Transactions on Vehicular Technology},
  volume={73},
  number={9},
  pages={12625--12637},
  year={2024},
  publisher={IEEE}
}

@article{he2025md,
  title={{MD-SLAM: a multi-task deep-learning-based visual SLAM system in dynamic environments}},
  author={He, Ying and Zeng, Xinyu and Yu, F Richard and Zhong, Zhengjun and Liu, Zhiquan and Zhou, Guang},
  journal={IEEE Transactions on Vehicular Technology},
  year={2025},
  publisher={IEEE}
}

@article{azak2024unified,
  title={{A Unified Monocular Vision-Based Driving Model for Autonomous Vehicles With Multi-Task Capabilities}},
  author={Azak, Salim and Bozkaya, F{\i}rat and T{\i}{\u{g}}l{\i}o{\u{g}}lu, {\c{S}}{\"u}kr{\"u}can and Yusefi, Abdullah and Durdu, Akif},
  journal={IEEE Transactions on Intelligent Vehicles},
  year={2024},
  publisher={IEEE}
}

@article{wang2025mtfenet,
  title={{MTFENet: A Multi-Task Autonomous Driving Network for Real-Time Target Perception}},
  author={Wang, Qiang and Xue, Yongchong and Lyu, Shuchang and Cheng, Guangliang and Yang, Shaoyan and Jin, Xin},
  journal={IEEE Open Journal of Vehicular Technology},
  year={2025},
  publisher={IEEE}
}

@article{dauner2026123d,
  title={{123D: Unifying Multi-Modal Autonomous Driving Data at Scale}},
  author={Dauner, Daniel and Charraut, Valentin and Berle, Bastian and Li, Tianyu and Nguyen, Long and Wang, Jiabao and Jing, Changhui and Igl, Maximilian and Caesar, Holger and Ivanovic, Boris and others},
  journal={arXiv preprint arXiv:2605.08084},
  year={2026}
}

@inproceedings{feng2024unitraj,
  title={{Unitraj: A unified framework for scalable vehicle trajectory prediction}},
  author={Feng, Lan and Bahari, Mohammadhossein and Amor, Kaouther Messaoud Ben and Zablocki, {\'E}loi and Cord, Matthieu and Alahi, Alexandre},
  booktitle={European Conference on Computer Vision},
  pages={106--123},
  year={2024},
  organization={Springer}
}

@article{wang2025alpamayo,
  title={{Alpamayo-r1: Bridging reasoning and action prediction for generalizable autonomous driving in the long tail}},
  author={Wang, Yan and Luo, Wenjie and Bai, Junjie and Cao, Yulong and Che, Tong and Chen, Ke and Chen, Yuxiao and Diamond, Jenna and Ding, Yifan and Ding, Wenhao and others},
  journal={arXiv preprint arXiv:2511.00088},
  year={2025}
}

@inproceedings{tian2025nuscenes,
  title={{NuScenes-SpatialQA: A Spatial Understanding and Reasoning Benchmark for Vision-Language Models in Autonomous Driving}},
  author={Tian, Kexin and Mao, Jingrui and Zhang, Yunlong and Jiang, Jiwan and Zhou, Yang and Tu, Zhengzhong},
  booktitle={Proceedings of the IEEE/CVF International Conference on Computer Vision},
  pages={4567--4576},
  year={2025}
}

@article{song2024robustness,
  title={{Robustness-aware 3d object detection in autonomous driving: A review and outlook}},
  author={Song, Ziying and Liu, Lin and Jia, Feiyang and Luo, Yadan and Jia, Caiyan and Zhang, Guoxin and Yang, Lei and Wang, Li},
  journal={{IEEE Transactions on Intelligent Transportation Systems}},
  year={2024},
  publisher={IEEE}
}

@article{mao20233d,
  title={{3D object detection for autonomous driving: A comprehensive survey}},
  author={Mao, Jiageng and Shi, Shaoshuai and Wang, Xiaogang and Li, Hongsheng},
  journal={{International Journal of Computer Vision}},
  volume={131},
  number={8},
  pages={1909--1963},
  year={2023},
  publisher={Springer}
}

@article{ma20233d,
  title={{3d object detection from images for autonomous driving: a survey}},
  author={Ma, Xinzhu and Ouyang, Wanli and Simonelli, Andrea and Ricci, Elisa},
  journal={{IEEE Transactions on Pattern Analysis and Machine Intelligence}},
  volume={46},
  number={5},
  pages={3537--3556},
  year={2023},
  publisher={IEEE}
}

@article{liu2023echoes,
  title={{Echoes beyond points: Unleashing the power of raw radar data in multi-modality fusion}},
  author={Liu, Yang and Wang, Feng and Wang, Naiyan and Zhang, Zhao-Xiang},
  journal={{Advances in Neural Information Processing Systems}},
  volume={36},
  pages={53964--53982},
  year={2023}
}

@article{wang2023multi,
  title={{Multi-modal 3d object detection in autonomous driving: A survey and taxonomy}},
  author={Wang, Li and Zhang, Xinyu and Song, Ziying and Bi, Jiangfeng and Zhang, Guoxin and Wei, Haiyue and Tang, Liyao and Yang, Lei and Li, Jun and Jia, Caiyan and others},
  journal={{IEEE Transactions on Intelligent Vehicles}},
  volume={8},
  number={7},
  pages={3781--3798},
  year={2023},
  publisher={IEEE}
}

@article{yao2023radar,
  title={{Radar-camera fusion for object detection and semantic segmentation in autonomous driving: A comprehensive review}},
  author={Yao, Shanliang and Guan, Runwei and Huang, Xiaoyu and Li, Zhuoxiao and Sha, Xiangyu and Yue, Yong and Lim, Eng Gee and Seo, Hyungjoon and Man, Ka Lok and Zhu, Xiaohui and others},
  journal={{IEEE Transactions on Intelligent Vehicles}},
  volume={9},
  number={1},
  pages={2094--2128},
  year={2023},
  publisher={IEEE}
}

@article{aung2024review,
  title={{A review of lidar-based 3d object detection via deep learning approaches towards robust connected and autonomous vehicles}},
  author={Aung, Nang Htet Htet and Sangwongngam, Paramin and Jintamethasawat, Rungroj and Shah, Shashi and Wuttisittikulkij, Lunchakorn},
  journal={{IEEE Transactions on Intelligent Vehicles}},
  year={2024},
  publisher={IEEE}
}

@article{luo2021multiple,
  title={{Multiple object tracking: A literature review}},
  author={Luo, Wenhan and Xing, Junliang and Milan, Anton and Zhang, Xiaoqin and Liu, Wei and Kim, Tae-Kyun},
  journal={{Artificial intelligence}},
  volume={293},
  pages={103448},
  year={2021},
  publisher={Elsevier}
}

@article{bashar2209multiple,
  title={{Multiple object tracking in recent times: A literature review.}},
  author={Bashar, M and Islam, S and Hussain, KK and Hasan, MB and Rahman, ABMA and Kabir, MH},
  journal={{arXiv preprint arXiv:2209.04796}},   
  year={2022}
}

@article{agrawal2024systematic,
  title={{A systematic survey on recent deep learning-based approaches to multi-object tracking}},
  author={Agrawal, Harshit and Halder, Agrya and Chattopadhyay, Pratik},
  journal={{Multimedia Tools and Applications}},
  volume={83},
  number={12},
  pages={36203--36259},
  year={2024},
  publisher={Springer}
}

@article{hassan2024multi,
  title={{Multi-object tracking: a systematic literature review}},
  author={Hassan, Saif and Mujtaba, Ghulam and Rajput, Asif and Fatima, Noureen},
  journal={{Multimedia Tools and Applications}},
  volume={83},
  number={14},
  pages={43439--43492},
  year={2024},
  publisher={Springer}
}

@article{rakai2022data,
  title={{Data association in multiple object tracking: A survey of recent techniques}},
  author={Rakai, Lionel and Song, Huansheng and Sun, ShiJie and Zhang, Wentao and Yang, Yanni},
  journal={{Expert systems with applications}},
  volume={192},
  pages={116300},
  year={2022},
  publisher={Elsevier}
}

@article{leon2021review,
  title={{A review of tracking and trajectory prediction methods for autonomous driving}},
  author={Leon, Florin and Gavrilescu, Marius},
  journal={{Mathematics}},
  volume={9},
  number={6},
  pages={660},
  year={2021},
  publisher={MDPI}
}

@article{huang2022survey,
  title={{A survey on trajectory-prediction methods for autonomous driving}},
  author={Huang, Yanjun and Du, Jiatong and Yang, Ziru and Zhou, Zewei and Zhang, Lin and Chen, Hong},
  journal={{IEEE Transactions on Intelligent Vehicles}},
  volume={7},
  number={3},
  pages={652--674},
  year={2022},
  publisher={IEEE}
}

@article{karle2022scenario,
  title={{Scenario understanding and motion prediction for autonomous vehicles—review and comparison}},
  author={Karle, Phillip and Geisslinger, Maximilian and Betz, Johannes and Lienkamp, Markus},
  journal={{IEEE Transactions on Intelligent Transportation Systems}},
  volume={23},
  number={10},
  pages={16962--16982},
  year={2022},
  publisher={IEEE}
}

@article{ding2023incorporating,
  title={{Incorporating driving knowledge in deep learning based vehicle trajectory prediction: A survey}},
  author={Ding, Zhezhang and Zhao, Huijing},
  journal={{IEEE Transactions on Intelligent Vehicles}},
  volume={8},
  number={8},
  pages={3996--4015},
  year={2023},
  publisher={IEEE}
}

@article{bharilya2024machine,
  title={{Machine learning for autonomous vehicle's trajectory prediction: A comprehensive survey, challenges, and future research directions}},
  author={Bharilya, Vibha and Kumar, Neetesh},
  journal={{Vehicular Communications}},
  volume={46},
  pages={100733},
  year={2024},
  publisher={Elsevier}
}

@article{huang2023review,
  title={{A review of deep learning-based vehicle motion prediction for autonomous driving}},
  author={Huang, Renbo and Zhuo, Guirong and Xiong, Lu and Lu, Shouyi and Tian, Wei},
  journal={{Sustainability}},
  volume={15},
  number={20},
  pages={14716},
  year={2023},
  publisher={MDPI}
}

@article{uhlemann2024evaluating,
  title={{Evaluating pedestrian trajectory prediction methods with respect to autonomous driving}},
  author={Uhlemann, Nico and Fent, Felix and Lienkamp, Markus},
  journal={{IEEE Transactions on Intelligent Transportation Systems}},
  year={2024},
  publisher={IEEE}
}

@article{chib2023recent,
  title={{Recent advancements in end-to-end autonomous driving using deep learning: A survey}},
  author={Chib, Pranav Singh and Singh, Pravendra},
  journal={{IEEE Transactions on Intelligent Vehicles}},
  volume={9},
  number={1},
  pages={103--118},
  year={2023},
  publisher={IEEE}
}

@article{chen2024end,
  title={{End-to-end autonomous driving: Challenges and frontiers}},
  author={Chen, Li and Wu, Penghao and Chitta, Kashyap and Jaeger, Bernhard and Geiger, Andreas and Li, Hongyang},
  journal={{IEEE Transactions on Pattern Analysis and Machine Intelligence}},
  year={2024},
  publisher={IEEE}
}

@article{tampuu2020survey,
  title={{A survey of end-to-end driving: Architectures and training methods}},
  author={Tampuu, Ardi and Matiisen, Tambet and Semikin, Maksym and Fishman, Dmytro and Muhammad, Naveed},
  journal={{IEEE Transactions on Neural Networks and Learning Systems}},
  volume={33},
  number={4},
  pages={1364--1384},
  year={2020},
  publisher={IEEE}
}

@article{singh2023end,
  title={{End-to-end autonomous driving using deep learning: A systematic review}},
  author={Singh, Apoorv},
  journal={{arXiv preprint arXiv:2311.18636}},
  year={2023}
}

@article{al2024end,
  title={{End-to-End Autonomous Driving in CARLA: A Survey}},
  author={Al Ozaibi, Youssef and Hina, Manolo Dulva and Ramdane-Cherif, Amar},
  journal={{IEEE Access}},
  year={2024},
  publisher={IEEE}
}

@article{dal2024joint,
  title = {{Joint Perception and Prediction for Autonomous Driving: A Survey}},
  volume = {26},
  number = {12},
  journal = {IEEE Transactions on Intelligent Transportation Systems},
  publisher = {Institute of Electrical and Electronics Engineers (IEEE)},
  author = {Dal'Col,  Lucas and Oliveira,  Miguel and Santos,  Vítor},
  year = {2025},
  month = Dec,
  pages = {21427--21452}
}

@inproceedings{xu2024towards,
  title={{Towards motion forecasting with real-world perception inputs: Are end-to-end approaches competitive?}},
  author={Xu, Yihong and Chambon, Lo{\"\i}ck and Zablocki, {\'E}loi and Chen, Micka{\"e}l and Alahi, Alexandre and Cord, Matthieu and P{\'e}rez, Patrick},
  booktitle={{2024 IEEE International Conference on Robotics and Automation (ICRA)}},
  pages={18428--18435},
  year={2024},
  organization={IEEE}
}

\vspace{11pt}
\vspace{-40pt}
\begin{IEEEbiography}
[{\includegraphics[width=1in,height=1.25in,clip,keepaspectratio]{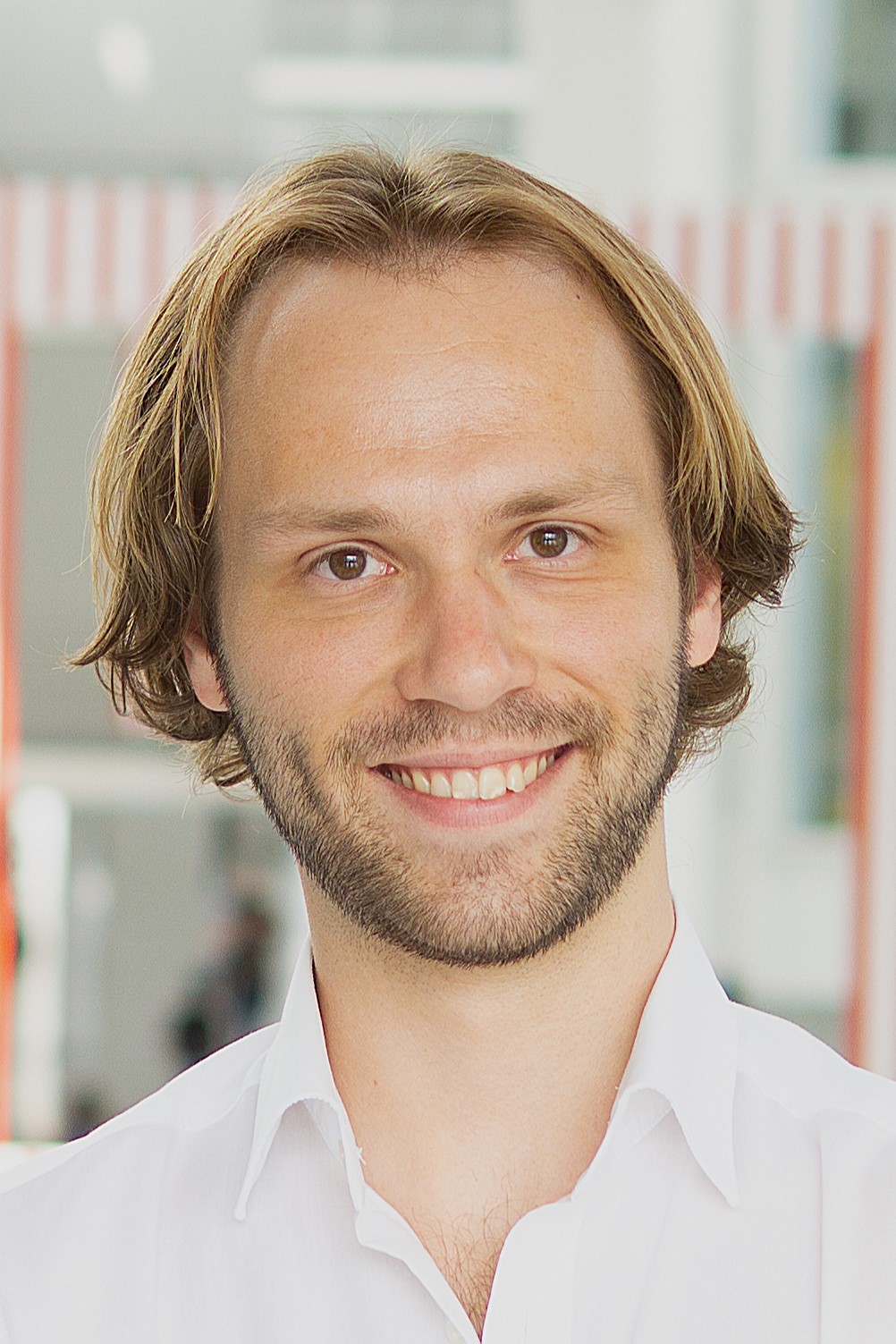}}]{Loïc Stratil}
received the B.Sc. degree from the University of Stuttgart, Stuttgart, Germany in 2020 and M.Sc. degree from the Technical University of Munich (TUM), Munich, Germany, in 2023, where he is currently pursuing the Ph.D. degree in mechanical engineering with the Institute of Automotive Technology. His research interests include object detection, multi-object-tracking and motion prediction as well as related applications for autonomous driving with a focus on robust real-world application. 
\end{IEEEbiography}

\vspace{11pt}
\vspace{-40pt}
\begin{IEEEbiography}
[{\includegraphics[width=1in,height=1.25in,clip,keepaspectratio]{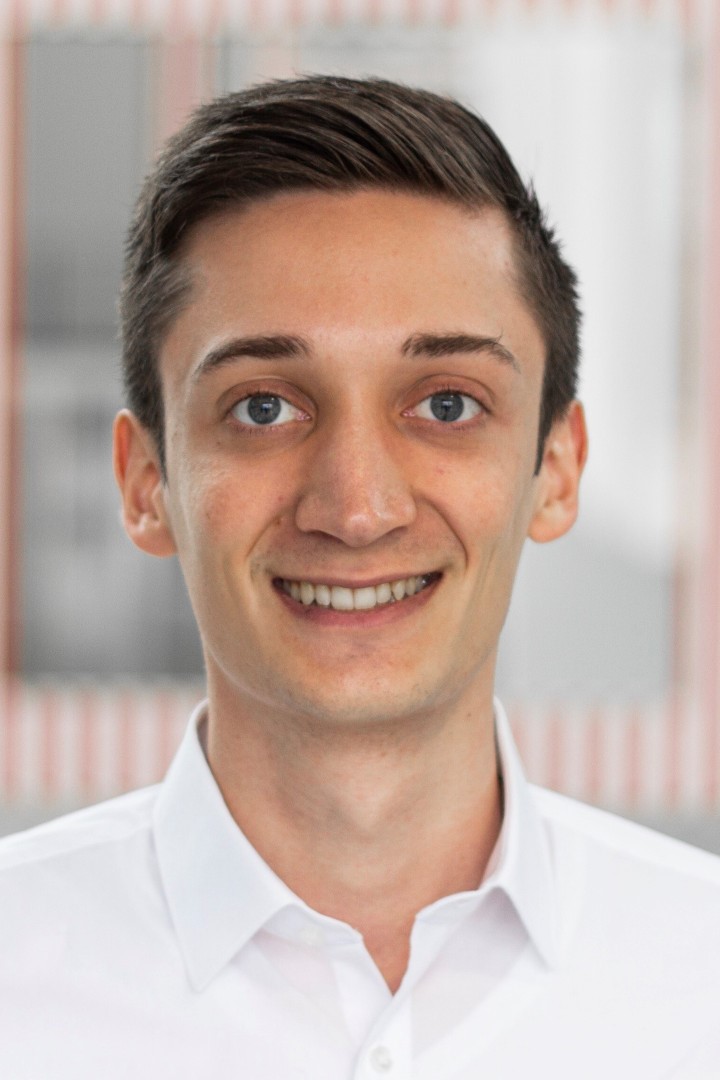}}]{Felix Fent}
received the B.Sc. and M.Sc. degrees from the Technical University of Munich (TUM), Munich, Germany, in 2018 and 2020, respectively, where he is currently pursuing a Ph.D. degree in mechanical engineering with the Institute of Automotive Technology. His research interests include radar-based perception, sensor fusion and multi-modal object detection approaches with a focus on real-world applications.
\end{IEEEbiography}

\vspace{11pt}
\vspace{-33pt}
\begin{IEEEbiography}
[{\includegraphics[width=1in,height=1.25in,clip,keepaspectratio]{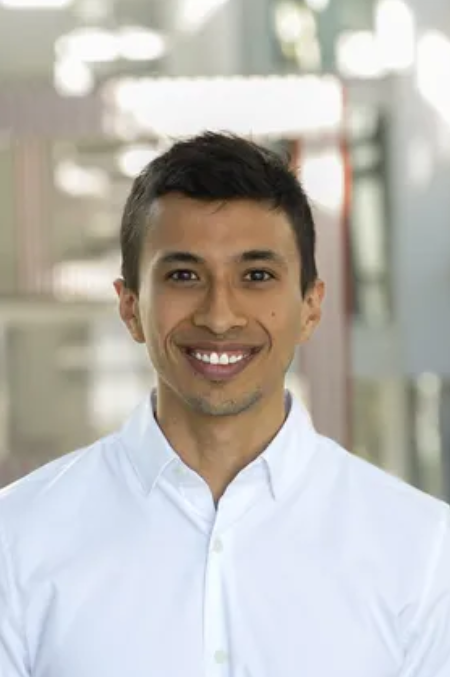}}]{Esteban Rivera}
received B.Sc. degrees in electronic engineering and physics from the Universidad de los Andes, Bogota, Colombia in 2016, and the M.Sc. degree in electronic engineering from the Karlsruhe Institute for Technology (KIT), Germany. Currently, he is pursuing a Ph.D. degree in mechanical engineering with the Institute of Automotive Technology at the Technical University of Munich (TUM). His research interests include data-efficient multi-modal perception, semi-supervised learning and visual foundational models.
\end{IEEEbiography}

\vspace{11pt}
\vspace{-40pt}
\begin{IEEEbiography}
[{\includegraphics[width=1in,height=1.25in,clip,keepaspectratio]{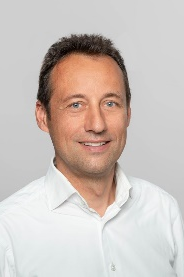}}]{Markus Lienkamp}
 professor of the Institute of Automotive Technology at the Technical University of Munich (TUM) since 2009, researches in the field of autonomous vehicles, aiming to create an open-source software platform. After studying mechanical engineering at TU Darmstadt and Cornell University, Prof. Lienkamp earned his doctorate from TU Darmstadt in 1995. He then joined Volkswagen’s trainee program, worked in a Ford–Volkswagen (VW) joint venture in Portugal, led VW’s commercial vehicle brake testing department, and later headed the Electronics and Vehicle Research department.
\end{IEEEbiography}

\vfill
\end{document}